\documentclass[twoside]{article}

\usepackage[accepted]{aistats2021}
\usepackage[round]{natbib}

\usepackage[utf8]{inputenc} % allow utf-8 input
\usepackage[T1]{fontenc}    % use 8-bit T1 fonts
\usepackage{hyperref}       % hyperlinks
\usepackage{url}            % simple URL typesetting
\usepackage{booktabs}       % professional-quality tables
\usepackage{amsfonts}       % blackboard math symbols
\usepackage{nicefrac}       % compact symbols for 1/2, etc.
\usepackage{microtype}      % microtypography
\usepackage{amsmath}
\usepackage{cuted}
\usepackage{subfig}
\usepackage{xspace}
\usepackage[nolist,smaller]{acronym}  
\usepackage{xcolor}
\usepackage{xfrac}

\usepackage{enumitem}

\newtheorem{remark}{Remark}
\newtheorem{proposition}{Proposition}

\usepackage{graphicx}

\newcommand{\acro}[1]{\textsc{#1}\xspace }

\newcommand{\ML}{\acro{\smaller ML}}%Machine Learning
\newcommand{\KLD}{\acro{\smaller KLD}}
\newcommand{\DP}{\acro{\smaller DP}}%Differential Privacy
\newcommand{\DGP}{\acro{\smaller DGP}}
\newcommand{\SDGP}{\acro{\smaller S-DGP}}%Synthetic-Data generating Process
\newcommand{\GAN}{\acro{\smaller GAN}}
\newcommand{\DPGAN}{\acro{\smaller DP-GAN}}
\newcommand{\PATEGAN}{\acro{\smaller PATE-GAN}}
\newcommand{\Mopen}{\acro{\smaller \textit{M}-open}}

\newcommand{\BD}{\acro{\smaller $\beta$D}}%beta-divergence

\DeclareMathOperator*{\argmin}{arg\,min}

\newtheorem{definition}{Definition}

\begin{document}



\twocolumn[

\aistatstitle{Foundations of Bayesian Learning from Synthetic Data}

\aistatsauthor{ Harrison Wilde \And Jack Jewson \And Sebastian Vollmer \And Chris Holmes }

\aistatsaddress{ Department of Statistics\\University of Warwick \And Barcelona GSE\\Universitat Pompeu Fabra \And Department of Statistics,\\ Mathematics Institute\\University of Warwick \And Department of Statistics\\University of Oxford;\\The Alan Turing Institute } 

]

\begin{abstract}
  There is significant growth and interest in the use of synthetic data as an enabler for machine learning in environments where the release of real data is restricted due to privacy or availability constraints. Despite a large number of methods for synthetic data generation, there are comparatively few results on the statistical properties of models learnt on synthetic data, and fewer still for situations where a researcher wishes to augment real data with another party's synthesised data. We use a Bayesian paradigm to characterise the updating of model parameters when learning in these settings, demonstrating that caution should be taken when applying conventional learning algorithms without appropriate consideration of the synthetic data generating process and learning task. Recent results from general Bayesian updating support a novel and robust approach to Bayesian synthetic-learning founded on decision theory that outperforms standard approaches across repeated experiments on supervised learning and inference problems.
\end{abstract}
    
% \section{TODO}
% \begin{itemize}
%   we only consider ii.d
%   truncate      
%\item trade of of using learners for evaluation
%     \item sdfdsThis research was supported in part through computational resources provided by The Alan Turing Institute and with the help of a generous gift from Microsoft Corporation.

% \end{itemize}

% https://docs.google.com/document/d/1BrFwalnFh15e15cG1BjECgumJ2Pc_Fdu2kJKqDKMLnY/edit?usp=sharing

\section{Introduction}

Privacy enhancing technologies comprise an area of rapid growth \citep{rspet}. An important aspect of this field concerns the release of privatised versions of data for learning.
% (PETs)mitigating risk of information disclosure whilst supporting the use of data for learning \citep{rspet}. One particular PET allows for the preservation of individual privacy via an information `keeper' $K$ holding sensitive data, and releasing a privatised version to a `learner' $L$. 
% The intricacies of this task depend significantly on the way in which privacy is defined, and 
Simply \textit{anonymising} data is not sufficient to guarantee individual privacy \citep[e.g.][]{rocher2019estimating}. Instead, 
we refer to the large body of work on 
Differential Privacy (\DP) \citep{dwork2006calibrating} 
%defines a robust and rigorous privacy \textit{guarantee}. This 
% constitutes a statistical promise in the form of a 
defining bounds on the probability that an adversary may identify whether a particular observation is present in a dataset in the situation where they have access to all other observations in the dataset. \DP{}'s formulation is context-dependent across the literature; we amalgamate definitions regarding adjacent datasets by \cite{dwork2014algorithmic, dwork2009differential} below:
\begin{definition}[$(\varepsilon, \delta)$-differential privacy]
A randomised function or algorithm $\mathcal{K}$ is said to be $(\varepsilon, \delta)$-differentially private if for all pairs of \textbf{adjacent, equally-sized} datasets $D$ and $D'$ that differ in one observation  % $\operatorname{HammingDistance}(D, D') \le 1$ 
and all $S \subseteq \operatorname{Range}(\mathcal{K})$,
\begin{equation}
\operatorname{Pr}[\mathcal{K}(D) \in S] \leq e^{\varepsilon} \times \operatorname{Pr}\left[\mathcal{K}\left(D^{\prime}\right) \in S\right]  + \delta
\end{equation}
\end{definition}
The current state of the art privatises Generative Adversarial Networks (\GAN{}s) \citep{goodfellow2014generative} through adjustments to their learning processes such that their outputs fulfil a \DP guarantee specified at the point of training \citep{jordon2018pate,xie2018differentially}.

Despite these contributions, there is a fundamental knowledge gap surrounding how, from a statistical perspective, one should learn from privatised synthetic data. Progress has been made for simple exponential family and regression models \citep{bernstein2018differentially, bernstein2019differentially}.
%, allowing for well-calibrated Bayesian posteriors in the face of privatised sufficient statistics. 
However, the uses of such `simple' models for modern \ML problems are limited. 
% and thus more work is required to offer fundamentals in the scenario where one wants to maximise their potential benefit from synthetic data outside of the constraints of particular model families.

In this paper we ask what does it mean to learn from synthetic data? And how can we improve upon our inferences and predictions given that we acknowledge the privatised synthetic nature of the data? In doing so we adopt the \Mopen world viewpoint \citep{bernardo2001bayesian} associated with model misspecification. We acknowledge that correctly modelling a privacy preserving mechanism such as a black-box generative model or complex noise convolution is often intractable. 

This results in models that are misspecificed by design, providing two insights that we explore in this paper: firstly that, when left unchecked, the Bayesian inference machine learns about the model parameters minimising the Kullback-Leibler divergence (\KLD) to the synthetic data generating process (\SDGP) \citep{berk1966limiting, walker2013bayesian} rather than the true data generating process (\DGP), and secondly that improved performance can be gained by considering robust inference methods that acknowledge this misspecification.

To achieve this, we investigate models based on a mix of simulated and real-world data to offer empirical insights on the learning procedure when a varying amount of real data is available to the user to be augmented with some unspecified amount of privatised synthetic data. 
% This reveals the existence of desirable quantities of synthetic data to use and model parameters to set. In turn, this motivates a procedure that allows for these points to be discovered in practice through partial two-way communication between some $L$ and `keeper' $K$. Alternatively, when communication is not available, $L$ may hold back some real data of their own in order to evaluate how much synthetic data is appropriate to use and thus must manage the trade-off between this hold-out set and the effect this loss of real data may have on their inference.

% Additionally, we find that the learner's specific task can often determine the efficacy and usability of synthetic data, especially when a strict privacy guarantee is required. In some cases it appears that synthetic data produced through even the most recent advances in \DP does not prove useful at all in learning a model when a small amount of real data is available; whilst in others mode collapse and concentration in the learned \SDGP arising through \GAN architectures can mean that small amounts of very private data can counter-intuitively be more representative and helpful than even real data drawn from the \DGP.

The contributions of our work can be summarised as:
\begin{enumerate}[leftmargin=*,noitemsep]
\vspace{-0.2cm}
    \item Learning from synthetic data can lead to unpredictable outcomes, due to varying levels of model misspecification introduced by generation and privacy preservation.
    \item Robust Bayesian inference offers improvements over classical Bayes when learning from synthetic data.
    \item Real and synthetic data can be used in tandem to achieve practical effectiveness through the discovery of desirable stopping points for learning, and optimal model configurations.
    \item Consideration of the preferred properties of the inference procedure are critical; the specific task at hand can determine how best to use synthetic data.
\end{enumerate}

Throughout this research we adopt the Bayesian standpoint utilising recent developments in generalised Bayesian updating \citep{bissiri2016general} and minimum divergence inference \citep{jewson2018principles}, but note that many of the observations hold in the frequentist paradigm also.

\section{Problem Formulation}{\label{Sec:Foundations}}

We outline the inference problem as follows,
\begin{itemize}[leftmargin=*,noitemsep]
\item Let $x_{1:n}$ denote a training set of $n$ exchangeable observations from Nature's
true \DGP, $F_{0}(x)$, such that $x_{1:n}\sim F_{0}(x)$; we suppose $x_{i}\in\mathbb{R}^{d}$. These real observations
are held privately by a data keeper $K$. 
\item $K$ uses data $x_{1:n}$ to produce an $(\epsilon, \delta)$-differentially private synthetic data generating mechanism (\SDGP). With a slight abuse of notation we use 
$\mathcal{G}_{\varepsilon, \delta}(x_{1:n})$ to denote the \SDGP, noting this will cover the case where $\mathcal{G}_{\varepsilon, \delta}$ is a fully  generative model as well as when it involves directly privatising the finite data $x_{1:n}$ (see discussion on the \SDGP below).%\footnote{In the case of the Laplace mechanism \citep{dwork2014algorithmic} the sampling distribution is not a function of the private data $x_{1:n}$. Instead, it corresponds to a convolution of the \DGP with a sufficiently scaled Laplace distribution. Sampling one $z_i$ in this case requires using one private data item of $K$. }.
%, $\mathcal{G}_{\varepsilon, \delta}(F_{0})$, and generates $m$ pseudo-observations $z_{1:m}$ \overset{\text{i.i.d.}}{\sim} \mathcal{G}_{\varepsilon, \delta}(F_{0})$ which they make available to $L$. Sampling $z_{1:m}$  can only depend on $F_0$ through samples $x_1,\dots,x_n$ - see Remark \ref{rem:sampling}. {\bf Alternative:} Let  $\mathcal{G}_{\varepsilon, \delta}(x_{1:n})$ to be the sampling distribution for the synthetic data.

%\vspace{-0.2cm}\item Let $z_{1:m}\sim$ $\mathcal{G}_{\varepsilon, \delta}(F_{0})$ denote $m$ pseudo-observations sampled using an $(\varepsilon, \delta)$-differentially private synthetic data generating mechanism (\SDGP), formulated using the real data $x_{1:n}$.
\item Let $f_{\theta}(x)$ denote a learner $L$'s model likelihood for $F_{0}(x)$, that is
parameterised by $\theta$ with prior $\tilde{\pi}(\theta)$, and marginal (predictive)
likelihood $p(x)=\int_{\theta}f_{\theta}(x)\tilde{\pi}(\theta)d\theta$.
\item $L$'s prior may already encompass some other set of real-data drawn from $F_0$ leading to $\tilde{\pi}(\theta) = \pi(\theta\mid x^L_{1:n_L})$, for $n_L \geq 0$ prior observations. %But for generality we suppress the dependence on $x^L_{1:n_L}$.
\end{itemize}

We adopt a decision theoretic framework \citep{Berger2013-qp}, in
assuming that $L$ wishes to take some optimal action $\hat{a}$
in a prediction or inference task; satisfying:
\begin{equation}
\hat{a}=\arg\max_{a\in\mathcal{A}}\int U(x,a)F_{0}(x)dx .\label{eq:decision_task}
\end{equation}
This is with respect to a user-specified utility-function $U(x,a)$ that evaluates actions
in the action space $\mathcal{A}$,
% , noting that this could be, for
% example, a space of probability functions. This makes precise $L$'s task and their motivation for seeking information from $K$.
and makes precise $L$'s desire to learn about $F_0$ in order to accurately identify $\hat{a}$.

%\jack{This is great as it defines why we want to learn about $F_0$. However, later we introduce this Divergence, which allows us to quantify how `useful' the of data from G when learning about $F_0$. Part of me likes the separation, the divergence is very abstract while `actions' sound mroe concrete. But we do slightly run the risk of confusing and repeating outselves. Thoughts?}

\textbf{Synthetic data generation mechanism.}
%The definition of $\mathcal{G}_{\varepsilon, \delta}$ allows for it to take two forms: 
In defining $\mathcal{G}_{\varepsilon, \delta}$, we believe it is important to differentiate between its two possible forms: 
\begin{enumerate}[leftmargin=*,noitemsep]
    \item $\mathcal{G}_{\varepsilon, \delta}(x_{1:n})=G_{\varepsilon, \delta}(z\mid x_{1:n})$, here $G$ is a privacy-preserving generative model fit on the real data such as the \PATEGAN \citep{jordon2018pate} or \DPGAN \citep{xie2018differentially}. These produce differentially private synthetic data by injecting heavy-tailed noise into gradient-based learning and/or training through partitioned collections, aggregations and subsets of the data. The \SDGP provides conditional independence between $z_{1:m}$ and $x_{1:m}$ and therefore no longer queries the real data after `fitting'. Alternative approaches in this class include fitting Bayesian Networks \citep[e.g.][]{Zhang2017-dt}. 
    \item $\mathcal{G}_{\varepsilon, \delta}=\int K_{\varepsilon, \delta}(x,dz) F_0(dx)$. A special case of this integral comprises the convolution of $F_0$ with noise distribution $H$ s.t  $\mathcal{G}_{\varepsilon, \delta}=F_{0}\star H_{\varepsilon, \delta}$. The sampling distribution is therefore not a function of the private data $x_{1:n}$. In this case, the number of samples that we can draw is limited to $m\leq n$ as drawing one data item requires using one sample of $K$'s data. Examples of this formulation include the Laplace mechanism \citep{dwork2014algorithmic} and transformation-based privatisation \citep{Aggarwal2004-vn}.
\end{enumerate}

% \begin{remark}
% In the case of Laplace mechanism the sampling distribution is not a function of the private data ${x_1,\dots,x_n}$. Instead  it corresponds to convolution of the data generating mechansims with the Laplace distribution. Sampling one $z$ in this case requires using one private data item.   be precise in the Laplace mechanism the sampling distribution corresponds to convolution with the underlying sampling $F_0$. Thus cannot be seen as a mapping from a fixed data set $x_1,\dots,x_n$ to a probability measure.
% {\bf Alternative: For Laplace mechanism, and similar mechanisms noising individual data items the sampling distribution cannot be obtained as a mapping from the dataset itself - as the sampling distribution corresponds to convolution of the data generating mechansims with the noise distribution.}
% \end{remark}

\textbf{{The fundamental problem of synthetic learning}} is that
%$L$ wants to learn about $F_0$ given $z_{1:m}\sim \mathcal{G}_{\varepsilon, \delta}$, and the prior $\pi$ as defined above, to make decisions based on the unknown $F_{0}$, and where 
$L$ wants to learn about $F_0$ but only has access to their prior $\tilde{\pi}(\theta)$ and to $z_{1:m}\sim \mathcal{G}_{\varepsilon, \delta}$, where:
\begin{itemize}[leftmargin=*,noitemsep]
\item $\mathcal{G}_{\varepsilon, \delta}\not\equiv F_{0}$. %for a \SDGP trained on infinite data, \jack{We need to be careful, given infinite data we no longer need to worry about the privacy of any one observation}
%$\mathcal{G}_{\varepsilon, \delta}(x)\equiv G(x\mid x_{1:n\to\infty});x_{i}\sim F_{0}(x)$. 
That is, the \SDGP $\mathcal{G}_{\varepsilon, \delta}(\cdot)$ is misspecified by design
\item $L$'s model, $p(x)$, is specified using beliefs about the
target $F_{0}(x)$ rather than $\mathcal{G}_{\varepsilon, \delta}(x)$, and the black-box nature of modern \SDGP's makes modelling them impossible. Therefore, the posterior predictive converges to a different distribution under real and synthetic data generating
processes such that $p(x\mid z_{1:m\to\infty})\not\equiv p(x\mid x_{1:n\to\infty})$
\end{itemize}

Learning from synthetic data is an intricate
example of learning under model misspecification, where the misspecification
is by $K$'s design. It is important, as shown below, that this is recognised in the updating of models. Fortunately we can adapt recent advances in Bayesian inference under model misspecification to help optimise learning with respect to $L$'s task.

\subsection{Bayesian Inference under model misspecification}

Bayesian inference under model misspecification has recently been
formalised \citep{walker2013bayesian,bissiri2016general} and represents
a growing area of research, see \citet{watson2016approximate,jewson2018principles,miller2018robust,lyddon2018generalized,grunwald2017inconsistency,knoblauch2019generalized}
to name but a few. Traditional Bayes rule updating in this context
can be seen as an approach that learns about the parameters of the
model that minimises the logarithmic score,
%as illustrated by Eq. \eqref{Equ:GeneralBayesRule}, 
or equivalently,
the Kullback-Leibler divergence (\KLD) of the model from the \DGP
of the data \citep{berk1966limiting,walker2013bayesian,bissiri2016general}, where $\KLD(g\,\|\,f) = \int \log \left(\sfrac{g}{f}\right)d\mathcal{G}_{\varepsilon, \delta}$.

%\jack{Is it obvious below that we have fixed \SDGP, $\mathcal{G}_{\varepsilon, \delta}$, estimated using fixed finite real data $x_{1:n}\sim F_0$, and then we consider sampling infinite $m$ from $\mathcal{G}_{\varepsilon, \delta}$? We do not want to consider infinite real data $x_{1:n}$ used to learn $\mathcal{G}_{\varepsilon, \delta}$ because in an infinite real sample privacy is no longer an issue}

As a result, if $L$ updates their model $f_{\theta}(x)$ using
synthetic data $z_{1:m}\sim \mathcal{G}_{\varepsilon, \delta}(x_{1:n})$, then as $m\to\infty$ they will be
learning about the limiting parameter that minimises the \KLD
to the \SDGP{}:
\begin{equation}
\theta_{\mathcal{G}_{\varepsilon, \delta}}^{\KLD}=\arg\min_{\theta\in\Theta}\operatorname{KLD}\left(g(\cdot)\,\|\,f_{\theta}(\cdot)\right),\label{Equ:KLDminimisation}
\end{equation}
and under regularity conditions the posterior distribution
concentrates around that point, $\pi(\theta\mid z_{1:m})\to\textbf{1}_{\theta_{\mathcal{G}_{\varepsilon, \delta}}^{\KLD}}$
as $m\to\infty$, where $g(\cdot)$ denotes the density function of
$\mathcal{G}_{\varepsilon,\delta}$. 

Moreover, the posterior will concentrate away from the model that is closest to $F_0$ in \KLD, corresponding to the limiting model that would be learnt given an infinite real sample $x_{1:\infty}$ from $F_0$: 
%the optimal model that is closest to $F_{0}(x)$ in \KLD, 
\begin{equation}
\theta_{\mathcal{G}_{\varepsilon, \delta}}^{\KLD} \not\equiv \theta_{F_{0}}^{\KLD}  =  \arg\min_{\theta\in\Theta}\operatorname{KLD}\left(f_{0}(\cdot)\,\|\,f_{\theta}(\cdot)\right)
\end{equation}

%which corresponds to the limiting model that would have been recovered through standard Bayesian inference on an infinite sample from $F_0$. 
This problem is exacerbated by the injection of noise by the \SDGP to ensure $(\varepsilon, \delta)$-\DP as this process produces data $z_{1:m}$ that is prone to outliers by design, with respect to $L$'s model $f_{\theta}(x)$.
%which if left unchecked can lead to significant difference between how $\theta_{\mathcal{G}_{\varepsilon,\delta}}^{\KLD}$ and $\theta_{F_{0}}^{\KLD}$ approximate $F_0$.
So, given that as we collect more synthetic data our inference is no longer minimising the \KLD towards $F_0$, we must carefully consider and investigate whether our inference is still `useful' for learning about $F_0$ at all.
%\jack{We note such a considertaion is not neccesary under the classical assumption...}

 % and in particular that $\KLD(F_0 \,\|\, f_{\theta^{\KLD}_{\mathcal{G}_{\varepsilon, \delta}}})$ can be large. 
 
%\jack{We define these theoretical limits so we know where our inference is heading}
 
 \subsection{The approximation to $F_0$}

%\jack{We need to introduce this metric before we start talking about robustness or the learning trajectory, because it is the metric $D$ that defines what it means for data from $\mathcal{G}_{\varepsilon, \delta}$ to be useful for learning about $F_0$}
%\jack{Could move the robustness to the next Section actually}

%In order to make concrete the way in which our inferences `approximate' $F_0$ we must introduce a notion of divergence between probability distributions. 
Before we proceed any further we must consider what it means for data from $\mathcal{G}_{\varepsilon,\delta}$ to be `useful' for learning about $F_0$. 
We can do so using the concepts of \textit{scoring rules} and statistical \textit{divergence}. %\jack{We note that such a concept is not necessary in normal Bayesian updating as the data is trivially useful about $F_0$} 
%A statistical divergence provides a measure of the `distance' between two probability distributions. scoring rules  \citep{dawid2007geometry,gneiting2007strictly}

\begin{definition}[Proper Scoring Rule]
The function $s: \mathcal{X}\times \mathcal{P}$ is a strictly proper scoring rule provided its difference function $D$ satisfies
\begin{align}\label{eq:proper}
    D(f_0 \,\|\, f) = \mathbb{E}_{x \sim f_0}\left[s(x, f(\cdot))\right] - \mathbb{E}_{x \sim f_0}\left[s(x, f_0(\cdot))\right]\nonumber
\end{align}
$D(f_1\,\|\,f_2) \geq 0$, $D(f\,\|\,f) = 0$ for all $f, f_1, f_2 \in \mathcal{P}(x)$
%\begin{enumerate}[leftmargin=*]
%    \vspace{-0.2cm}\item $D(f_1\,\|\,f_2) \geq 0$ for all $f_1, f_2 \in \mathcal{P}(x)$,
%    \vspace{-0.2cm}\item $D(f\,\|\,f) = 0$ for all $f \in \mathcal{P}(x)$;
%\end{enumerate} 
$$\mathcal{P}(x) := \left\{\int f(x) : f(x) \geq 0\, \forall x\in \mathcal{X}, \int_{\mathcal{X}}f(x)dx = 1\right\}$$
\end{definition}

%\begin{definition}[Statistical Divergence]
%A function $D: \mathcal{P}(x)\times\mathcal{P}(x)\rightarrow \mathbb{R}^{+}$ is a statistical divergence if
%\begin{enumerate}
%    \vspace{-0.2cm}\item $D(f_1\,\|\,f_2) \geq 0$ for all $f_1, f_2 \in \mathcal{P}(x)$
%    \vspace{-0.2cm}\item $D(f\,\|\,f) = 0$ for all $f \in \mathcal{P}(x)$
%\end{enumerate} 
%\vspace{-0.2cm}
%$$\mathcal{P}(x) := \left\{\int f(x) : f(x) \geq 0\, \forall x\in \mathcal{X}, \int_{\mathcal{X}}f(x)dx = 1\right\}$$
%\end{definition}
%
%Note that there is no requirement that a divergence satisfies the triangle inequality s.t one is not necessarily a metric. A subset of divergences emit the following representation involving proper scoring rules \citep{dawid2007geometry,gneiting2007strictly}. 
%
%\begin{definition}[Proper Scoring Rule]
%If there exists a function $s: \mathcal{X}\times \mathcal{P}$ such that
%\begin{align}
%    D(f_0 \,\|\, f) = \mathbb{E}_{x \sim f_0}\left[s(x, f(\cdot))\right] - \mathbb{E}_{x \sim f_0}\left[s(x, f_0(\cdot))\right]
%\end{align}
%In such a case the function $s(x, f)$ is called a proper scoring rule
%\end{definition}

%The function $D$ is called a \textit{divergence} and can be interpreted as follows: $D(f_0 \,\|\, f)$ is the additional expected penalty, $s(\cdot, \cdot)$, for thinking the data was distributed according to $f$ when it was instead distributed according to $f_0$ \citep{dawid2007geometry}. 
The function $D$ measures a distance between two probability distributions. 
$s(x, f)$ arises as the divergence is minimised when $f_0 = f$ \citep{gneiting2007strictly,dawid2007geometry}. A further advantage of this representation is that is allows for the minimisation of $D(f_0\,\|\,\cdot)$ using only samples from $f_0$,
\begin{align}
\,&\argmin_{f\in\mathcal{F}} D(f_0\,\|\,f) = \argmin_{f\in\mathcal{F}} \mathbb{E}_{x\sim f_0}[s(x, f(\cdot))]\nonumber\\
&\quad\quad\xleftarrow[n \to \infty]{} \frac{1}{n}\sum_{i=1}^n s(x_i, f(\cdot)), \quad x_i \sim f_0
\end{align}
Here, $f_0$ is the density of the \DGP $F_0$ generating real data; thus the approximating density $f$ becomes our predictive inferences made using synthetic $z_{1:m}\sim\mathcal{G}_{\varepsilon,\delta}$. Henceforth, we define any concepts of closeness (or `usefulness') in terms of a chosen divergence $D$ and associated scoring rule $s$. Given that inference using $\mathcal{G}_{\varepsilon,\delta}$ is no longer able to exactly capture $f_0$, $L$ can use this notion of closeness to define what aspects of $F_0$ they are most concerned with capturing. % through using data from $\mathcal{G}_{\varepsilon,\delta}$.  
The importance of this specification is illustrated in Section \ref{Sec:Experiments}.

% For what is to come we canonically we consider the logarithmic scoring rule and \KLD (introduced above Eq. \eqref{Equ:KLDminimisation}), owing to its intrinsic relationship with traditional Bayesian updating (Section \ref{Sec:Foundations}), but reinforce that the formulation holds in greater generality.

\section{Improved learning from the \SDGP}{\label{Sec:LearningSyntheticData}}

% Learning from synthetic data necessitates a unique consideration regarding how inference is learning a model. 
The classical assumptions underlying statistics are that minimising the \KLD is the optimal way to learn about the \DGP, and that more observations provide more information about this underlying \DGP; such logic does not necessarily apply here. $L$ wishes to learn about the private \DGP, $F_0$, but must rely on observations from the \SDGP $\mathcal{G}_{\varepsilon,\delta}$ to do so.
% which as mentioned earlier has a varying dependence on $F_0$ according to the privacy level required and the $(\varepsilon, \delta)$-\DP mechanism with which it is convolved. 
%
%\jack{The `Learner' is $L$ and the `Keeper' is $K$}
%Learning from synthetic data necessitates a unique consideration regarding how much data is desirable to use in learning a model. The classical assumption underlying statistics is that more observations provide more information about the \DGP and allow for a greater reduction in uncertainty; such logic does not necessarily apply here. $L$ wishes to learn about the private \DGP, $F_0$, but must rely on observations from the \SDGP $\mathcal{G}_{\varepsilon,\delta}$ which as mentioned earlier has a varying dependence on $F_0$ according to the privacy level required and the $(\varepsilon, \delta)$-\DP mechanism with which it is convolved. 
In this section we acknowledge this setting to propose a framework for improved learning from synthetic data. In so doing we pose the following question and detail our solutions in turn:
%
%Given metric $D$, learning using synthetic data poses two challenges to the Learner: 
%\begin{itemize}
%\vspace{-0.2cm}\item[1.] How to improve the robustness of the learning procedure to acknowledge the misspecifcation and outlier prone nature of $z_{1:m}\not\sim F_{0}(x)$? \jack{Here we want our limiting approximation to $G$ to be closer to $F_0$}
%\vspace{-0.2cm}\item[2.] Starting from the prior predictive, $p(x)$, detecting if or when during the learning procedure the posterior update is moving further away from $F_{0}(x)$. That is, when 
%\[
%\KLD\left[f_{0}(\cdot)\,\|\,p(x\mid z_{1:j+1})\right]>\KLD\left[f_{0}(\cdot)\,\|\,p(x\mid z_{1:j})\right]
%\]
%other notions of distance such as log-loss and Wasserstein are considered
%below.
%\end{itemize}
%\begin{itemize}
%\vspace{-0.2cm}\item[1.] Starting from the prior predictive, $p(x)$, when does learning using $z_{1:m}\not\sim F_0$ stop making you closer to $F_{0}(x)$. That is, when 
%\begin{equation}
%    D\left(f_{0}(\cdot)\,\|\,p(x\mid z_{1:j+1})\right) > D\left(f_{0}(\cdot)\,\|\,p(x\mid z_{1:j})\right)\nonumber
%\end{equation}
%\vspace{-0.2cm}\item[2.] How to improve the robustness of the learning procedure at approaimating $F_0$ to acknowledge the misspecifcation and outlier prone nature of $z_{1:m}\not\sim F_{0}(x)$?
%\end{itemize}
%We now deal with each challenge in turn.
%
Given the scoring criteria $D$, 
%learning using synthetic data poses the question 
is $\theta^{\KLD}_{\mathcal{G}_{\varepsilon, \delta}}$ the best the learner can do? 
\begin{itemize}[leftmargin=*,noitemsep]
\item[1.] Can the robustness of the \textit{learning procedure} be improved in approximating $F_0$ by acknowledging the misspecifcation and outlier prone nature of $z_{1:m}$? 
\item[2.] Starting from the prior predictive, $p(x)$, for a given learning method when does learning using $z\sim \mathcal{G}_{\varepsilon, \delta}$ stop improving inference for $F_{0}(x)$? That is, when 
%\vspace{-0.2cm}\item[2.] Starting from the prior predictive, $p(x)$, for a given learning method when does learning using $z_{1:m}$ stop improving inference for $F_{0}(x)$? That is, when 
%\begin{equation}
%    D\left(f_{0}(\cdot)\,\|\,p(\cdot\mid z_{1:j+1})\right) > D\left(f_{0}(\cdot)\,\|\,p(\cdot\mid z_{1:j})\right)\nonumber
%\end{equation}
\begin{equation}
    E_{z}\left[D\left(f_{0}(\cdot)\,\|\,p(\cdot\mid z_{1:j+1})\right)\right] > E_{z}\left[D\left(f_{0}(\cdot)\,\|\,p(\cdot\mid z_{1:j})\right)\right]\nonumber
\end{equation}
%\jack{Should we go all the way to $\infty$ or are we better to stop before}
\end{itemize}
%We now deal with each challenge in turn.

%\jack{It is important to define D before we start talking about the learning trajectory or robustness, because D defines what it means for data from G to be useful. I am still unsure of the order, should we do robustness before the learning trajectory or the other way around}

%\subsection{Statistical Divergences and Proper Scoring Rules}

%\jack{Data is usually trivially useful but now it is only useful if it helps us minimise sme score.}

%\jack{Under standard Bayesian inference there is a clear separation between the inference, which aims to learn about the process that generated the data, and the decision making which uses this inference as a best guess at the data generating process in order to minimise future expected losses. Such a separation relies on the assumption that given enough data the Learner is able to learn about the data generating process exactly, and will therefore correctly estimate expected losses/utilities and take the correct optimal decision according to this.}

%As we have introduced above, the Learner can no longer rely on the fact that their synthetic data, `collected' from $\mathcal{G}_{\varepsilon,\delta}$, is necessarily taking them towards $F_0$. In order to analyse this we need a way to quantify how useful the data from $\mathcal{G}_{\varepsilon,\delta}$ is at learning about $F_0$. One way to establish such a closeness is by using concepts from divergences and proper scoring rules. 

\subsection{General Bayesian Inference}

In order to address these issues we adopt a general Bayesian, minimum divergence paradigm for inference \citep{bissiri2016general, jewson2018principles} inspired by model misspecification, where $L$ can coherently update beliefs about their model parameter $\theta$ from prior $\pi(\theta)$ to posterior $\pi(\theta|z_{1:m})$ using:
\begin{align}
    \pi^{\ell}(\theta|z_{1:m}) \propto \frac{\tilde{\pi}(\theta)\exp\left(-\sum_{i=1}^m \ell(z_j, f_{\theta})\right)}{\int\tilde{\pi}(\theta)\exp\left(-\sum_{i=1}^m \ell(z_j, f_{\theta})\right)d\theta},\label{Equ:GeneralBayesRule}
\end{align}
where $\ell(z, f_{\theta})$ is the loss function used by $L$ for inference. The logarithmic score $\ell_0(z, f_{\theta}) = - \log f_{\theta}(z)$ recovers traditional Bayes rule updating. The predictive distribution associated with such a posterior and the model $f_{\theta}$ is:
\begin{equation}
p^{\ell}(x|z_{1:m}) = \int f_{\theta}(x)\pi^{\ell}(\theta|z_{1:m})d\theta \label{Equ:GeneralBayesianPredictive}   
\end{equation}

\subsection{Robust Bayes and dealing with outliers}{\label{Sec:RobustStatistics}}

%\jack{First we change the limit to make it more robust, then we acknowledge the point along the path to the limit which will benefit the inference, and clear if the limit is closer to the truth then the path should be too}

%As well as seeking to optimise a particular point along the learning trajectory itself, we can also think about adapting the particular inference methods, acknowledging the inherently misspecified nature of the problem in order to `optimise' the learning trajectory itself. 

In the absence of the ability to correctly model the \SDGP, robust
statistics \citep[see e.g.][]{berger1994overview} provide an alternative
option to guard against artefacts of the generated synthetic data.
%We characterise robust statistics from a Bayesian standpoint into two categories.
We can gain increased robustness in our learning procedure to data $z_{1:m}$ by changing the loss function, $\ell(z, f_{\theta})$ used for inference in Eq. \eqref{Equ:GeneralBayesRule}. Here we consider two alternative loss functions to the standard logarithmic score underpinning standard Bayesian statistics,
\begin{align}
    \ell_w(z, f_{\theta}) :&= -w\log f_{\theta}(z)\label{Equ:wKLDloss}\\
    \ell^{(\beta)}(z,f_{\theta}) :& =\frac{1}{\beta+1}\int f_{\theta}(y)^{\beta+1}dy-\frac{1}{\beta}f_{\theta}(z)^{\beta}.\label{Equ:betaDloss}
\end{align}
Loss function $\ell_w(z, f_{\theta})$ introduces a learning parameter $w>0$ into the Bayesian
update \citep[e.g.][]{lyddon2018generalized,grunwald2017inconsistency,miller2018robust,holmes2017assigning}. Down-weighting, $w<1$ will generally produce a less confident posterior than in the case of traditional Bayes' rule, with a greater dependence on the prior. Conversely,
$w>1$ will have the opposite effect. The value of $w$ can have ramifications for inference and prediction \citep{rossell2018tractable,grunwald2017inconsistency}. However, we note that as the sample size grows using the weighted likelihood posterior will still learn about $\theta_{G}^{\ast}$ if $w$ is fixed. Choosing $w=\sfrac{s}{m}$ instead averages the log-likelihood. In this case $s$ can be seen as a notion of effective sample size.

Alternatively, minimising $\ell^{(\beta)}(x,f(\cdot))$ in expectation over the \DGP is equivalent to minimising the $\beta$-divergence (\BD)  \citep{basu1998robust}. Therefore, analogously to the \KLD and the log-score, using $\ell^{(\beta)}(x,f(\cdot))$ \citep{bissiri2016general,jewson2018principles,ghosh2016robust} produces a Bayesian update targeting:
\begin{align}
\theta_{\mathcal{G}_{\varepsilon, \delta}}^{\BD}: & =\arg\min_{\theta\in\Theta}\BD\left(g(\cdot)\,\|\,f_{\theta}\right).
\end{align}
As $\beta\rightarrow0$, then $\BD\rightarrow\KLD$,
but as $\beta$ increases it provides increased robustness through
skepticism of new observations relative to the prior. 
We demonstrate the robustness properties of the \BD in some simple scenarios in the Supplementary material and refer the reader to e.g. \cite{knoblauch2019generalized, knoblauch2018doubly} for further examples. 
%This increased robustness through minimising the \BD over the \KLD can be seen in Fig.\ref{Fig:OutlierInfluence}, where one outlying observation can be seen to move \KLD-Bayes inference away from the prior predictive, while the \BD-Bayes is less sensitive to such an observation and still learn from other observations in agreement with the prior.
%
%\begin{figure}[h]
%\centering{}\includegraphics[clip,width=0.7\columnwidth,trim= {0.0cm 0.0cm 0.0cm 0.0cm}]{figures/NIG_posterior_predictives_inlier-1}\\
%\includegraphics[clip,width=0.7\columnwidth,trim= {0.0cm 0.0cm 0.0cm 0.0cm}]{figures/NIG_posterior_predictives_outlier-1}
%trim={<left> <lower> <right> <upper>}
%\caption{ Influence of Outliers: NIG Prior predictive ({\color{black}{\textbf{black}}})
%and posterior predictives using a Gaussian model under the \KLD-Bayes
%({\color{red}{\textbf{red}}}) and \BD-Bayes ({\color{blue}{\textbf{blue}}})
%after an inlying (left) and outlying (right) observation ({\color{gray}{\textbf{grey}}}) }
%\label{Fig:OutlierInfluence} 
%\end{figure}
%
We note there are many possible divergences providing greater robustness properties than the \KLD, e.g. Wasserstein or Stein discrepancy \citep{barp2019minimum}, but for our exposition we focus on the \BD for its convenience and simplicity.

A key difference between the two robust loss functions considered above is that while $\ell_w(z, f_{\theta})$ down-weights the log-likelihood of each observation equally, $\ell^{(\beta)}(x,f(\cdot))$ does so adaptively, based on how likely the new observation is under the current inference \citep{cichocki2011generalized}. It is this adaptive down-weighting that allows the \BD to target a different limiting parameter to $\theta^{\KLD}_{\mathcal{G}_{\varepsilon, \delta}}$.
This, in particular, allows the \BD to be robust to outliers and/or heavy tailed contaminations. As a result, we believe that 
$D\left(F_0\,\|\,f_{\theta_{\mathcal{G}_{\varepsilon, \delta}}^{\BD}}\right) < D\left(F_0\,\|\,f_{\theta_{\mathcal{G}_{\varepsilon, \delta}}^{\KLD}}\right)$ across a wide class of \SDGP{}s. That is to say that the \BD minimising approximation to $\mathcal{G}_{\varepsilon, \delta}$ 
%, $f_{\theta_\mathcal{G}_{\varepsilon, \delta}^{\beta}}$ 
is a better approximation of $F_0$ than the KLD minimising approximation.
%to $f_{\theta_{\mathcal{G}_{\varepsilon, \delta}^{KLD}}$.

A strength of the \BD is that, unlike standard robust methods using heavier tailed models or losses \citep{berger1994overview,huber1981robust,beaton1974fitting}, $\ell^{(\beta)}(x,f(\cdot))$ does not change the model used for inference. In the absence of any specific knowledge about the \SDGP, updating using the \BD maintains the model $L$ \textit{would} have used to estimate $F_0$, but updates its parameters robustly. This also has advantages in the data combination scenario where $L$ is combining inferences from their own private data $x^L_{1:n_L}$ with synthetic data $z_{1:m}$. They can maintain the same model for both datasets, with the same model parameters, yet update robustly about $z_{1:m}$ whilst still using the log-score for $x^L_{1:n_L}$ (i.e. to produce $\tilde{\pi}(\theta)$).
%\jack{Give the betaD its own subsection. We need to stress that in absence of knowledge of the Synth-DGP that you don't want to change the model (as you don't know what the S-DGP is!) but you want robustness so keep your model for $F_{0}(x)$ but update robustly (rather than change your model to a robust one)} 

%\jack{Idea of both methods down-weighting the data (changing the effective sampel size), wKLD does i somewhat vanillaly, betaD is more adaptive}

% We stress again that the model agnostic nature of the \BD means it
% can be applied in any modelling scenario (i.e. for both regression
% and classification) without changing the model  . This provides a useful
% platform on which to compare this approach with traditional ones,
% and allows one to retain their initial model (which they believe represents
% $F_{0}$ on some level) but learn its parameters in a robust fashion.
% This feature will become particularly useful in what is to come.

\subsection{The Learning Trajectory}\label{sec:learningTraj}

%\harry{Good way to think about this is the beta-divergence carrying momentum from the real data's trajectory towards $F_0$ carrying it on a closer path to it on the way to $\mathcal{G}_{\varepsilon,\delta}$}

%As well as allowing us to consider the desirabiliy of the limiting approximation to $F_0$ under different learning methods, the concept of closeness provided by $D$, between the inferences made using data collected from $\mathcal{G}_{\varepsilon,\delta}$ and the true \DGP $F_0$ 
Moreover, the concept of closeness provided by $D$ allows us to consider how $L$'s approximation to $F_0$ changes as more data is collected from the \SDGP.
%we can consider how this measure of closeness 
%changes as we use more data from the \SDGP. 
%We have established previously that as the learner draws more observations from $\mathcal{G}_{\varepsilon,\delta}$, they will learn about $\theta_{\mathcal{G}_{\varepsilon,\delta}}^{\KLD}$ the model minimising the \KLD to $\mathcal{G}_{\varepsilon,\delta}$. The quality of $f_{\theta_{\mathcal{G}_{\varepsilon,\delta}}^{\KLD}}$ at approximating $F_0$ can then be evaluated using $K_{\infty} = D(f_0 \,\|\, f_{\theta_{\mathcal{G}_{\varepsilon,\delta}}^{\KLD}})$. 
Firstly, we provide a trivial theorem that says using more data and approaching the limit $\theta^{\KLD}_{\mathcal{G}_{\varepsilon, \delta}}$ is not necessarily the optimal target to learn about according to criteria $D$. 

\begin{proposition}[Suboptimality of learning \SDGP]
For \SDGP $\mathcal{G}_{\varepsilon,\delta}$, model $f_{\theta}(\cdot)$, and divergence $D$, there exists prior $\tilde{\pi}(\theta)$, private \DGP $F_0$ and $0\leq m <\infty$ such that 
%\begin{equation}
%    D\left(F_0\,\|\,p(\cdot|z_{1:m})\right) \leq D(F_0\,\|\,f_{\theta^{\KLD}_{\mathcal{G}_{\varepsilon,\delta}}})\label{Equ:SuboptimalittG0D}
%\end{equation}
\begin{equation}
    \mathbb{E}_{z}\left[D\left(F_0\,\|\,p(\cdot|z_{1:m})\right)\right] \leq D(F_0\,\|\,f_{\theta^{\KLD}_{\mathcal{G}_{\varepsilon,\delta}}})\label{Equ:SuboptimalittG0D}
\end{equation}
where $\theta^{\KLD}_G := \argmin_{\theta\in\Theta} \KLD(\mathcal{G}_{\varepsilon,\delta}\,\|\,f_{\theta})$ and $p(x|z_{1:m})$ is the Bayesian posterior predictive distribution (using $\ell_0$) based on (synthetic) data $z_{1:m}$, see Eq. \eqref{Equ:GeneralBayesianPredictive}.
\label{Thm:SuboptimalittG0D}
\end{proposition}

% We know that under regularity conditions as $m\rightarrow\infty$ Bayes rule will concentrate about the parameter $\theta^{\ast}_{\KLD} := \argmin_{\theta\in\Theta} \KLD(\mathcal{G}_{\varepsilon,\delta}\,\|\,f(\cdot;\theta))$ \citep{berk1966limiting}, as a result the takeaway here is that given an infinite sample from an \SDGP $\mathcal{G}_{\varepsilon,\delta}\neq F_0$ is is not necessarily optimal to use all of the data available for learning, contrary to the logic of standard statistical analyses. 
The proof of Proposition \ref{Thm:SuboptimalittG0D} involves a simple counter example in which the prior is a better approximation to $F_0$ according to divergence $D$ than $\mathcal{G}_{\varepsilon,\delta}$. While trivial, this could reasonably occur if $L$ has strong, well-calibrated expert belief judgements, or if they have a considerable amount of their own data before incorporating synthetic data. Further, we argue next that by considering the path, in the number of synthetic observations $m$, between the prior and the \SDGP $\mathcal{G}_{\varepsilon,\delta}$ it is possible to get even `closer' to $F_0$. We call such a path the learning trajectory. 

Changing the divergence used for inference as suggested in Section \ref{Sec:RobustStatistics} changes these trajectories by changing their limiting parameter. However, Prop. \ref{Thm:SuboptimalittG0D}, which considered learning minimising the \KLD, can equally be shown for learning minimising the \BD (see Supplementary material). In the following Sections we talk generally about optimising such a trajectory for a given learning method, before focusing on the comparing methods and trajectories in Section \ref{Sec:Experiments}. %As a result consider $p^{\ell}(x|z_{1:m})$ as the Bayesian predictive distribution for the given learning methods/loss function, see Eq. \eqref{Equ:GeneralBayesianPredictive}.

%We draw an analogy here with the bias-variance trade-off associated with statistical optimality criteria such a mean-squared-error. Often practitioners are willing to accept some bias to their analysi in order to reduce its variance. We view the addition of synthetic data . While adding observations from $\mathcal{G}_{\varepsilon,\delta}$ can be seen to `bias' the analysis away from $F_0$. Additional observations in the same direction as $F_0$ can reduce the variance. Finding the optimal point on this trajectory can be seen to optimise this trade-off.
%\jack{We can actually do the maths for a MSE example}

%\subsection{The Learning Trajectory}

\subsection{Optimising the Learning Trajectory}

%\jack{Here we are using (sacrificing) some of the data from $F_0$ in order to guide the inference using synthetic data}

%\jack{So we cannot advocate infinite communication. But can we for some canonical models place enough structure on the communication process to allow this to be possible, i.e. fixed predictive models, logistic regression, random forest, neural networks, where you specify your learning method, KLD, betaD, wKLD and then have some fixed prior hyper-parameters. Yes they can just do it all in house, fit the model on part of the synthetic data, and then use the rest to test on.}

%\jack{Is the AUC a proper scoring rule, no but I suppose the breir score is, and it's the test set Breir score I have to privatise rather than the individual one so should be fine.}

The insights from the previous section raise two questions for a learner $L$ using synthetic data:
\begin{enumerate}[leftmargin=*,noitemsep]
    \item Is synthetic data able to improve $L$'s inferences about $F_0$ according to divergence $D$? If so,
    \item What is the optimal quantity to use in getting the learner closest to $F_0$ according to $D$?
\end{enumerate}
\vspace{-0.2cm}
%In this section we seek to use a small amount of data from $F_0$ to answer these questions and guide our inference towards $F_0$.
%Given a sample $z_{1:M}\sim \mathcal{G}_{\varepsilon, \delta}$ $M\leq \infty$, 
Both questions can be solved by the estimation of 
%\begin{strip}
\begin{equation} \label{eqn:ltraj}
%\begin{aligned}
%m^{\ast} := \argmin_{{0\leq m \leq M}} \left[ D(f_0 \,\|\, p(\cdot|z_{1:m})\right],
m^{\ast} := \argmin_{{0\leq m \leq M}} \mathbb{E}_{z}\left[ D(f_0 \,\|\, p^{\ell}(\cdot|z_{1:m})\right],
% \end{aligned}
\end{equation}
%\chris{should this be an expectation over $z_{1:m} \sim G$??}
%\end{strip}

However, clearly the learner never has access to the data generating density. Instead we take advantage of the representation of proper scoring rules and propose using a `test set' $x^{\prime}_{1:N}\sim F_0$ to estimate
\begin{align}
  \hat{m} :&= \argmin_{0\leq m \leq M} \frac{1}{N}\frac{1}{B}\sum_{b=1}^{B}\sum_{j=1}^N s(x^{\prime}_j, p^{\ell}(\cdot|z^{(b)}_{1:m}))\label{Equ:tilde_mSDGP}\\
  &\textrm{with } \{z_{1:m}^{(b)}\}_{b_{1:B}}\sim \mathcal{G}_{\varepsilon, \delta}.\nonumber
\end{align}
As such we use a small amount of data from $F_0$ to guide the synthetic data inference towards $F_0$. We consider doing so in a tailored fashion for $L$'s specific inference problem, or put the onus on $K$ to evaluate the general ability of their \SDGP to capture $F_0$.
%Next we consider calculating this objective.
%\sebastian{we should. cconsider three cases  prior is futh8er away then using infinte amount of syn data, prior is closer but we can sfill benfit from syn data we cannot benefit from syn data}

%\jack{The question here is do we optimise the learning trajectory for one fixed, ordered sample $z_{1:m}$, in which case we do not take expectations, or do we optimise the learning trajectory for a fixed \SDGP, in which case we want to take expectation over $z_{1:m}\sim G$}
%\jack{If we do so for a fixed dataset then I think we just throw away $z_{m+1, M}$, i.e. don't do any averaging when we learn}

\subsubsection{Optimising for $L$'s inference}

%In the presence of an independent testing data set $x^{\prime}_{1:N} \sim F_0$ we can take advantage of the proper scoring rule and estimate
%\begin{enumerate}
%    \item the expectation with respect to $f_0$ for calculating the scoring rule see \eqref{eqn:proper}
%    \item the expectation with respect to synthetic data using samppling of the empirical distribution of the synthethic data. 
%\end{enumerate}$m^{\ast}$ using
%\begin{align}
%  \hat{m} :&= \argmin_{0\leq m \leq M} \frac{1}{N\cdot B}\sum_{j=1}^{N}\sum_{i=1}^B s(x^{\prime}_j, p^{\ell}(\cdot|z^{(b)}_{1:m}))\label{Equ:hat_m_oneDataset}
%\end{align}

Consider two potential sources of an independent test set $x^{\prime}_{1:N}\sim F_0$ allowing the learner $L$ to calculate the $\hat{m}$ associated with the specific learning trajectory of their problem.
The first option is for $L$ to sacrifice some of their own data $x^L_{1:n_L}$ when constructing their prior. The second requires that $K$ hold, $x^{\prime}_{1:N}$ out when it trains the \SDGP, which can then be queried by $L$ in order to estimate $\hat{m}$. Clearly $K$ is not able to share the observations with $L$ as this would violate the \DP guarantee. Instead a secure protocol for two-way communication between $L$'s model and $K$'s test set must be established; promising directions include \citep{Cormode2019-xn,De_Montjoye2018-wz} and a practical use case \citep{UK_Health_Data_Research_Alliance2020-xu}. 

\begin{remark}
We may consider data-dependent $m$, for a concrete stream of data $z_{1:m}$,
\begin{align}
  \hat{m} :&= \argmin_{0\leq m \leq M} \frac{1}{N}\sum_{j=1}^{N} s(x^{\prime}_j, p^{\ell}(\cdot|z^{}_{1:m})).\label{Equ:hat_m_oneDataset}
\end{align}
%This considers when to stop for a concrete dataset comprising all the data available to $L$. 
This introduces an undesirable dependency on the ordering of the data but can be mitigated by averaging different realisations of the synthetic data, which in turn can be shown to improve any convex proper scoring rule, see Prop. \ref{thm:avg}. See the Supplementary material for a proof of this Remark.
\end{remark}

\subsubsection{A Broader Study}
%\sebastian{that is why I find the bootstrap interesting if $K$ releases mechanism for syn data}
When the previous, problem and data specific methods are not available we have to fall back on a broader study. 
Here we recommend that alongside releasing synthetic data, $K$ optimises the learning trajectory themselves, under some default model, loss and prior setting by repeatedly partitioning $x_{1:n}$ into test and training sets. For example, when releasing classification data, $K$ could release an $\hat{m}$ associated with logistic regression and BART, for the log-score, under some vaguely informative priors, providing learners an idea of what to expect. While this is less tailored to any specific inference problem it still allows $K$ to communicate a broad measure of the quality of its synthetic data for learning about $F_0$. 

\subsection{Posthoc improvement through averaging }

If more synthetic data is available (e.g. sampling $z_{1:m}, m\to\infty$ from a \GAN), we can average the posterior predictive distribution across different realisations ensuring we do not waste synthetic data when $\hat{m}$ is less than the maximum available. Jensen's inequality allows us to improve the performance of the predictive distribution if we consider convex proper scoring rules such as the logarithmic score: 
%\begin{theorem}
%\label{thm:avg}The expected scoring rule decreases when averaging
%over different realisations of the (modified) posterior predictive depending
%on different synthetic data sets
%
%\begin{align*}
%\mathbb{E}_{z}D\left(F_{0}\,\|\,\frac{1}{B}\sum_{b=1}^{B}\tilde{p}\left(x|z_{1:m^{(b)}\right)\right) & \ge\mathbb{E}_{z}\frac{1}{B}\sum_{b=1}^{B}D\left(F_{0}\,\|\,\tilde{p}\left(x|z_{1:\hat{m}_{b}}^{(b)}\right)\right)\\
% & =\mathbb{E}_{z}D\left(F_{0}\,\|\,\tilde{p}\left(x|z_{1:\hat{m}_{b}}^{(b)}\right)\right).
%\end{align*}
%\end{theorem}
%
%\begin{proof}
%We use the Jensen inequality and the identical distribution of $\tilde{p}\left(x|z_{1:\hat{m}_{b}}^{(b)}\right)$
%across varying $b$ for the equality. 
%\end{proof}

\begin{proposition}[Predictive Averaging]
\label{thm:avg}Given divergence $D$ with convex scoring rule, averaging
over different realisations of the posterior predictive depending on different synthetic data sets improves inference by Jensen's inequality:
\begin{align*}
\mathbb{E}_{z}D\left(F_{0}\,\|\,\frac{1}{B}\sum_{b=1}^{B}\tilde{p}\left(x|z_{1:m}^{(b)}\right)\right)\leq\\\quad\mathbb{E}_{z}\frac{1}{B}\sum_{b=1}^{B}D\left(F_{0}\,\|\,\tilde{p}\left(x|z_{1:m}^{(b)}\right)\right)=\\\quad\mathbb{E}_{z}D\left(F_{0}\,\|\,\tilde{p}\left(x|z_{1:m}^{(b)}\right)\right).
\end{align*}
\end{proposition}
The significance of this is that more synthetic data can always be used to improve the predictive distribution, but not by naïvely using all of it to learn at once.

\section{Experimental Setup and Results}{\label{Sec:Experiments}}

In order to investigate the concepts and methodologies outlined above, we consider two experiment types:
\begin{enumerate}[leftmargin=*,noitemsep]
    \item Learning the location and variance of a Gaussian distribution.
    \item Using Bayesian logistic regression for binary classification on a selection of real-world datasets.
\end{enumerate}
In these contexts we investigate the learning trajectory of classical Bayesian updating alongside the robust adjustments discussed in Section \ref{Sec:RobustStatistics}. In order to draw comparisons between these methods, we study the trajectories' dependence on values spanning a grid of data quantities $n_L$ and $m$, robustness parameters $w$ and $\beta$, prior values, and the parameters of chosen \DP mechanisms (Full experiment specifications are included in the Supplementary material). 
The varying amounts of unprivatised data available to $L$ were used to construct increasingly informative priors $\tilde{\theta} = \pi(\theta\mid x^L_{1:n_L})$, using standard Bayesian updating as no robustness is required when learning using data from $F_0$.
%We use standard Bayesian updating to learn from the varying amounts of real data alongside an uninformative prior to give a varying prior $\pi(\theta\mid x^L_{1:n_L})$ due to our assumption that the original model is correct for data coming from $F_0$. 
%Learning trajectories can be calculated for models spanning the full grid of possibilities defined above by utilising an unseen dataset $x_{1:n}$ (this could be the keeper $K$'s data or some subset of $L$'s data not used in training).
Learning trajectories are then estimated utilising an unseen dataset $x^{\prime}_{1:N}$ (mimicking that defined by either $K$'s data or some subset of $L$'s data not used in training).

%When undertaking experiments we seek to investigate the value of $\hat{m}$ as defined in Eq. \eqref{Equ:hat_m_oneDataset}. Carrying out repeat experiments and observing expected performance leads to us repeating the computation of $\hat{m}$ across multiple realisations of the synthetic data. In so doing we still seek to mimic the procedure for optimising $\hat{m}$ for a specific synthetic data sample $z_{1:m}$,
%\begin{align} \label{eq:emin}\
%  \mathbb{E}_z \min_{0\leq m \leq M} \frac{1}{N}\sum_{j=1}^{N} s(x^{\prime}_j, p(\cdot|z_{1:m})).
%\end{align}
%This should not be confused with $\tilde{m}$ defined in Eq. \eqref{Equ:tilde_mSDGP}, which rather than focusing on finding the optimal $m$ for any one dataset, provided a value of $m$ that was best for an average data set from $\mathcal{G}_{\varepsilon, \delta}$, i.e. valid for the \SDGP on average rather than any specific realisation.

To this end we use optimised MCMC sampling schemes \citep[e.g.][]{hoffman2014no} to sample from the classical and adjusted posteriors in each case and draw comparisons across the grid laid out above, repeating experiments to mitigate %to be relatively sure of our observations in spite of 
any sources of noise. This results in an extensive computational task, made feasible through a mix of Julia's Turing PPL \citep{ge2018t}, MLJ \citep{Blaom2020-le} and Stan \citep{carpenter2017stan}.

The majority of the experiments are carried out with $\varepsilon = 6$, which is seen to be a realistic value respective of practical applications \citep{lee2011much, erlingsson2014rappor, tang2017privacy, applepriv} and upon observation of the relationship between privacy and misspecification shown in the figure included in the Supplementary material.

\subsection{Simulated Gaussian Models}

We first introduce a simple but illustrative simulated example in which we infer the parameters of a Gaussian model $f_{\theta} = \mathcal{N}(\mu, \sigma ^ 2)$  where $\theta = (\mu, \sigma^2)$. We place conjugate priors on $\theta$ with $\sigma ^ 2 \sim \text{InverseGamma}(\alpha_p, \beta_p)$ and $\mu \sim \mathcal{N}(\mu_p, \sigma_p * \sigma)$ respectively. We consider $x_{1:n}$ drawn from \DGP $F_0 = \mathcal{N}(0, 1^2)$ and adopt the Laplace mechanism  \citep{dwork2014algorithmic} to define our \SDGP. This perturbs samples drawn from the \DGP with noise drawn from the Laplace distribution of scale $\lambda$, calibrated via the sensitivity $\mathcal{S}$ of the \DGP in order to provide $(\varepsilon, 0)$-\DP per the Laplace mechanism's definition with $\varepsilon = \sfrac{\mathcal{S}}{\lambda}$. To achieve finiteness of $\mathcal{S}$ in this case, we adjust our model to be that of a truncated Gaussian; restricting its range to $\pm 3\sigma$ to allow for meaningful $\varepsilon$'s to be calculated under the Laplace mechanism. 

We then compare and evaluate the empirical performances of the models defined below (formulations are given explicitly in the Supplementary material):

\begin{enumerate}[leftmargin=*,noitemsep]
    \item The standard likelihood formulated with a $w$ parameter as in Eq. \eqref{Equ:wKLDloss}.% to allow for robustness from reweighting.
    \item The posterior under the \BD loss as in Eq. \eqref{Equ:betaDloss}.
    \item The `Noise-Aware' likelihood where the \SDGP can be tractably modelled using the Normal-Laplace convolution \citep{reed2006normal, amini2017letter}.
\end{enumerate}

\subsubsection{Results and Discussion}

\begin{figure*}
    \centering
    \includegraphics[width=\textwidth]{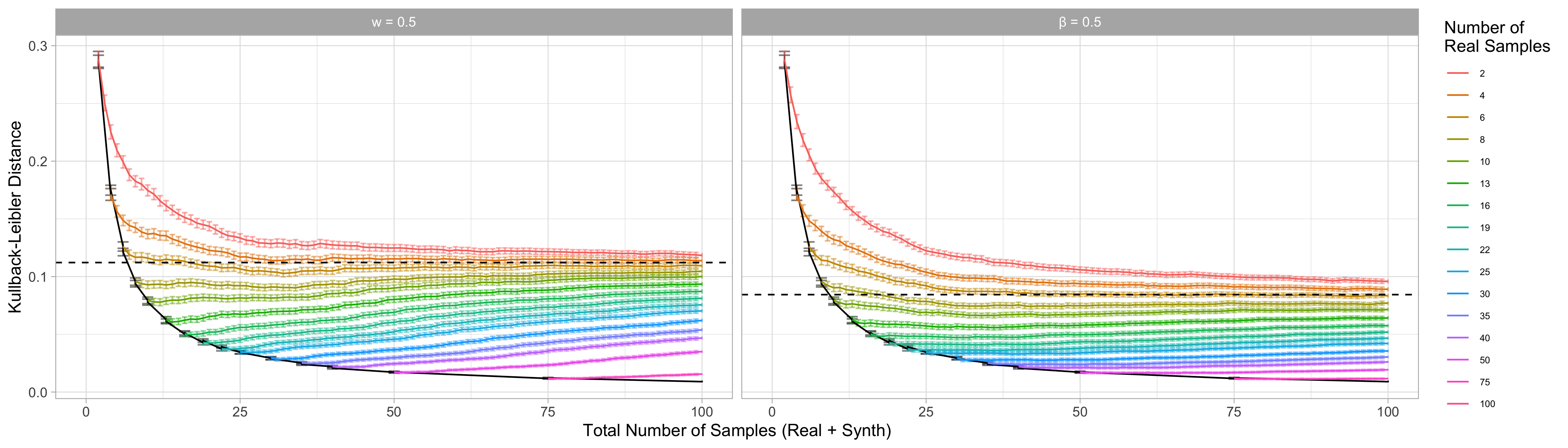}
    \caption{Shows how the \KLD to $F_0$ changes as we add more synthetic data, starting with increasing amount of real data. This demonstrates the effectiveness of the best $\BD$ configuration found compared to the the closest alternative traditional model in terms of performance with down-weighting $w = 0.5$; the black dashed line represents $\KLD(F_0\,\|\,f_{\theta^{\ast}})$ for $\theta^{\ast} = \theta^{\KLD}_{\mathcal{G}_{\varepsilon, \delta}}$ on the left and $\theta^{\ast} = \theta^{\BD}_{\mathcal{G}_{\varepsilon, \delta}}$ on the right, representing the approximation to $F_0$ given an infinite sample from $\mathcal{G}_{\varepsilon, \delta}$ under the two learning methods. }
\label{Fig:GaussBranched} 
\end{figure*}

\begin{figure*}
    \centering
    \includegraphics[width=\textwidth]{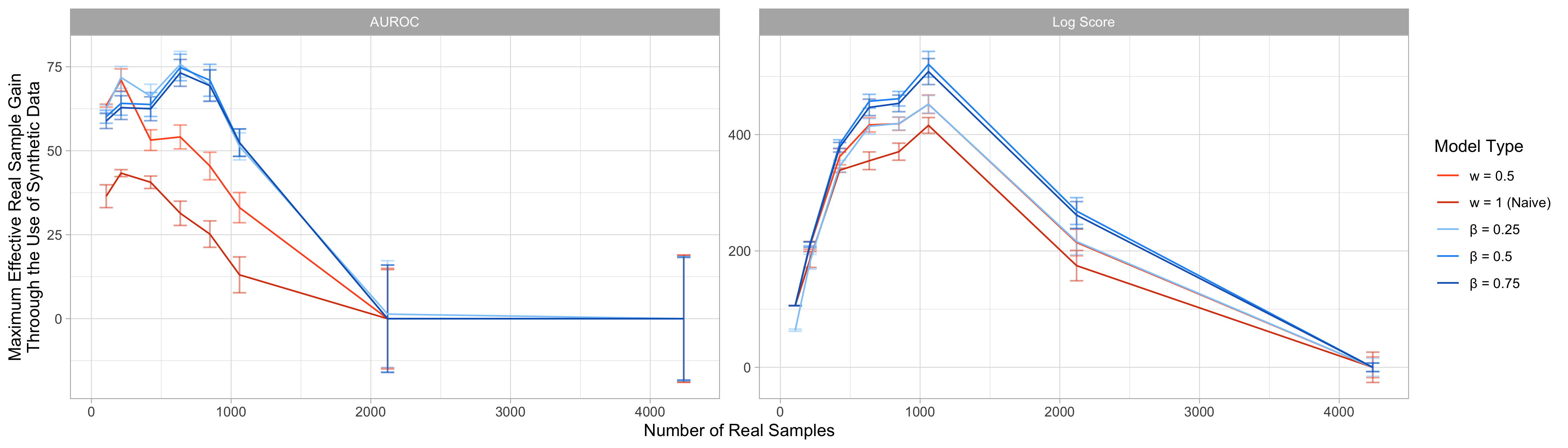}%
    \caption{ Shows the effective number of real samples gained through optimal $\hat{m}$ synthetic observations alongside varying amounts of real data usage w.r.t the AUROC and log-score performance criteria. These are calculated and presented here via bootstrapped averaging under a logistic regression model learnt on the Framingham dataset. The amount of effective real samples is significantly affected by the learning task's focused criteria. }%
\label{fig:Neff}%
\end{figure*}

\begin{figure*}
    \centering
    \subfloat[][\centering{Simulated Gaussian, real $n = 10$}]{
        \includegraphics[width=0.45\textwidth]{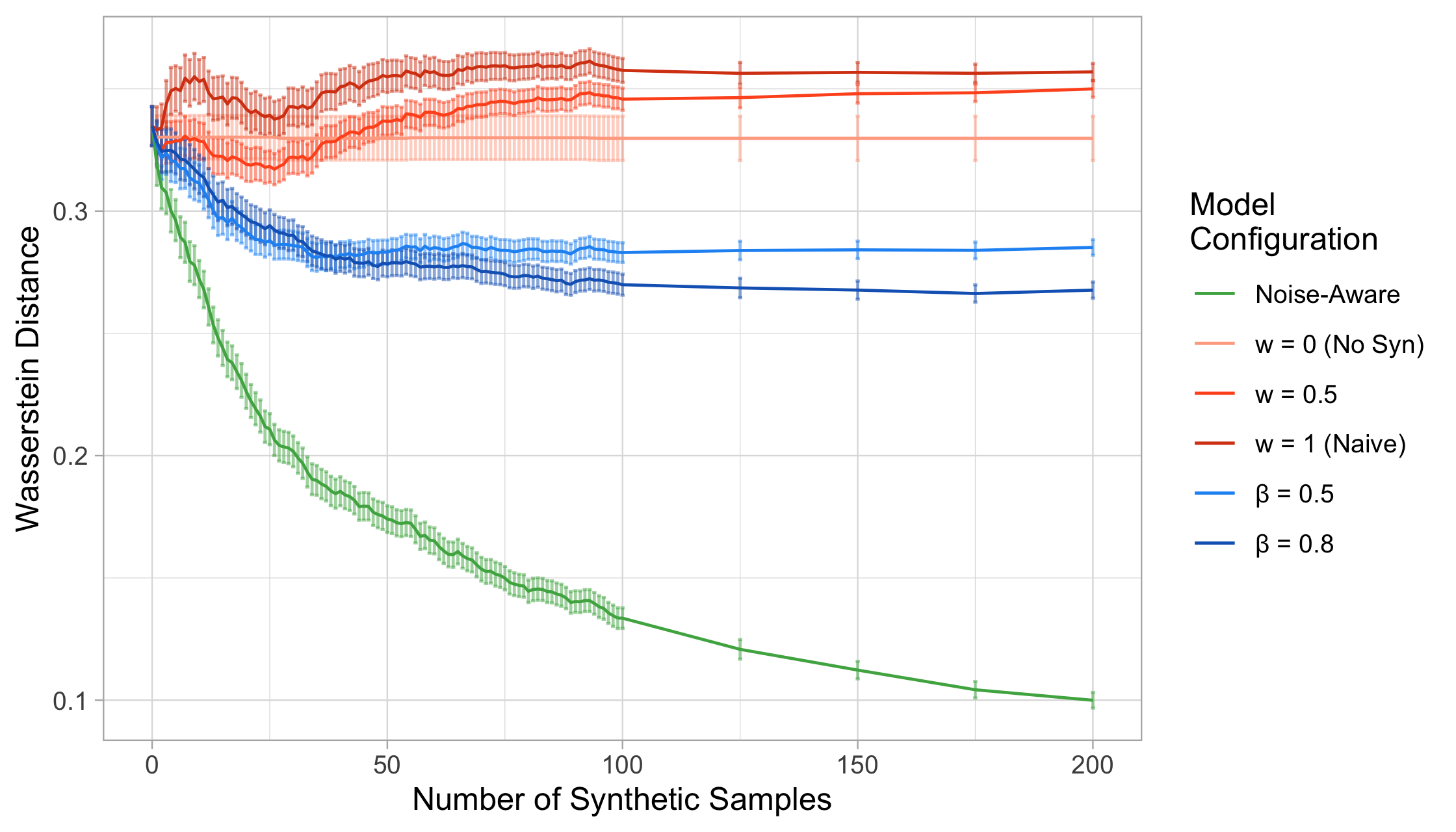}
    }%
    \subfloat[][\centering{UCI Heart Dataset, real $n = 30$ $(10\%)$}]{
        \includegraphics[width=0.45\textwidth]{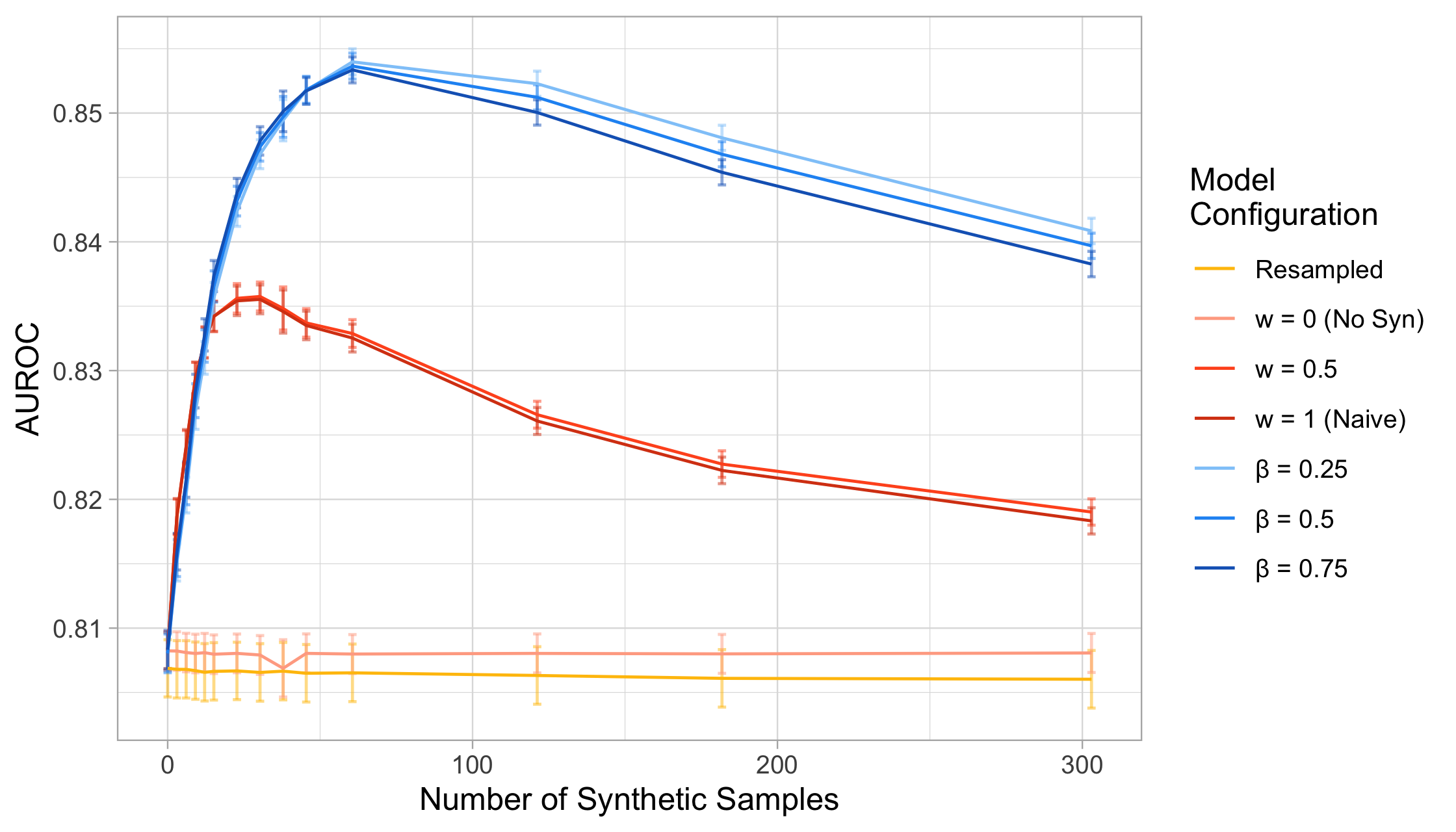}
    }%
    \caption{Given a fixed real amount of data, we can compare model performances directly by focusing on one of the `branches' in the class of diagrams shown in Figures \ref{Fig:GaussBranched} \& \ref{Fig:EpsComparison}, to see that the \BD's performance falls between that of the noise-aware model and the other models, exhibiting robust and desirable behaviour across a range of $\beta$. Naïve and reweighting-based approaches fail to gain significantly over not using synthetic data (shown by $w=0$'s flat trajectory); the resampled model in the logistic case can also be seen to perform very poorly in comparison to models that leverage the synthetic data. }
\label{Fig:ModelComparison} 
\end{figure*}

We observe that three different categories of learning trajectory occur across the models, these are illustrated in the `branching' plots in Figure \ref{Fig:GaussBranched}:
\begin{enumerate}[leftmargin=*,noitemsep]
    \item The prior $\tilde{\pi}$ is sufficiently inaccurate or uninformative (in this case due to low $n_L$) such that the synthetic data continues to be useful across the range of $m$ we consider. As a result the learning trajectory is a monotonically decreasing curve in the criteria of interest.
    \item A turning point is observed; synthetic data initially brings us closer to $F_0$ before further synthetic data moves the inference away. We see that in the majority of cases these trajectories lie under the limiting \KLD and \BD approximations to $\mathcal{G}_{\varepsilon, \delta}$ demonstrating the efficacy of `optimising the learning trajectory'.
    %that represents an initial benefit from the use of synthetic data that is then overcome by the distance in the relevant space between $\mathcal{G}_{\varepsilon,\delta}$ and $F_0$.
    \item The final scenario occurs under a sufficiently informative prior $\tilde{\pi}$ (here due to large $n_L$) such that synthetic data is not observably of any use at all and immediately causes the model to perform worse.
\end{enumerate}

We can further quantify what is perhaps the most interesting characteristic of these experiments: the turning points. To do this we formulate bootstrapped averages of the number of `effective real samples' that correspond to the estimated optimal quantity of synthetic data. This is done by comparing these minima with the black curves in the `branching' plots representing the learning trajectory under an increasing $n_L$ and $m=0$. These calculations are shown in Figure \ref{fig:Neff}; the Supplementary contains more details.

In general, we observe a significant increase in performance from the \BD (see Figures \ref{Fig:GaussBranched}, \ref{Fig:ModelComparison}), indicated by its proximity to even the noise-aware model at lower values of $n_L$, and more modest improvements from reweighting methods. The \BD achieves more desirable minimum-trajectory log score, \KLD and Wasserstein values in the majority of cases compared to the other model types, and also exhibits greater robustness to larger amounts of synthetic data where other approaches can lose out significantly.

%Observation 1 and 2 can be illustrate by the following schematic - where we consider prior predictive, $F_0$ and $G_0$ on the space of probability measures over $x$. SV: I would only add the figure - a detailed discussion of a geometric interpreation can be found in 

\subsection{Logistic Regression}

We now move on to a more prevalent and practical class of models that also exhibit the potentially dangerous behaviours of synthetic data in a real-world scenario on datasets concerning subjects that have legitimate potential privacy concerns. Namely, we build logistic regression models for the UCI Heart Disease dataset \citep{Dua2017-vp} and the Framingham Cohort dataset \citep{splansky2007third}. Clearly, we are now only able to access the empirical distribution $F{_n^\ast}$, where $n^\ast$ is the total amount of data present in each dataset. We use $x_{1:n^T}$ to train an instance of the aforementioned \PATEGAN $\mathcal{G}_{\varepsilon, \delta}$ and keep back $x_{1:(n^\ast - n^T)}$ for evaluation; we then draw synthetic data samples $z_{1:m} \sim \mathcal{G}_{\varepsilon, \delta}$. As before, we investigate how the learning trajectories are affected across the experimental parameter grid.

%The standard log-likelihood of our posterior in the case of logistic regression is that of a logit-parameterised Bernoulli distribution and this defines our first model along with the $w$ robustness parameter product on the likelihood. The second is again the posterior arising from adapting the standard likelihood via the \BD (both formulations are included in Supplementary ??). In this case we cannot formulate a ``Noise-Aware'' model due to the black-box nature of the GAN; this highlights the reality of the model misspecification setting we find ourselves in outside of the simplest of simulated examples.

Again we consider learning using $\ell_w$ and $\ell_{\beta}$ applied to the logistic regression likelihood, $f_{\theta}$ (the exact formulations of these are provided in the Supplement). In this case we cannot formulate a `Noise-Aware' model due to the black-box nature of the GAN; this highlights the reality of the model misspecification setting we find ourselves in outside of simple simulated examples, but can instead define a `resampled' model that recycles the real data in the prior.

\subsubsection{Results and Discussion}

\begin{figure}
\includegraphics[width=\columnwidth]{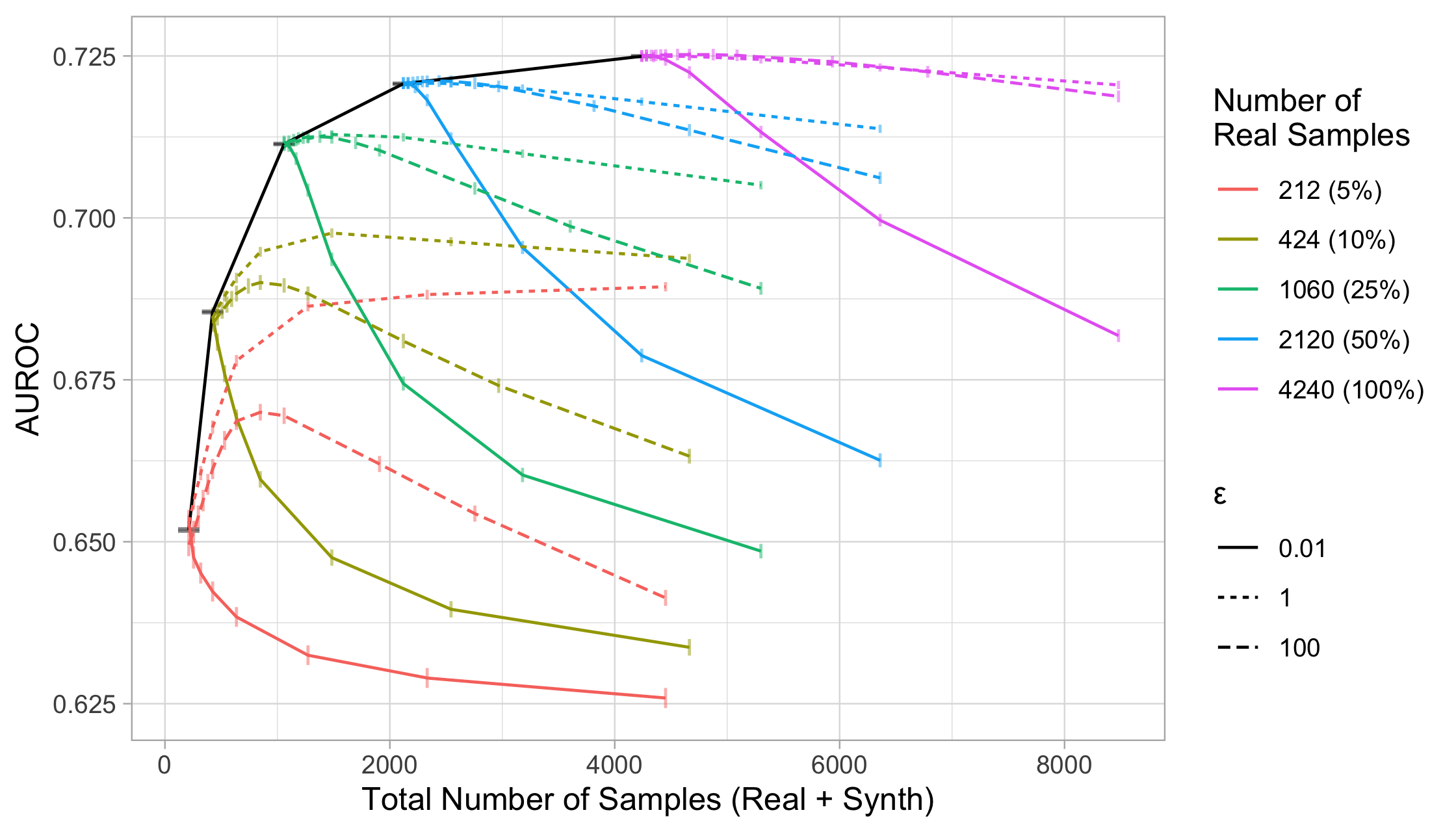}
\caption{ This plot illustrates an interesting and important observation made when varying $\varepsilon$ for a \GAN based model, we observe that there is a privacy `sweet-spot' around $\varepsilon = 1$ whereby more private data performs better than less private data (see the curves for $\varepsilon = 100$).\protect\footnotemark%\\
}
\label{Fig:EpsComparison} 
\end{figure}

Here the learning trajectories are defined with respect to the AUROC as well as the log score; whilst not technically a divergence this gives us a decision theoretic criteria to quantify the closeness of our inference to $F_0$. Referring to Figures \ref{fig:Neff}, \ref{Fig:ModelComparison} and \ref{Fig:EpsComparison} we see that the learning trajectories observed in this more realistic example mirror those observed in our simulated Gaussian experiments. There are however some cases in which the reweighted posterior outperforms the \BD, and importantly we see large discrepancies in $\hat{m}$ when comparing log score to AUROC values meaning the learning task is critical to define.

One particularly interesting observation unique to experiments using synthetic data from a \GAN is that: we see an improvement in performance as epsilon decreases up to a point. We believe this is due to potential mode collapse in the \GAN learning process on imbalanced datasets, and concentration of $\mathcal{G}_{\varepsilon, \delta}$ as the injected noise increases such that a small number of synthetic samples can actually be \textit{more} representative of $F_0$ than even the real data. This effect is short lived as presumably these samples then become over-represented through the posterior distribution and performance begins to fall off, see Figure \ref{Fig:EpsComparison}.

\footnotetext{This plot exhibits the effect under the \BD model on the Framingham dataset with $\beta = 0.5$, but is observable across all model types in both AUROC and log score.}

\section{Conclusions}

%It is clear that synthetic data can help design analysis pipelines that can then be run on the real data by matching the type of data present continuous, missingness and categorical and preserve certain imbalances and characteristics. 
We consider foundations of Bayesian learning from synthetic data that acknowledge the intrinsic model misspecification and learning task at hand. 
% In this article we considered the utility of synthetic data for learning a predictive distribution considering multiple inference techniques and synthetic data generation mechanisms. We have demonstrated that depending on privacy requirements the data can have very limited or no utility. % and contrasted it to how many effective data items it corresponds to. 
%This paper we make two important observations when updating using synthetic data. The fthetic data is not guaranteed to help.
Contrary to traditional statistical inferences, conditioning on increasing amounts of synthetic data is not guaranteed to help you learn about the true data generating process or make better decisions. Down-weighting the information in synthetic data (either using a weight $w$ or divergence \BD) provides a principled approach to robust optimal information processing and warrants further investigation. 

\section*{Acknowledgements}

HW is supported by the Feuer International Scholarship in Artificial Intelligence. JJ was funded by the Ayudas Fundación BBVA a Equipos de Investigación Cientifica 2017 and Government of Spain's Plan Nacional PGC2018-101643-B-I00 grants whilst working on this project. SJV is supported by The Alan Turing Institute (EPSRC grant EP/N510129/) and the University of Warwick IAA funding. CH is supported by The Alan Turing Institute, Health Data Research UK, the Medical Research Council UK, the Engineering and Physical Sciences Research Council (EPSRC) through the Bayes4Health programme Grant EP/R018561/1, and AI for Science and Government UK Research and Innovation (UKRI).

\bibliography{aistats_2021}
\bibliographystyle{plainnat}

\end{document}



% Supplementary material: To improve readability, you must use a single-column format for the supplementary material.
\onecolumn
\section{The $\beta$-Divergence}

The $\beta$-divergence (\BD) was fist introduced by \cite{basu1998robust} under the name `density power divergence', as a robust and efficient alternative to frequentist maximum-likelihood estimation. It has since been used to produce Bayesian posteriors, firstly in \cite{ghosh2016robust} before being unified by generalised Bayesian updating \citep{bissiri2016general} in \cite{jewson2018principles}. Recent applications of the \BD are vast, representing a growing area of research.

The $\beta$-divergence between two distributions with densities $g$ and $f$ with dominating measure $\mu$ (which we in general assume to be the Lebesque measure) is
\begin{align}
    \BD(g\,\|\,f) := \frac{1}{\beta + 1}\int f^{\beta + 1}d\mu - \frac{1}{\beta}\int f^{\beta}g d\mu + \frac{1}{\beta(\beta + 1)}\int g^{\beta + 1} d\mu.
\end{align}
When minimising $\BD(g\,\|\,f_{\theta})$ for $\theta$, the final term $\frac{1}{\beta(\beta + 1)}\int g^{\beta + 1} d\mu$ can be ignored and therefore 
\begin{align}
    \theta^{\BD}_{G} := \argmin_{\theta\in\Theta} \BD(g\,\|\,f) = \argmin_{\theta\in\Theta} \mathbb{E}_{z \sim g}\left[\ell^{(\beta)}(z, f_{\theta})\right],
\end{align}
with the law of large numbers providing that only a sample $\{x_i\}_{1:n}\sim g$ is needed to instantiate such a minimiser
\begin{align}
     \mathbb{E}_{z \sim g}\left[\ell^{(\beta)}(z, f_{\theta})\right] \xleftarrow[n \to \infty]{} \frac{1}{n}\sum_{i=1}^n \ell^{(\beta)}(z_i, f_{\theta}), \quad x_i \sim g,
\end{align}
and from Eq. (9) of the main paper 
\begin{align}
    \ell^{(\beta)}(z,f_{\theta}) :& =\frac{1}{\beta+1}\int f_{\theta}(y)^{\beta+1}dy-\frac{1}{\beta}f_{\theta}(z)^{\beta}.\nonumber
\end{align}
The fact that 
\begin{align}
    \lim_{\beta\rightarrow 0} \frac{1}{\beta}x^{\beta} = \log x,
\end{align}
is then enough to prove that $\lim_{\beta\rightarrow0} \BD = \KLD$.

\subsection{The Robustness of the \BD}{\label{Sec:betaDRobustness}}

This section provides some illustrations demonstrating the robustness of the general Bayesian update using $\ell_w(x, f_{\theta})$ and $\ell^{(\beta)}(x, f_{\theta})$ compared with traditional Bayesian updating (using $\ell_0(x, f_{\theta})$).

\begin{figure}%[h]
\centering{}
\includegraphics[clip,width=0.49\columnwidth,trim= {0.0cm 0.0cm 0.0cm 0.0cm}]{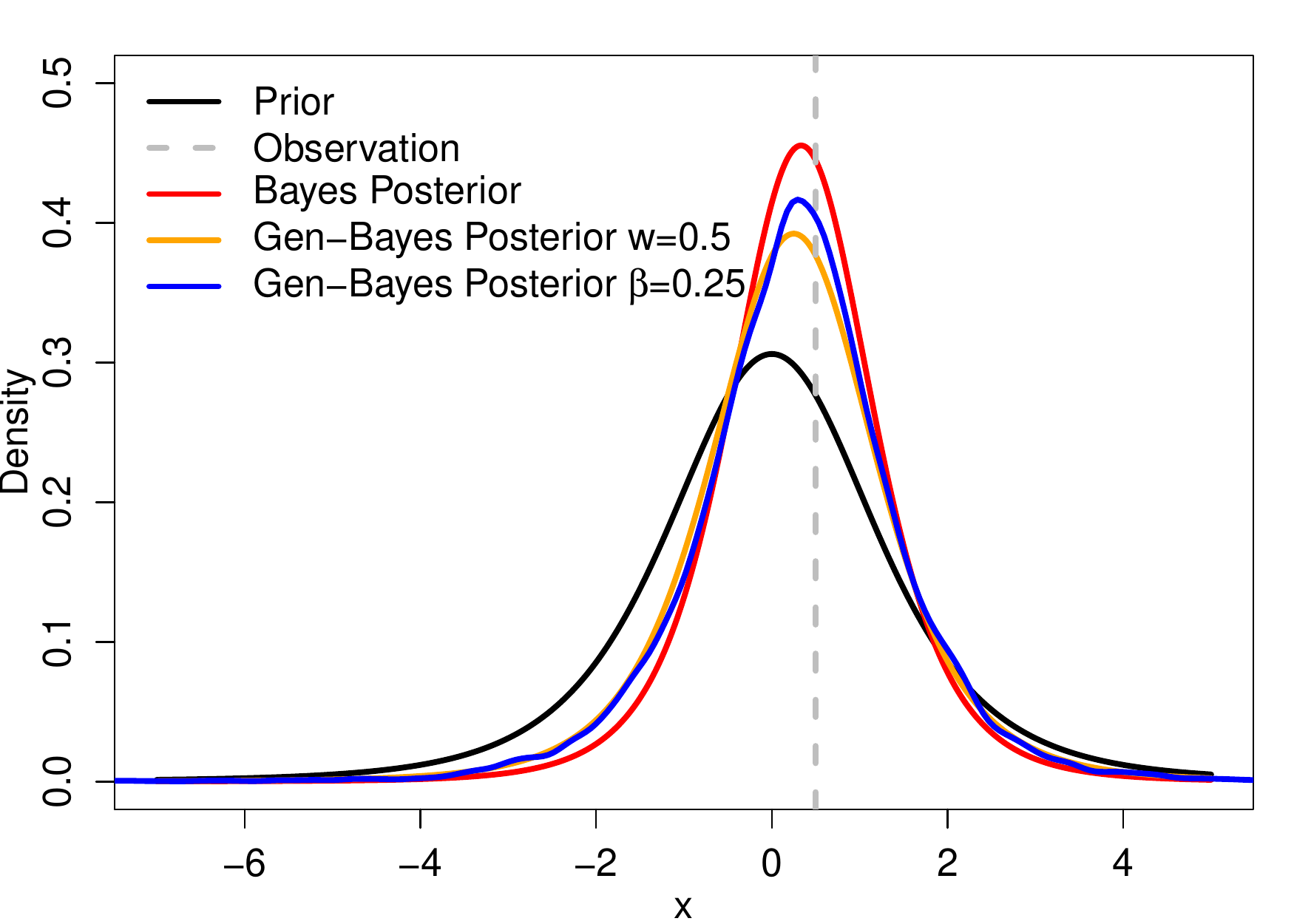}
\includegraphics[clip,width=0.49\columnwidth,trim= {0.0cm 0.0cm 0.0cm 0.0cm}]{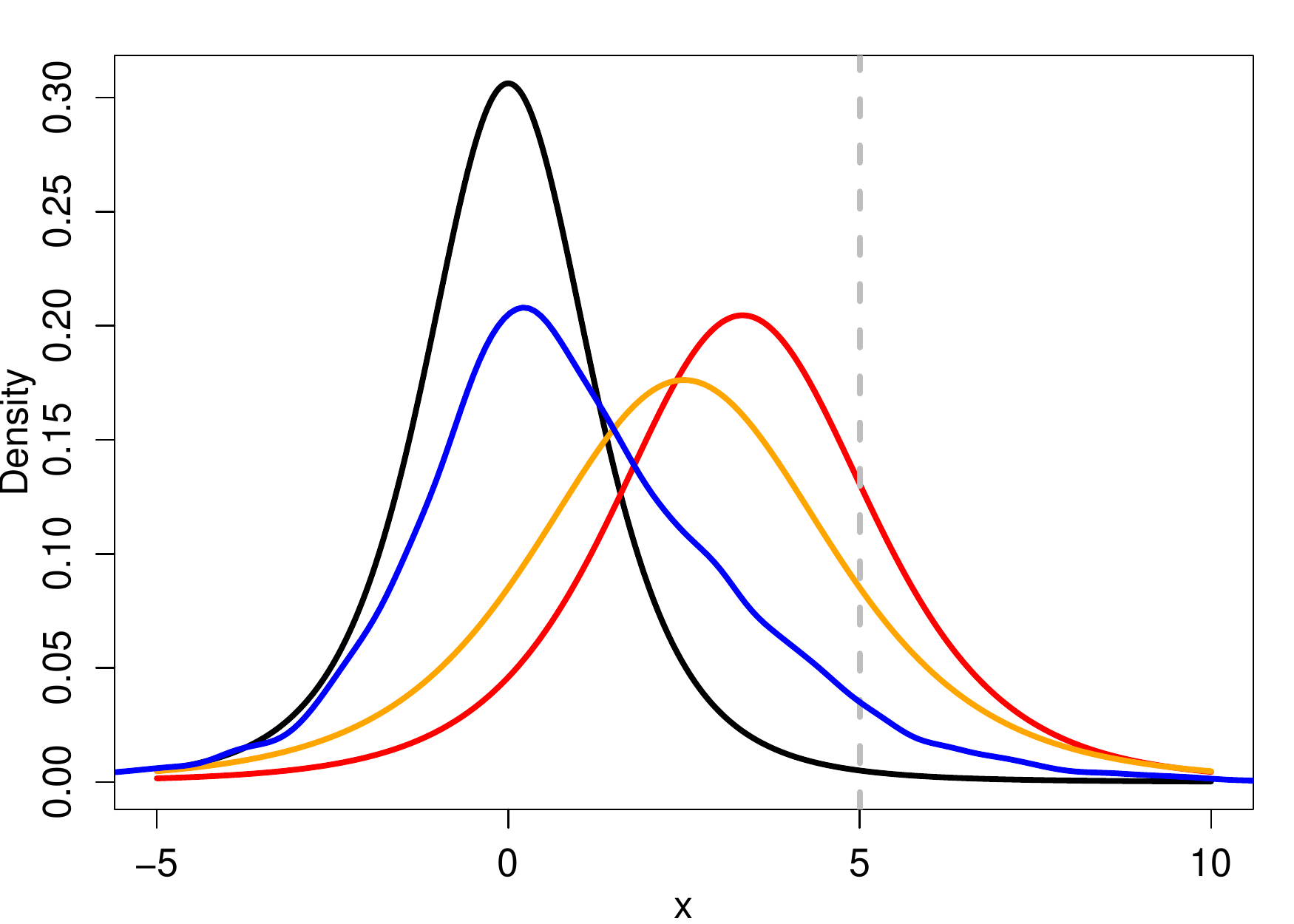}
\includegraphics[clip,width=0.49\columnwidth,trim= {0.0cm 0.0cm 0.0cm 0.0cm}]{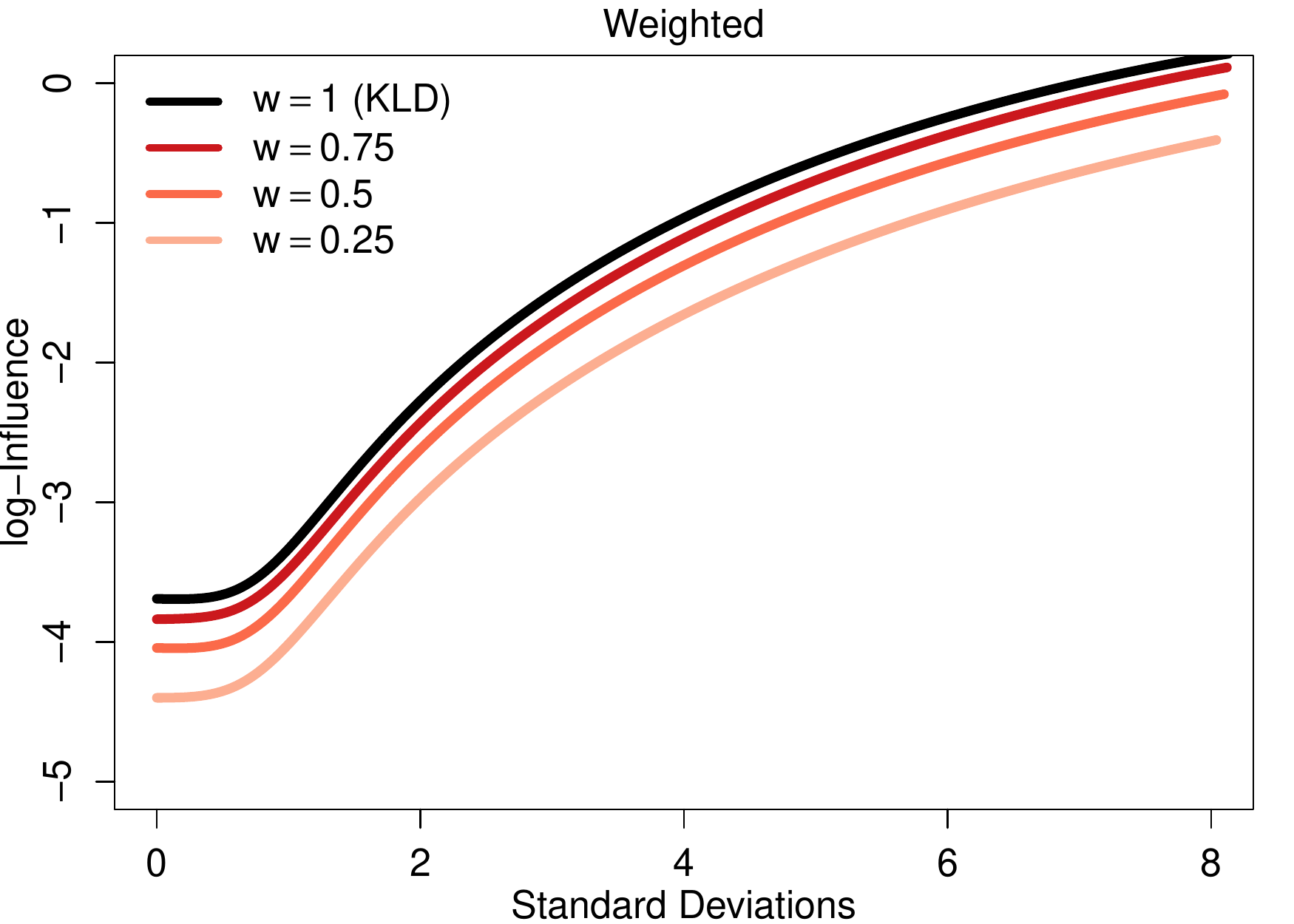}
\includegraphics[clip,width=0.49\columnwidth,trim= {0.0cm 0.0cm 0.0cm 0.0cm}]{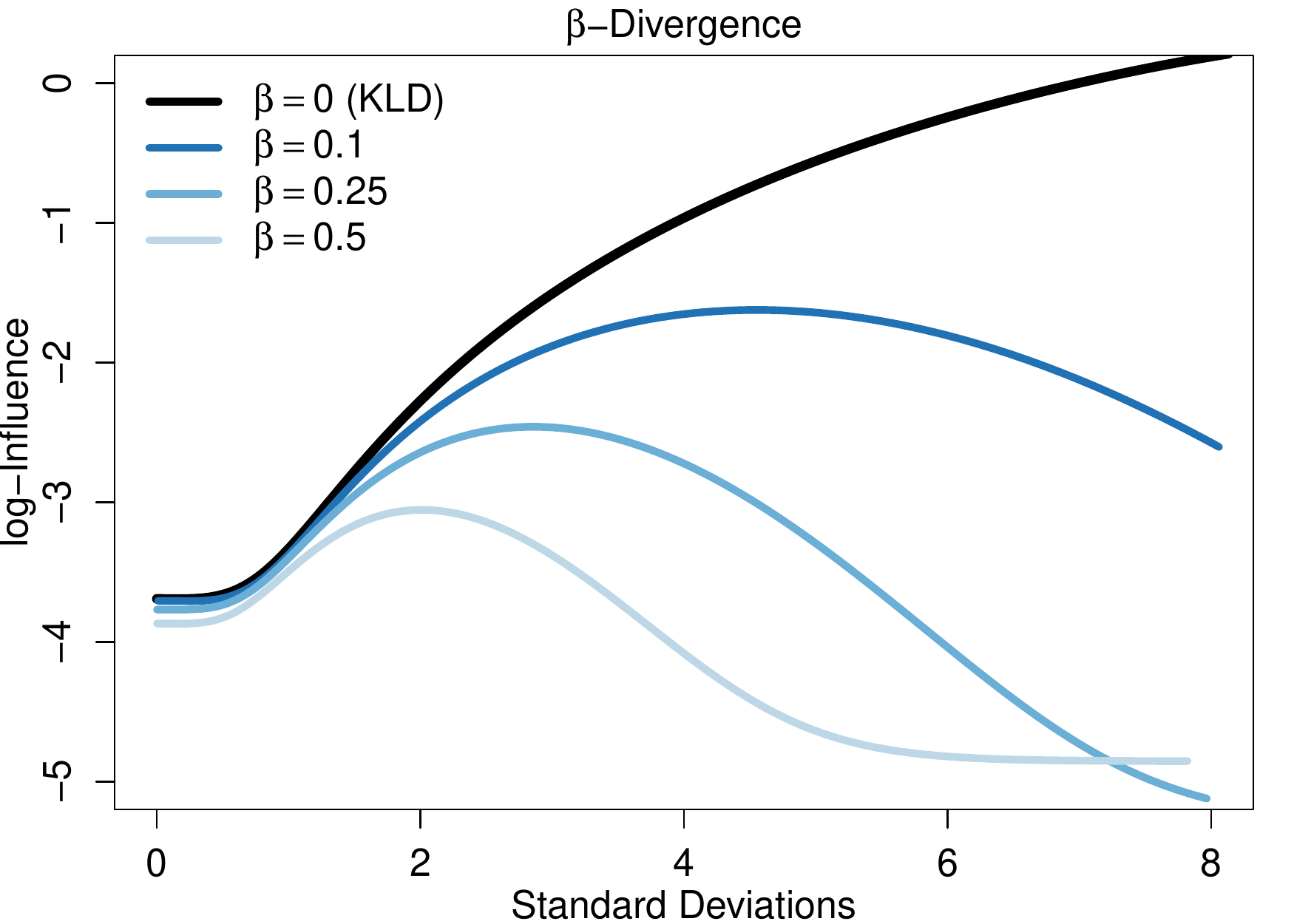}
%trim={<left> <lower> <right> <upper>}
\caption{ Influence of Outliers. \textbf{Top}: NIG Prior predictive ({\color{black}{\textbf{black}}})
and posterior predictives using a Gaussian model, $f_{\theta} = \mathcal{N}(\mu, \sigma^2)$, under traditional Bayesian updating  ($\ell_0(x, f_{\theta})$)
({\color{red}{\textbf{red}}}) and general Bayesian updating with $\ell_w(x, f_{\theta})$ ({\color{orange}{\textbf{orange}}}) and $\ell^{(\beta)}(x, f_{\theta})$ ({\color{blue}{\textbf{blue}}})
after an inlying (\textbf{Left}) and outlying (\textbf{Right}) observation ({\color{gray}{\textbf{grey}}}).
\textbf{Bottom}: log-Fisher-Rao-metric \citep{kurtek2015bayesian} between the general Bayesian posterior with or without one observation at different posterior standard deviations away from the previous posterior mean for $\ell_w(x, f_{\theta})$ (\textbf{Left}) and $\ell^{(\beta)}(x, f_{\theta})$ (\textbf{Right}) under model $f_{\theta} = \mathcal{N}(\mu, \sigma^2)$.
}
\label{Fig:OutlierInfluence} 
\end{figure}

Firstly we consider how observations are `downweighted' compared with the prior under different learning / updating procedures. First, we let the model be a Gaussian location scale model $f_{\theta} = \mathcal{N}(\mu, \sigma^2)$. We start with a conjugate Normal-Inverse-Gamma (NIG) prior, $NIG(\mu, \sigma^2;a_0, b_0, \mu_0, \kappa_0) = \mathcal{IG}(\sigma^2; a_0, b_0)\times \mathcal{N}(\mu; \mu_0, \sigma^2/\kappa_0)$ with $(a_0, b_0, \mu_0, \kappa_0) = (2, 1, 0, 1/2)$ and consider the posterior after observing an observation `in agreement' with the prior $x_{in} = 0.5$ and one not in agreement with the prior $x_{out} = 5$. Figure \ref{Fig:OutlierInfluence}, plots the prior and posterior predictives' densities after observing one observation in these two cases. 

After seeing an `inlying' observation consistent with the prior, all three methods learn similarly, with their posterior predictives' modes being shifted towards the observation. However, we see that using either $\ell_w(x, f_{\theta})$ or $\ell^{(\beta)}(x, f_{\theta})$ gives more relative weight to the prior than traditional Bayesian updating, as they continue to produce larger posterior variances driven by the prior. After seeing an `outlier' the three methods produce very different inferences. One outlying observation can be seen to move the traditional Bayesian inference away from the prior predictive; the same effect is witnessed when using $\ell_w(x, f_{\theta})$, although to a lesser extent in that the posterior predictive also carries a large variance compared to the traditional Bayesian one. Inference under the \BD is very different given an outlier. The posterior predictive mode stays in agreement with the prior but the right tail of the posterior is heavier than the left in order to `acknowledge' the outlying observation.

Next we formalise the influence \citep{kurtek2015bayesian, jewson2018principles} given to different observations under the different learning methods, $\ell_w(x, f_{\theta})$ and $\ell^{(\beta)}(x, f_{\theta})$ w.r.t the Gaussian model $f_{\theta} = \mathcal{N}(\mu, \sigma^2)$.  Here we examine $R(\pi^{(\ell}(\theta|z_{1:n}, x), \pi^{\ell}(\theta|z_{1:n}))$: the Fisher-Rao metric between the general Bayesian posterior based on observations $\{z_{1:n}, x\}$ and the general Bayesian posterior based only on $z_{1:n}$, providing an idea of how an observation at $x$ influences the posterior. Figure \ref{Fig:OutlierInfluence} shows this for variable $x$, loss functions  $\ell_w(x, f_{\theta})$ and $\ell^{(\beta)}(x, f_{\theta})$ for varying $w$ and $\beta$, and $z_{1:n} \sim \mathcal{N}(0, 1)$ with $n = 200$.
The influence plots under $\ell_w(x, f_{\theta})$ are monotonically increasing, showing that as an observation becomes less likely under the current inference its influence over the analysis increases. Decreasing $w<1$ decreases the influence of a new observation, but we can see this happens uniformly meaning an outlier is downweighted by the same amount as an observation near the current posterior mode. Under $\ell^{(\beta)}(x, f_{\theta})$ the influence curves are no longer monotonic as the observation moves away from the current posterior mean. Initially, the influence of observations increases, mimicking inference under $\ell_w(x, f_{\theta})$, but then after a point the influence starts to decrease as these observations become increasingly unlikely given the current inference. This allows \BD-inference to adaptively reject the influence of outliers.  

\begin{figure}%[h]
\centering{}
\includegraphics[clip,width=0.49\columnwidth,trim= {0.0cm 0.0cm 0.0cm 0.0cm}]{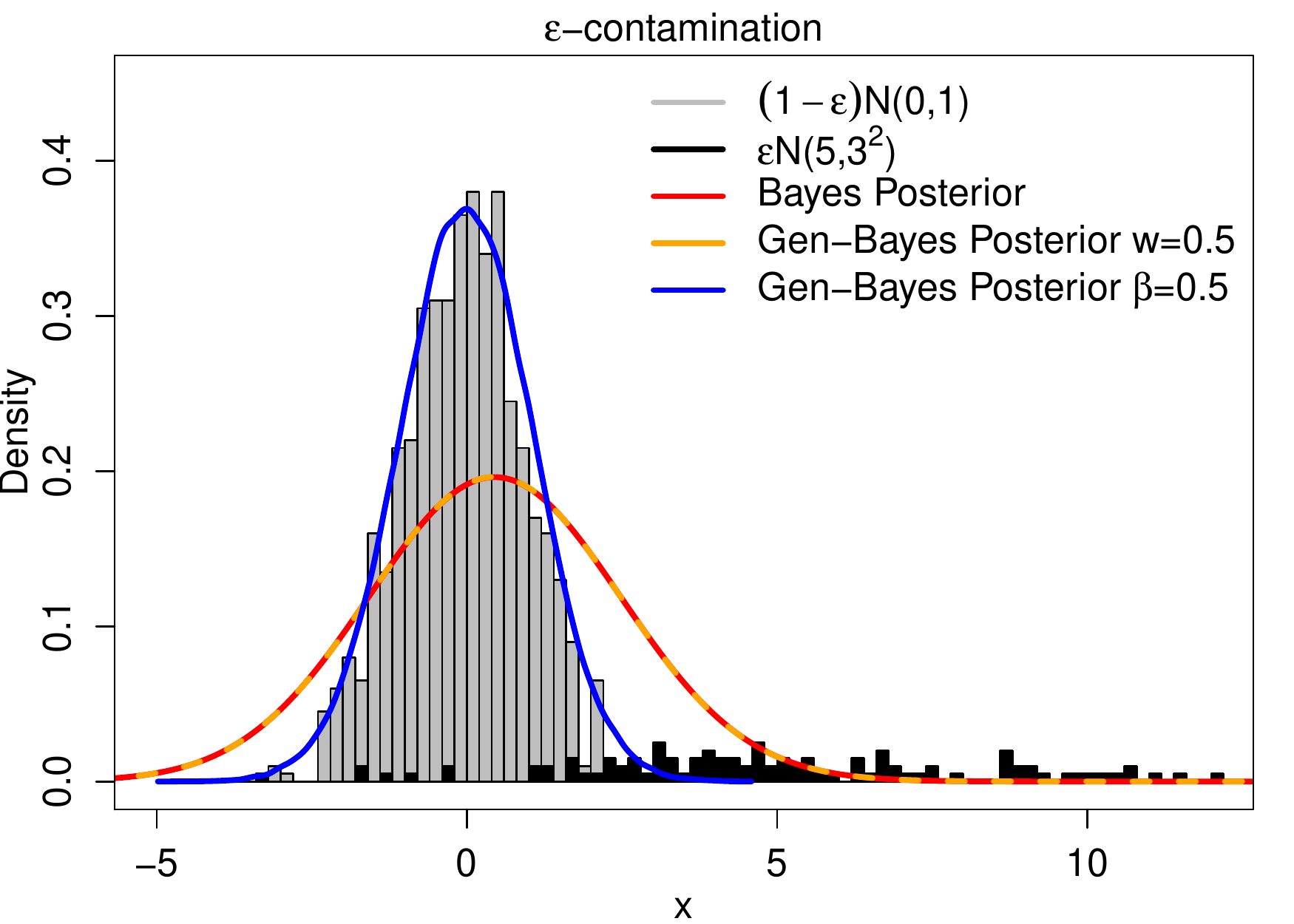}
\includegraphics[clip,width=0.49\columnwidth,trim= {0.0cm 0.0cm 0.0cm 0.0cm}]{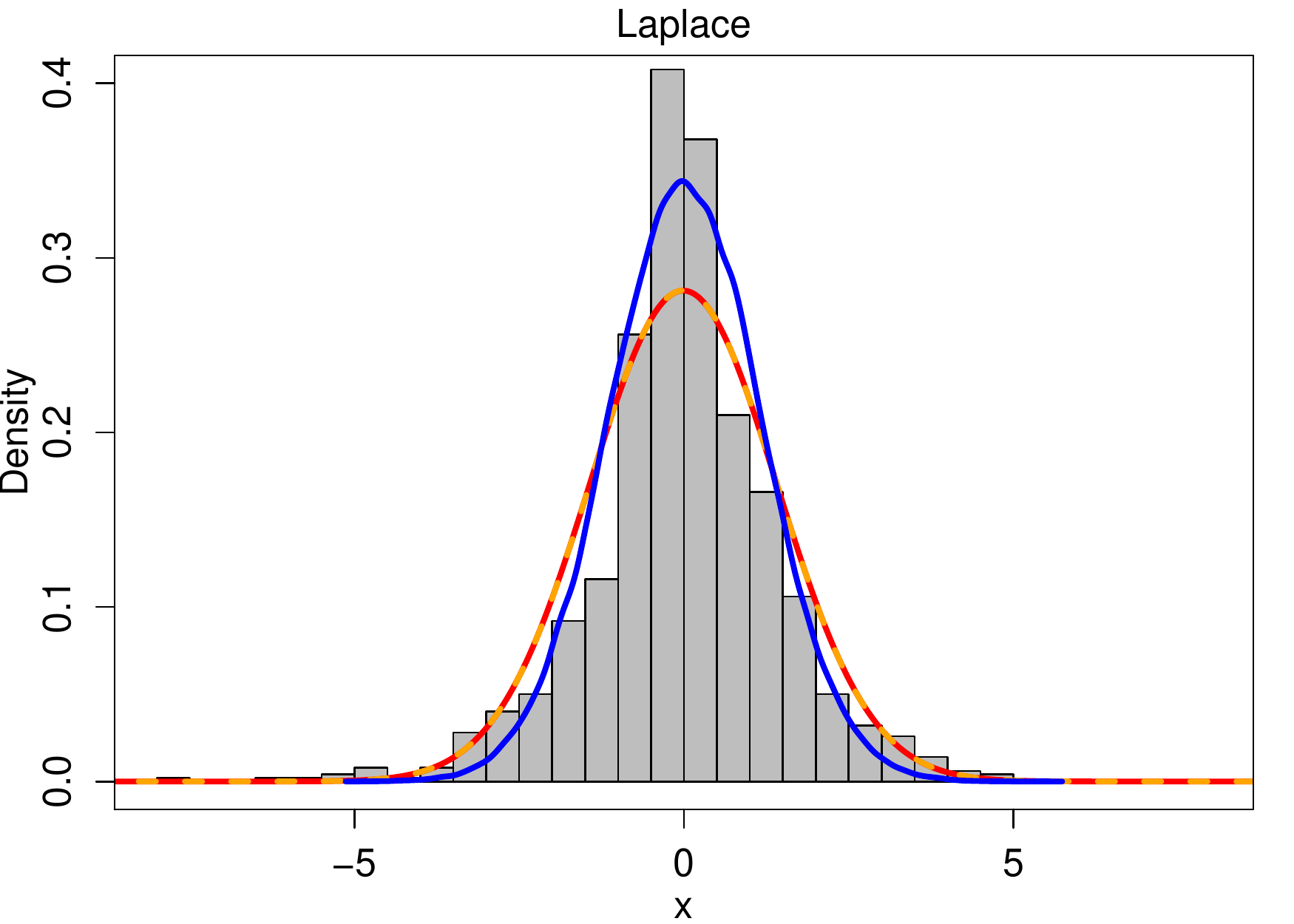}
%trim={<left> <lower> <right> <upper>}
\caption{General Bayesian Predictive densities from $\ell_0(x, f_{\theta})$ ({\color{red}{\textbf{red}}}), $\ell_w(x, f_{\theta})$ ({\color{orange}{\textbf{orange}}}) and $\ell^{(\beta)}(x, f_{\theta})$ ({\color{blue}{\textbf{blue}}}) given $n = 1000$ observations from $g_1(x) = 0.9\mathcal{N}(0, 1^2) + 0.1\mathcal{N}(5, 3^2)$ (\textbf{Left}) and $g_2(x) = \text{Laplace}(0, 1)$ (\textbf{Right}) under model $f_{\theta} = \mathcal{N}(\mu, \sigma^2)$}
\label{Fig:epsilonContamination} 
\end{figure}

Lastly we show how the downweighting of the influence of observations illustrated above affects inference for large samples. We consider inference for a Gaussian model $f_{\theta} = \mathcal{N}(\mu, \sigma^2)$ based on two datasets of size $n = 1000$ generated from  $g_1(x) = 0.9\mathcal{N}(0, 1^2) + 0.1\mathcal{N}(5, 3^2)$ and $g_2(x) = \mathcal{L}(0, 1)$. Generating process $g_1$ is referred to as an $\epsilon$-contamination, where the model is correct for $(1-\epsilon)$\% of observations but is contaminated with $\epsilon$\% outliers, whilst $g_2$ has heavier tails that $f_{\theta}$. Figure \ref{Fig:epsilonContamination} plots the posterior predictive approximation of both $g_1$ and $g_2$ for traditional Bayesian updating and general Bayesian updating with $\ell_w(x, f_{\theta})$ and $\ell^{(\beta)}(x, f_{\theta})$. Firstly, for $n = 1000$ there is little difference between the traditional Bayesian inference and the general Bayesian inference using $\ell_w(x, f_{\theta})$. Additionally we see that minimising the \BD allows the general Bayesian inference to be less concerned with correctly capturing the tails of $g_1$ and $g_2$ and as a result allows it to provide a more accurate approximation to their modes.

\section{Proof of Propositions}

\subsection{Proof of Proposition 1}

Next we prove that for a given \SDGP $\mathcal{G}_{\varepsilon,\delta}$, model $f(\cdot;\theta)$ and infinite synthetic data sample $z_{1:\infty}\sim \mathcal{G}_{\varepsilon,\delta}$, there exists prior $\pi(\theta)$ and private \DGP $F_0$ such that the $L$ is able to get closer to $F_0$ in terms of D, than if they were to use the \KLD limiting approximation to \SDGP $\mathcal{G}_{\varepsilon,\delta}$ $\theta^{\KLD}_{\mathcal{G}_{\varepsilon,\delta}}$

\begin{proposition}[Suboptimality of learning from the \SDGP]
For \SDGP $\mathcal{G}_{\varepsilon,\delta}$, model $f_{\theta}(\cdot)$, and divergence $D$, there exists prior $\tilde{\pi}(\theta)$, private \DGP $F_0$ and $0\leq m <\infty$ such that 
%\begin{equation}
%    D\left(F_0\,\|\,p(\cdot|z_{1:m})\right) \leq D(F_0\,\|\,f_{\theta^{\KLD}_{\mathcal{G}_{\varepsilon,\delta}}})\label{Equ:SuboptimalittG0D}
%\end{equation}
\begin{equation}
    \mathbb{E}_{z}\left[D\left(F_0\,\|\,p(\cdot\mid z_{1:m})\right)\right] \leq D\left(F_0\,\|\,f_{\theta^{\KLD}_{\mathcal{G}_{\varepsilon,\delta}}}\right)\label{Equ:SuboptimalittG0D}
\end{equation}
where $\theta^{\KLD}_{\mathcal{G}_{\varepsilon,\delta}} := \argmin_{\theta\in\Theta} \KLD(\mathcal{G}_{\varepsilon,\delta}\,\|\,f_{\theta})$ and $p(x\mid z_{1:m})$ is the Bayesian posterior predictive distribution (using $\ell_0$) based on (synthetic) data $z_{1:m}$, 
\begin{equation}
p(x\mid z_{1:m}) = \int f_{\theta}(x)\pi(\theta\mid z_{1:m})d\theta \label{Equ:GeneralBayesianPredictive}   
\end{equation}.
\label{Thm:SuboptimalityG0D}
\end{proposition}

\begin{proof}
Firstly fix the the divergence D between the \KLD minimising model to $\mathcal{G}_{\varepsilon,\delta}$ and \DGP $F_0$ as
\begin{equation}
K_{\infty} = D\left(F_0\,\|\,f_{\theta^{\KLD}_{\mathcal{G}_{\varepsilon,\delta}}}\right).
\end{equation}
Now either $\min_{\theta\in\Theta} D(F_0\,\|\,f_{\theta})) = K_{\infty}$ also, in which case the $D$-minimising approximation to $F_0$ is the same distance from $F_0$ (in terms of distance D), as the \KLD minimising approximation to $\mathcal{G}_{\varepsilon,\delta}$ and Eq. \eqref{Equ:SuboptimalittG0D} hold with equality. Such a situation would happen if $\mathcal{G}_{\varepsilon,\delta} = F_0 = f(\cdot;\theta_0)$ for example.
Or we can find $\pi$ such that $\mathbb{E}_{z}\left[D\left(F_0\,\|\,p(\cdot\mid z_{1:m})\right)\right] < K_{\infty}$, for example $\pi(\theta) = 1_{\theta^{\prime}}$ for $\theta^{\prime}$ such that $D(F_0\,\|\,f(\cdot;\theta)) < K_{\infty}$ and therefore Eq. \eqref{Equ:SuboptimalittG0D} holds with $m = 0$.
\end{proof}

We know that under regularity conditions and as $m\rightarrow\infty$, Bayes rule will concentrate about the parameter $\theta^{\KLD}_{\mathcal{G}_{\varepsilon,\delta}} := \argmin_{\theta\in\Theta} \KLD(\mathcal{G}_{\varepsilon,\delta}\,\|\,f(\cdot;\theta))$ \citep{berk1966limiting}; as such we can conclude that given an infinite sample from an \SDGP $\mathcal{G}_{\varepsilon,\delta}\neq F_0$, it is not necessarily optimal to use all of the data available, contrary to the logic of standard statistical analyses. 

Note that in general there is nothing about Proposition \ref{Thm:SuboptimalityG0D} that is specific about using traditional Bayesian updating and learning about $\theta^{\KLD}_{\mathcal{G}_{\varepsilon,\delta}}$ in the limit. The proof of the proposition is unchanged if we consider for example general Bayesian updating using $\ell^{(\beta)}(x, f_{\theta})$ and limiting parameter  $\theta^{\BD}_{\mathcal{G}_{\varepsilon,\delta}} := \argmin_{\theta\in\Theta} \BD(\mathcal{G}_{\varepsilon,\delta}\,\|\,f(\cdot;\theta))$.

%Further to this we believe there are examples where $0 < m < \infty$ is the best approach to use.

\subsection{Proof of Proposition 2}

Next we provide a result that ensures we do not waste synthetic data when $\hat{m}$ is less than the maximum amount of synthetic data available. If more synthetic data is available than the $\hat{m}$ that are used for inference (e.g. sampling $z_{1:m}, m\to\infty$ from a \GAN), we can average the posterior predictive distribution across different realisations and improve the performance of the predictive distribution if we consider convex proper scoring rules such as the logarithmic score. The proof of this result is simple and relies on Jensen's inequality. 

\begin{proposition}[Predictive Averaging]
\label{thm:avg}Given divergence $D$ with convex scoring rule, $s$, averaging
over different realisations (formulated using different realisations of $z_{1:m}$ indicated by superscript $(b)$) of the posterior predictive depending on different synthetic data sets improves inference:
\begin{align*}
\mathbb{E}_{z}D\left(F_{0}\,\|\,\frac{1}{B}\sum_{b=1}^{B}\tilde{p}\left(x\mid z_{1:m}^{(b)}\right)\right) \leq \mathbb{E}_{z}D\left(F_{0}\,\|\,\tilde{p}\left(x\mid z_{1:m}^{(b)}\right)\right).
\end{align*}

\end{proposition}

\begin{proof}

\begin{align}
\mathbb{E}_{z}D\left(F_{0}\,\|\,\frac{1}{B}\sum_{b=1}^{B}\tilde{p}\left(x\mid z_{1:m}^{(b)}\right)\right)\nonumber
&= \mathbb{E}_{z}\mathbb{E}_{x\sim f_0}\left[s\left(x, \frac{1}{B}\sum_{b=1}^{B}\tilde{p}\left(\cdot\mid z_{1:m}^{(b)}\right)\right)\right] - \mathbb{E}_{z}\mathbb{E}_{x\sim f_0}\left[s\left(x, f_0\right)\right]\nonumber\\
&\leq \mathbb{E}_{z}\mathbb{E}_{x\sim f_0}\left[\frac{1}{B}\sum_{b=1}^{B}s\left(x, \tilde{p}\left(\cdot\mid z_{1:m}^{(b)}\right)\right)\right] - \mathbb{E}_{z}\mathbb{E}_{x\sim f_0}\left[s\left(x, f_0\right)\right]\label{Equ:PredictiveAveraging_Jensens}\\
&= \frac{1}{B}\sum_{b=1}^{B}\mathbb{E}_{z}\mathbb{E}_{x\sim f_0}\left[s\left(x, \tilde{p}\left(\cdot\mid z_{1:m}^{(b)}\right)\right)\right] - \mathbb{E}_{z}\mathbb{E}_{x\sim f_0}\left[s\left(x, f_0\right)\right]\label{Equ:PredictiveAveraging_IID}\\
&= \mathbb{E}_{z}\mathbb{E}_{x\sim f_0}\left[s\left(x, \tilde{p}\left(\cdot\mid z_{1:\hat{m}}\right)\right)\right] - \mathbb{E}_{z}\mathbb{E}_{x\sim f_0}\left[s\left(x, f_0\right)\right]\nonumber\\
&= \mathbb{E}_{z}D\left(F_{0}\,\|\,\tilde{p}\left(x\mid z_{1:m}^{(b)}\right)\right)\nonumber
\end{align}

Where \ref{Equ:PredictiveAveraging_Jensens} uses Jensen's inequality and the convexity of $s$ and \ref{Equ:PredictiveAveraging_IID} uses the identical distribution of $\tilde{p}\left(x\mid z_{1:m}^{(b)}\right)$ across varying $b$. 

\end{proof}

The significance of this is that more synthetic data can always be used to improve the predictive distribution, but not by naïvely using all of it to learn at once.

%We are averaging over the data so we can't think about different m's or m being a random variable anymroe I don't think, we would use all of our data to estimate the learning trajectory i think, i.e. consider many repeats of size $m$. I.e. even given a stream we would still want to calculate an expectation over the synthetic data generating process.

\subsection{Elaboration of Remark 1}

Another way to obtain $\hat{m}$ could be dependent on the concrete data stream through the consideration of the trajectory of the evaluation criteria for a concrete sequence of data items. This involves finding the minimum of:

%\jack{I don't understand the objective function in Remark 1, what is sub m sup N and s is the scoring rule so takes data as an argument D is the divergence and would take the DGP density as the argument}

\begin{align*}
   \hat{m}:=\argmin_{0\leq m\leq M}\frac{1}{N}\sum_{j=1}^{N}s(x_{j}^{\prime},p^{\ell}(\cdot|z_{1:m}))
    \end{align*}
Finding this minimum in practice involves potentially adapting the upper bound via optimisation i.e. successively extending a search interval until the local minima is contained (see e.g. \url{http://www.optimization-online.org/DB_FILE/2007/10/1801.pdf}). This estimator depends on the order of the data; the performance of the resulting predictor can be improved using an analog of Proposition 2 by averaging across different shuffles.

\section{The Differential Privacy for Synthetic Data generated under the Normal-Laplace mechanism}

Here we formalise how the Laplace mechanism \citep{dwork2014algorithmic} provides synthetic data with differential privacy guarantees.

\begin{proposition}[Synthetic Data via the Laplace Mechanism]
Given real data $x_{1:n} \in D^n$, synthetic data, $z_{1:n} \in D^n$, generated according to the Laplace mechanism, $\mathcal{T}_{\mathcal{A}}: \mathbb{R}^n \rightarrow \mathbb{R}^n$ with $z_{1:n} = \mathcal{T}_{\mathcal{A}}(x_{1:n}) = x_{1:n} + \delta_{1:n}$ where $\delta_{1:n} \overset{\text{i.i.d.}}{\sim} \textrm{Laplace}(0, \lambda)$, is $(\varepsilon, 0)$-differentially private with $\varepsilon = \max_{x_n, x^{\prime}_n}\frac{|x_n - x^{\prime}_n|}{\lambda}$.
\end{proposition}

\begin{proof}
Fix $x_{1:n} \in D^n$ and consider $x^{\prime}_{1:n} = \{x_{1:n-1}, x^{'}_n\} \in D^n$
\begin{align}
    &\left|\ln\left(\frac{P(\mathcal{T}_{\mathcal{A}}(x_{1:n}) = z_{1:n})}{P(\mathcal{T}_{\mathcal{A}}(x_{1:n}^{\prime}) = z_{1:n})}\right)\right|\nonumber\\
    =&\left|\ln\left(\frac{P(x_{1:n} + \delta_{1:n} = z_{1:n})}{P(x^{\prime}_{1:n} + \delta_{1:n} = z_{1:n})}\right)\right|\nonumber\\
    =&\left|\ln\left(\frac{\frac{1}{\left(2\lambda\right)^n}\exp\left(\frac{\sum_{i=1}^n|x_i - z_i|}{\lambda}\right)}{\frac{1}{\left(2\lambda\right)^n}\exp\left(\frac{\sum_{i=1}^n|x^{\prime}_i - z_i|}{\lambda}\right)}\right)\right|\nonumber\\
    =&\left|\frac{|x_n - z_n|}{\lambda} - \frac{|x^{\prime}_n - z_n|}{\lambda}\right|\nonumber\\
    \leq& \frac{|x_n - x^{\prime}_n|}{\lambda}\nonumber,
\end{align}

As a result, such a procedure provides differential privacy of $ \varepsilon = \max_{x_n, x^{\prime}_n}\frac{|x_n - x^{\prime}_n|}{\lambda}$ by Definition 1 of \cite{dwork2006calibrating}
\end{proof}

In situations where $|x_n - x^{\prime}_n|$ is unbounded, an artificial upper bound $B$ can be imposed with corresponding truncation bounds $\{a, b\}$ with $b - a = B$. For example, this could take the form $\{a, b\} = \{\mu + \frac{B}{2}, \mu - \frac{B}{2}\}$ where $\mu$ is the mean of $F_0$. Any observation $x_i \notin (a, b)$ is instead considered as $\tilde{x}_i = \arg\min_{y \in \{a, b\}} |y - x_i|$ before the addition of the Laplace noise $\delta_i$. Unlike \cite{bernstein2018differentially} we cannot simply redact observations outside the truncation bounds as this would change the dimension of the response and leak privacy. As a result, the trunacated Laplace mechanism for unbounded real data is defined as $z_{1:n} = \mathcal{T}_{\mathcal{A}}(x_{1:n}) = \{\min(\max(a, x), b)\}_{1:n} + \delta_{1:n}$ with $\delta_{1:n} \overset{\text{i.i.d.}}{\sim} \textrm{Laplace}(0, \lambda)$ and provides $(\varepsilon, 0)$-differential privacy with $\varepsilon = \frac{b - a}{\lambda}$.

We note here that the Laplace mechanism provides a more naïve and much simpler method for producing synthetic data compared with methods such as the \DPGAN \citep{xie2018differentially} or \PATEGAN \cite{jordon2018pate}, and that clearly if estimation of variance is important then this mechanism constitutes a poor method to produce synthetic data. However, if one is interested in measures of central tendency, for example estimating coefficients for a regression model then the Laplace mechanism will preserve such features in expectation across the data, which is not necessarily guaranteed by the \GAN based methods. This shows that different methods, producing the same differential privacy guarantee, can have differing desirability to Learner, $L$, depending on which aspects of the \DGP $L$ wishes to capture.

% The discussion surrounding learning trajectories has demonstrated that a promising way to limit the learning is to use \BD combined with reducing the number of synthetic data items used. This has the undesirable effect thatthis depends on
% %
% \begin{enumerate}
%     \item how we are using the realised synthetic data to come up with $\hat{m}$ 
%     \item which $m$(possibly a random variable) data items are selected. 
% \end{enumerate}

\section{Motivating Schematic}

\begin{figure}[h]
\begin{center}
\includegraphics[trim= {9.0cm 1.25cm 9.0cm 1.75cm}, clip,  
width=0.4\columnwidth]{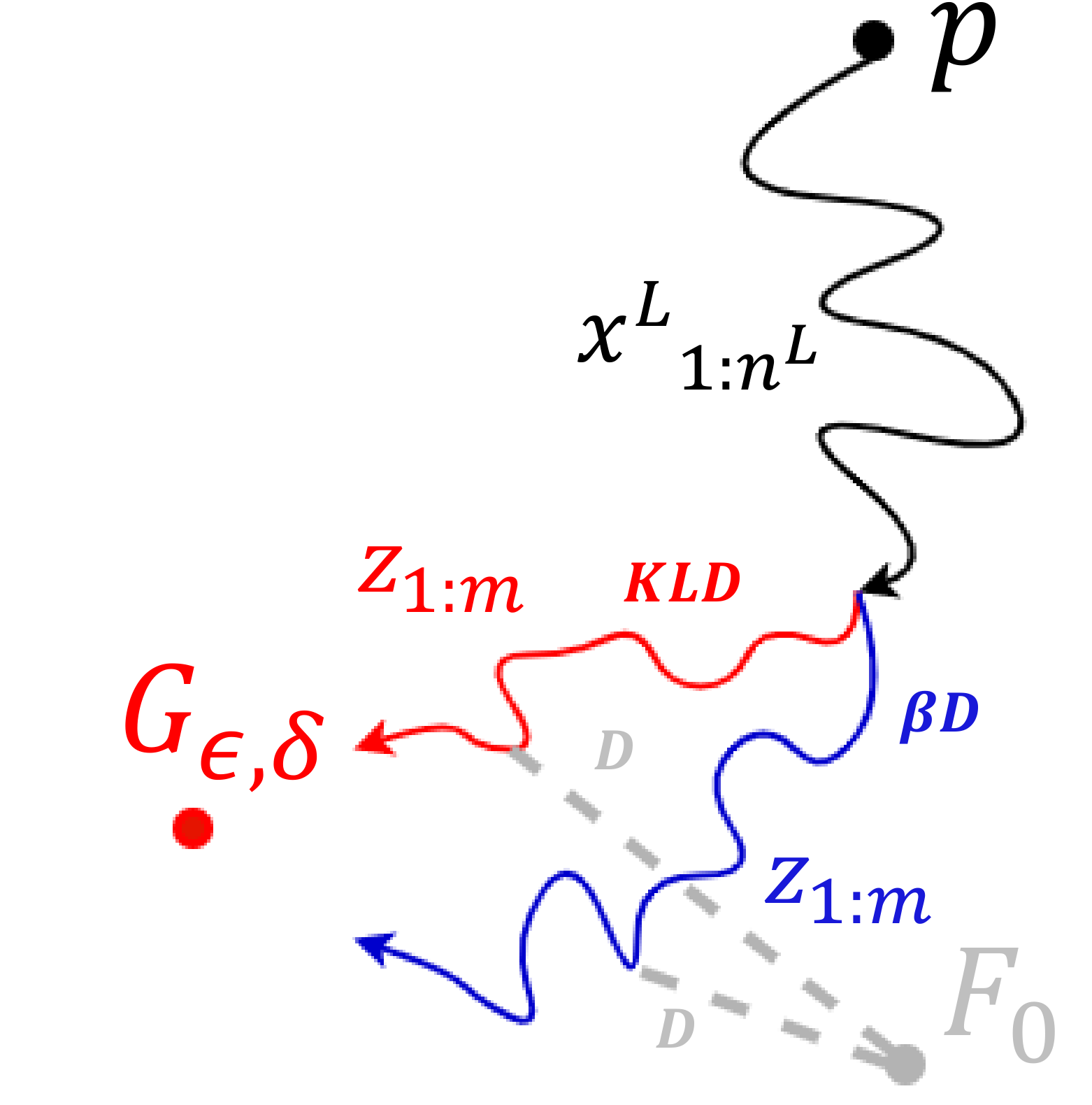}
\includegraphics[trim= {9.0cm 1.25cm 9.0cm 1.75cm}, clip,  
width=0.4\columnwidth]{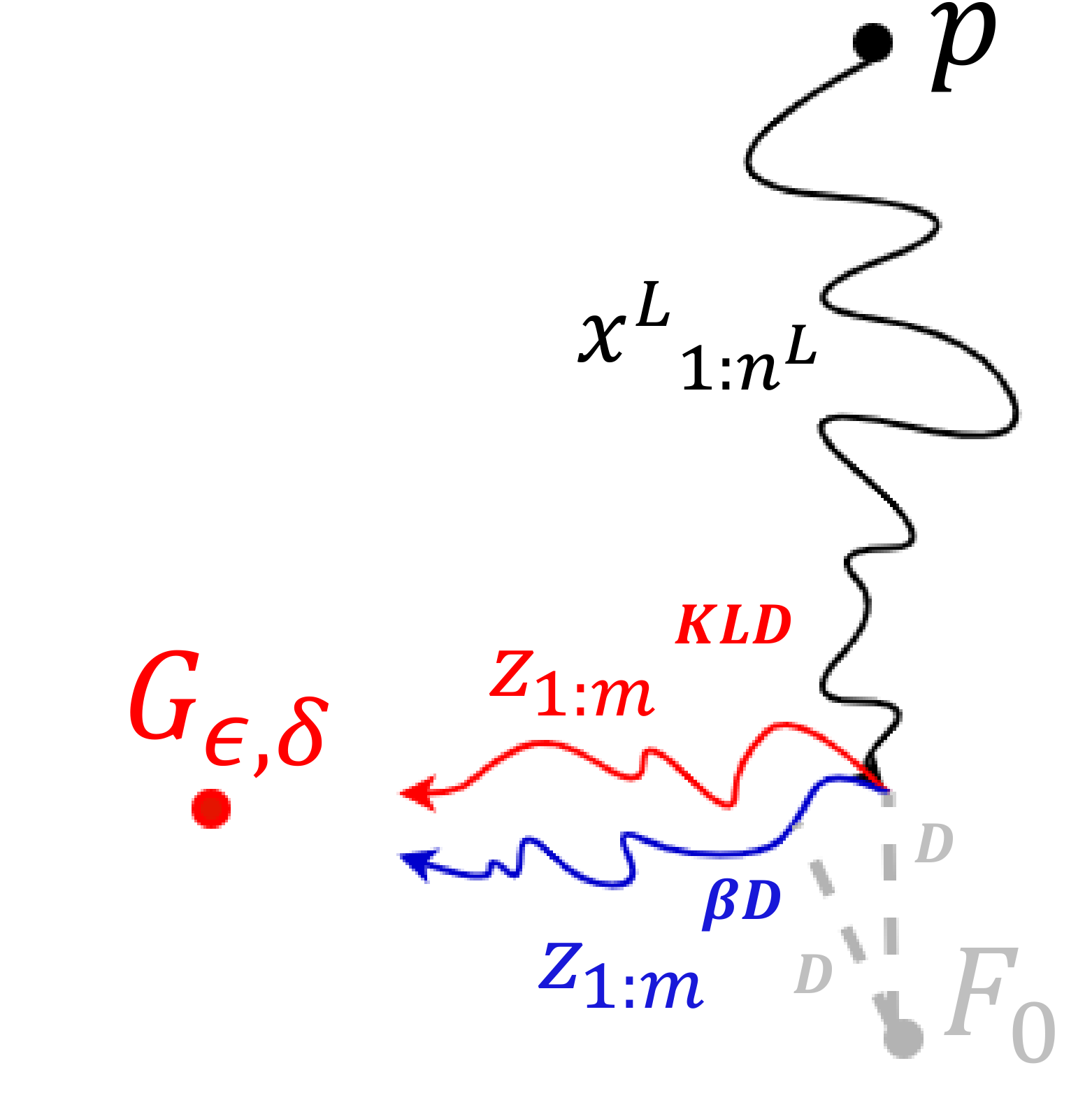}
%trim={<left> <lower> <right> <upper>}
\caption{
The statistical geometry of learning using synthetic data: Starting from prior predictive ($p$), updating using $x^L_{1:n_L}\sim F_0$ takes inference towards $F_0$, before $z_{1:m}\sim \mathcal{G}_{\varepsilon,\delta}$ takes inference towards $\mathcal{G}_{\varepsilon,\delta}$ as $m\rightarrow\infty$.  {\color{red}{\textbf{Red}}} is traditional Bayesian updating (using $\ell_0(z, f_{\theta})$) and {\color{blue}{\textbf{blue}}} is general Bayesian updating using $\ell^{(\beta)}(z, f_{\theta})$. Distances are defined by the divergence $D$. Left: Fewer $n_L$ mean that both learning methods are able to use synthetic data to improve inference for $F_0$ according to $D$. Right: Greater $n_L$ means that before using synthetic data $L$ is closer to $F_0$; when adding synthetic data here traditional Bayesian updating immediately takes inference away from $F_0$ with respect to $D$.
%Right: the model $f(\cdot;\theta)$ is misspecified for $\mathcal{G}_{\varepsilon,\delta}$ and the \BD minimising approximation to $\mathcal{G}_{\varepsilon,\delta}$ is `closer' to $F_0$ than the achieved when minimising the \KLD.
}
\label{Fig:StatisticalGeometrySchematic}
\end{center}
\end{figure}

Fig. \ref{Fig:StatisticalGeometrySchematic} provides a cartoon representing our interpretation of the learning trajectory in the space of distributions for data. Such an interpretation is substantiated by the experiments in Section 4 of the main paper and further in Section \ref{Sec:FurtherResults}. We consider two cases, one where the synthetic data is able to get inference closer to $F_0$ and one where synthetic data immediately makes things worse under traditional Bayesian updating because the beliefs prior to observing the synthetic data were sufficiently informative. Note that `distances' on this schematic are according to the chosen divergence $D$.

The learner $L$ starts at their prior predictive before observing any data, $p(x) = \int f_{\theta}(x)\pi(\theta)d\theta$, and uses their own data $x^L_{1:n_L}$ and Bayes rule (Eq. (6) from the main paper and $\ell_0(z, f_{\theta})$) to update their beliefs. Data $x^L_{1:n_L}\sim F_0$ is not privatised and thus using standard updating draws inference towards the \DGP, $F_0$. By Bayesian additivity this posterior, $\pi(\theta\mid x^L_{1:n_L})$ given observations from $F_0$ becomes the prior for inference using observations $z_{1:m}$. We thus interpret any inference that $L$ can do on their own data as providing a strongly informative prior about $F_0$. Again, we stress that our framework allows for the possibility that $n_L = 0$ here. However, we show later that $n_L>0$ offers an inferential `momentum' in the direction of $F_0$ under the \BD and allows for the synthetic data to bring inferences closer to $F_0$ on the learning trajectory than is possible under traditional Bayesian updating. 

After the initial steps towards $F_0$ following the use of $x^L_{1:n_L}$, $L$ starts learning from the synthetic data $z_{1:m}$ and therefore inference begins to move towards $\mathcal{G}_{\varepsilon,\delta}$. We argue that this does not imply inference is necessarily getting farther from $F_0$ providing acceptance of our minimal assumption that $\mathcal{G}_{\varepsilon,\delta}$ captures some useful information about $F_0$, i.e. we assume that in the model space defined by $D$, $\mathcal{G}_{\varepsilon,\delta}$ and $F_0$ are proximal. Such a phenomenon is depicted on the left-hand side of Fig. \ref{Fig:StatisticalGeometrySchematic}: the red line corresponds to using Bayes' rule when updating using the synthetic data, and we indicate a point on this trajectory that is closer to $F_0$ (according to the chosen divergence $D$) than the inference using only $x_{1:n}$. Up until this point learning about $\mathcal{G}_{\varepsilon,\delta}$ was also helping to learn about $F_0$ but after such a point inference begins to be pulled away from $F_0$. However, on the right-hand side of Figure \ref{Fig:StatisticalGeometrySchematic} we acknowledge that this is not always the case, there can be situations where synthetic data immediately takes inference away from $F_0$. This will happen if the prior information (including the learner's own data) is very strong, or if the \SDGP is far from $F_0$. 

%We differentiate between the cases where  $f(\cdot;\theta)$ is flexible enough to capture $\mathcal{G}_{\varepsilon,\delta}$ and not. When $f(\cdot;\theta)$ is `correctly specified' for $\mathcal{G}_{\varepsilon,\delta}$, minimising either the \KLD or the \BD will correctly capture $\mathcal{G}_{\varepsilon,\delta}$. Therefore, the trajectories have the same limit as $m\rightarrow\infty$ but take different routes to reach this limit. When $f(\cdot;\theta)$ is misspecified for $\mathcal{G}_{\varepsilon,\delta}$ the limiting approximation to $\mathcal{G}_{\varepsilon,\delta}$ will be different, and in particular a \SDGP with heavier tails (i.e. after the injection of Laplace noise) than $F_0$, makes the \BD minimising approximation to $\mathcal{G}_{\varepsilon,\delta}$ closer to $F_0$ than the \KLD minimising approximation to $\mathcal{G}_{\varepsilon,\delta}$. 
%
%So we ask not only whether the limiting parameter to $\mathcal{G}_{\varepsilon,\delta}$ are close to $F_0$, this has already been covered in the robust statistics literature, but where on the trajectory between the prior and the $\mathcal{G}_{\varepsilon,\delta}$ is closest to $F_0$

Additionally, in blue we plot an example trajectory for learning using the \BD.
We argue that the \BD has the ability to get closer to $F_0$ because of its robustness properties. The examples in Section \ref{Sec:betaDRobustness} have demonstrated that inference using the \BD, unlike traditional Bayesian inference, is able to ignore outliers while still learning from inlying observations. As a result, initially the \BD inference is able to learn from observations from the \SDGP that support the inference based on $x^L_{1:n_L}$ and dynamically downweight those that do not, therefore getting closer to $F_0$. Conversely, traditional Bayesian inference is influenced more by observations that disagree with $F_0$ and thus gets pulled more quickly towards $\mathcal{G}_{\varepsilon,\delta}$. This further reinforces the benefits that can be gained by beginning analysis using real data to impart some `momentum' towards $F_0$ when learning from synthetic samples under the \BD. %This also implies that standard reweighing approaches and varying $\beta$'s using the robustness techniques discussed in Section \ref{Sec:RobustStatistics} will further adjust these trajectories. %Next we observe the phenomenon depicted in Fig. \ref{Fig:StatisticalGeometrySchematic} in the following toy example.

\section{Prescribed Methodology}

The discussion surrounding learning trajectories has demonstrated that a promising way to improve inference using synthetic data is to use the \BD-loss combined with a reduced number of synthetic data items. A resulting question is how can one actually action such a procedure for inference given a realisation $z_{1:M}\sim \mathcal{G}_{\varepsilon, \delta}$, for $0< M\leq \infty$ and independent testing set $x^{\prime}_{1:N}\sim F_0$, for $0< N\leq \infty$. To answer this we must address the following questions. 
\begin{enumerate}
    \item How exactly can realisations of real and synthetic datasets be used to estimate $\hat{m}$?
    \item Given that we estimate $\hat{m}<M$, which $\hat{m}$ data items out of the $M$ available should we use for inference?
\end{enumerate}

We provide answers to these two questions in the next subsections.

\subsection{Finding $\hat{m}$}

The optimal $m^{\ast}$ as defined in Eq. (12) of the main paper is an expectation over synthetic data $z\sim \mathcal{G}_{\varepsilon, \delta}$ and real data from the \DGP, $x\sim F_0$, where the second expectation is hidden inside the definition of the Divergence (Definition 2). Given that, at best, we have a sample from the \DGP, $x^{\prime}_{1:N}\sim F_0$, and a sample from \SDGP $z_{1:M}\sim \mathcal{G}_{\varepsilon, \delta}$, $m^{\ast}$ can be estimated by $\hat{m}$ as defined in Eq. (13) of the main paper. We note that even in the case where the \SDGP density could be given to the learner $L$, the expectation in Eq. (12) would likely be intractable and sampling would be required to estimate this integral regardless.

In the case where $L$ is given the ability to sample from the \SDGP arbitrarily, they can continue to sample independently from $\mathcal{G}_{\varepsilon, \delta}$ to calculate $\hat{m}$. Clearly the more samples that they draw, the more accurate this estimation will be, but in reality fixing a computational sampling budget within this scheme is both inevitable and sensible.

When this is not the case, namely there exists $z_{1:M}\sim \mathcal{G}_{\varepsilon, \delta}$, for $0< M< \infty$, the learner can repeatedly sample with or without replacement from $z_{1:M}$ to estimate $\hat{m}$. Clearly repeated sampling is only beneficial to the extent to which new samples are not too dependent on previous ones, something that will be determined by the relative size of the $m$'s under consideration compared with $M$. If $\hat{m}\approx M$ (i.e. $m$ is of the same order of magnitude as $M$) then estimating this integral from one sample is the best that the learner can do. 

\subsubsection{A $p$-value for the Use of Synthetic Data}

In order to guard against the possible variance in calculating $\hat{m}$ from collections of real and synthetic data that might not be sufficiently large, we wish to provide a minimum guarantee that inference using $\hat{m}$ synthetic data samples is no worse than inference using only prior $\tilde{\pi}$ (including any of their own data and expert knowledge). In order to do so, the testing set $x^{\prime}_{1:N}\sim F_0$ can be split into independent subsets, one of which is used to estimate $\hat{m}$ using Eq. (12) of the paper, whilst the other subset is used to construct a $p$-value that the divergence $D$ is significantly reduced using $\hat{m}$ synthetic data samples compared to using no synthetic data at all, i.e. $m = 0$. The divergences are estimates as sums and therefore the central limit theorem can be invoked to construct such a $p$-value 

Moreover, to guard against the variance in splitting a possible small testing set, this procedure can be repeated. For example, repeatedly splitting the data $K$ times allows for the production of $K$ $p$-values. Then a similar procedure to that considered in \cite{watson2020machine} can be used to `aggregate' these $K$ $p$-values. They adapt a procedure of \cite{meinshausen2009p} which given a set of $K$ $p$-values $\{p_1, \ldots, p_K\}$ produces valid aggregated $p$-value, 
\begin{equation}
    p_{\textrm{aggregated}}^{(\textrm{median})} := \min(1, \textrm{Median}(2p_1, \ldots, 2p_K)). 
\end{equation}

This facilitates a valid and robust test for whether to use synthetic data or not.
\begin{itemize}
    \item If this test fails to reject the null hypothesis then the synthetic data from $\mathcal{G}_{\varepsilon, \delta}$ is disregarded ($\hat{m} = 0$) and the learner should just continue with $\tilde{\pi}$.
    \item If this test rejects the null hypothesis in favour of using synthetic data then $\hat{m}$ can be re-estimated using the whole testing set $x^{\prime}_{1:N}$ and returned to the learner.
    \end{itemize}

%In practice the synthetic data might be sorted a particular attribute and this picking the first $m$ might skew the inference. Remedies of this include 
%\begin{enumerate}
%    \item shuffling the data;
%     \item on random chunks e.g. shuffle and take batches of size $m$ and average
%    \item using Proposition ??? and averaging predictive distribution
%    \item exhaustively \begin{align}
%\frac{1}{\left(\begin{array}{c}
%n\\
%m
%\end{array}\right)}\sum_{(i_{1},\dots,i_{m})\subseteq\{1:n\}}\tilde{p}(\cdot\mid z_{i_{1}:i_{m}})
%\end{align}
%  
%    \item sample random subsets of size $m$ and approximate the above using Monte Carlo 
%\end{enumerate}
%Only the exhaustive method ensures that predictive distribution does not depend on the order of the data. The others still depend on the reordering. The effect of a random reordering will be lower for the 2) and 5).

\subsection{Inference Given $\hat{m}$}

Given that $L$ has estimated $\hat{m}<M$, how should they do inference using only $\hat{m}$ out of a possible $M$ samples? Should they simply take the first $\hat{m}$ samples? This would introduce an undesirable dependence on the ordering of the data. Whilst this could be remedied by shuffling the data, we invoke Proposition 2 which shows that inference is improved by averaging posterior predictives over many synthetic data sets, $z_{1:\hat{m}}$ of size $\hat{m}$. By Proposition 2 the learner should repeatedly sample subsets of size $\hat{m}$ from the $M$ available, with or without replacement, conduct posterior inferences using each one independently, and then average their posterior predictions. This has been shown to perform better in expectation according to divergence $D$ than using any single synthetic data subset of size $\hat{m}$. 

%TODO Given  $\hat{m}$ what should you - by Prop 2 you should average. Oviosuly the more with diminishing returns - so instead of exhaustaive averaging we fix computational budget and compute B predictive distributions average them. 

%\subsubsection{Evaluating $\hat{m}$}

\section{Additional Experimental Details and Results}

%If you are given a synth dataset, or a DGP, fix a comp budget num of repeats, then subset / sample from DGP to estimate $\hat{m}$.

This section details all of the information referred to in Section 4 of the main paper; it provides explicit model definitions, explicit grids that were explored to produce the plots included in the paper, further plots to these experiments, and other information regarding the reproduction of the code including reference to our GitHub repository etc.

\subsection{Explicit Loss Function Formulations}

In the two sections below, we formally define the models that were used in the experiments discussed in the main paper; the parameters that we searched across are formally introduced here as well, to be followed with a full experimental grid.

\subsubsection{Simulated Gaussian}

For the simulated Gaussian experiments, the first loss function is given by the standard log-likelihood for the posterior when learning the parameters of a Gaussian distribution, $f_{\theta}(x) = \mathcal{N}\left(x; \mu, \sigma^2\right)$. It can be written as below, with the addition of a $w$ parameter to indicate that this loss function also encompasses the reweighting approach mentioned in Section 3.2:

\begin{equation}
\begin{aligned}
\ell_{w}(x_i ; \theta) = - w \cdot \log f(x_i ; \theta)
\end{aligned}
\end{equation}

Our second loss function leverages the \BD in lieu of standard Bayesian updating (and its connection to the \KLD) and can be written in closed form:
\begin{equation}
\begin{aligned}
\ell_{\beta}(x_i, f(x_i ; \theta)) = - w_\beta \left(\frac{1}{\beta} f(x_i ; \theta)^{\beta} - \frac{1}{\beta + 1} \int f(z ; \theta)^{\beta + 1} dz\right)= - w_\beta \left(\frac{1}{\beta} f(x_i ; \theta)^{\beta} - \left( (2\pi) ^ {\frac{\beta}{2}} (1 + \beta)^{\frac{3}{2}}\sigma ^ \beta\right) ^{-1}\right),
\end{aligned}
\end{equation}
where $w_\beta$ is a `calibration weight' \citep{bissiri2016general}, calculated via $\beta$ and the data, that upweights the loss function to account for the `cautiousness' of the \BD. By this we mean that the \BD-loss is generally smaller than the log-loss everywhere, rather than only in outlying regions of the data space.

In the case of these simulated examples we can also define a model (and associated log-loss function) that captures the noise-induced privatisation via the Laplace mechanism through the density of a Normal-Laplace convolution, which was first defined in \citep{reed2006normal}, later corrected in \citep{amini2017letter} and now reformulated for our specific case of centred Laplace noise below:

\begin{equation}
\begin{aligned}
&\ell_{\text{Noise-Aware}}(x_i ; \mu, \sigma, \lambda) =\\&\quad -\log \Bigg( \frac{1}{4\lambda} \Bigg( e^{\frac{\mu - y}{\lambda} + \frac{\sigma ^ 2}{2\lambda ^ 2}} \left(1 + \text{erf}\left(\frac{y - \mu}{\sigma \sqrt{2}} - \frac{\sigma}{\lambda\sqrt{2}}\right)\right) + e^{\frac{y - \mu}{\lambda} + \frac{\sigma ^ 2}{2\lambda ^ 2}} \left(1 - \text{erf}\left(\frac{y - \mu}{\sigma\sqrt{2}} + \frac{\sigma}{\lambda\sqrt{2}}\right)\right)\Bigg)\Bigg)
\end{aligned}
\end{equation}

\subsubsection{Logistic Regression}

In the case of our logistic regression examples on real-world datasets, $w$ again allows us to reweight the standard log-likelihood for robustness when learning on the synthetic data to formulate our first loss function based on the logit-parameterised Bernoulli density function:

\begin{equation}
\begin{aligned}
\ell_{w}(x_i ; y_i, \alpha, \theta) = - w \cdot \log\Big( \text{logistic}\left( \alpha + x_i \cdot \theta \right) ^{y_i} + \big(1 - \text{logistic}\left( \alpha + x_i \cdot \theta \right)\big) ^ {(1 - y_i)} \Big),
\end{aligned}
\end{equation}

where:

\begin{equation}
\begin{aligned}
\text{logistic}(x) = \frac{1}{1 + e^{-x}}.
\end{aligned}
\end{equation}

Applying the \BD-loss to the same logit-parameterised Bernoulli density function becomes our second loss function:

\begin{equation}
\begin{aligned}
\ell_{\beta}(x_i ; y_i, \alpha, \theta) = -w_\beta\left(\frac{1}{\beta}\Big( \text{logistic}\left( \alpha + x_i \cdot \theta \right)^{y_i} + \big(1 - \text{logistic}\left( \alpha + x_i \cdot \theta \right)\big)^{(1 - y_i)} \Big)^\beta + \right. \\\quad\left.\frac{1}{\beta + 1} \left( \text{logistic}\left( \alpha + x_i \cdot \theta \right)^{\beta + 1} + \big(1 - \text{logistic}\left( \alpha + x_i \cdot \theta \right) \big)^{\beta + 1} \right)\right)
\end{aligned}
\end{equation}

Here, we cannot formulate a `noise-aware' counterpart as the privatisation is via the black-box generations of the \PATEGAN.

\subsection{Evaluation Criteria}

Here we explicitly define each of our evaluation criteria, which are in general calculated via an evaluation set and an approximation to the posterior predictive using the samples drawn from MCMC chains. For these definitions, we let $P$ and $Q$ be two probability measures.

\subsubsection{\KLD}
The \KLD is defined as: 
$$D_\text{KL}(P \parallel Q) = \int_{\mathcal{X}} \log\left(\frac{dP}{dQ}\right)\, dP$$

\subsubsection{Log Score}

The log score as in \cite{gneiting2007strictly} is a special case of a proper scoring rule defined as:
\begin{align}
    \mathbb{E}_{\theta\sim \pi(\theta|x_{1:n})}\left[ \log f(z;\theta)\right]
\end{align}

\subsubsection{Wasserstein Distance}

Following \cite{wasserstein} we define the 
 Wasserstein distance as:
 $$D_{W}(P,Q)=\sup\left(\int f dP - \int f dQ\mid\text{Lipschitz(f)\ensuremath{\leq1}}\right).$$
 Where $\text{Lipschitz}(f)=\sup_{x\neq y} \frac{\mid f(x)- f(y)\mid }{\left\lVert x-y\right\rVert}$
 
\subsubsection{AUROC}
For a probabilistic binary classification algorithm, the receiver operating characteristic (ROC) curve plots the true positive rate against the false positive rate as a parametric plot across different threshold settings. The area under this curve (AUROC) is equal to the probability that a classifier will rank some random positively labelled datapoint higher than a randomly chosen negatively labelled one; it comprises a common means of evaluating the performance of such classifiers. 

$$ A =  \int_{x=0}^{1} \mbox{TPR}(\mbox{FPR}^{-1}(x)) \, dx $$
where $\mbox{FPR}^{-1}(x) $ is the pseudo-inverse of the FPR that maps a false positive rate of $x$ to the corresponding choice of threshold.
Following \cite{Calders2007-ir} it can also be estimated via: $$\frac{\sum_{t_{0}\in\mathcal{D}^{0}}\sum_{t_{1}\in\mathcal{D}^{1}}\textbf{1}[f(t_{0})<f(t_{1})]}{|\mathcal{D}^{0}|\cdot|\mathcal{D}^{1}|}$$

\subsection{Reproducing Our Experiments}

In order to reproduce the results shown in this supplement and in the main paper, one must execute large scale experiments to explore the effect of various parameters across large grids. This amounts to a significant computational workload that was facilitated by recent advances in probabilistic programming in Julia's Turing PPL \citep{ge2018t}, MLJ \citep{Blaom2020-le} and Stan \citep{carpenter2017stan} and the use of large compute nodes. Specifically, we predominantly relied upon a SLURM cluster managed by XXXXX.%the University of Warwick's RTP research computing platform.

All of the code, experimental configuration specifications and other requirements are laid out in our \href{https://github.com/****%HarrisonWilde/Foundations-of-Bayesian-Learning-from-Synthetic-Data
}{GitHub repository}\footnote{\url{https://github.com/****%HarrisonWilde/Foundations-of-Bayesian-Learning-from-Synthetic-Data
}}; below are the parameter ranges and other values used to produce the plots in the case of the two experiment types discussed in the paper:

\begin{itemize}
    \item 6000 MCMC samples were taken per chain in the case of logistic regression; 4000 in the case of the Gaussian simulations. In both cases, 500 warm-up samples were sampled and subsequently discarded.
    \item We used Stan's NUTS \citep{hoffman2014no} sampler and the NUTS sampler provided by Turing to carry out the majority of the inference tasks; we monitored $\hat{R}$ and other convergence criteria when designing the experiments and during their execution to ensure consistent convergence.
    \item For the \BD based models, $\beta \in \{0.1, 0.2, 0.25, 0.4, 0.5, 0.6, 0.75, 0.8. 0.9\}$. We also upweight each posterior using a multiplicative $w_\beta = 1.25$ to account for the fact that the \BD is more `cautious' in general than standard updating.
    \item For the standard reweighted models, $w\in\{0.0, 0.25, 0.5, 0.75, 1.0\}$.
    \item Data quantities:
    \begin{itemize}
        \item For the simulated Gaussian experiment we jointly varied $n$ and $m$ with $\text{n}\in\{2, 4, 6, 8, 10, 13, 16, 19, 22, 25, 30, 35, 40, 50, 75, 100\}$ and $m\in\{1,2,\dots,99,100,120,140,160,180,200\}$.
        \item For logistic regression we traversed a grid of proportional quantities of data rather than explicit $n$'s for ease of comparison across the chosen datasets, $\alpha_\text{real} \in \{0.025, 0.05, 0.1, 0.25, 0.5, 1.0\}$ and $\alpha_\text{synth} \in \{0.0, 0.01, 0.02, 0.03, 0.04, 0.05, 0.075, 0.1, 0.15, 0.2, 0.4, 0.6, 1.0\}$.
    \end{itemize}
    Note that in both cases, we also ran a consecutive stream of real data values without any synthetic data to produce the black lines plotted on the branching plots and to give a means of comparing the performance of synthetic data, through the expected minima (or maxima in the case of AUROC), with the best case scenario of more real data allowing us to calculate the approximate effective number of real samples that the synthetic data could provide (see the following Section \ref{sec:FigMeaning}).
    \item For evaluation, we generated an additional 500 samples from $F_0$ for the Gaussian, and utilised 5-fold cross validation for the logistic regressions where one fold was used for evaluation in each of the five steps.
    \item Priors:
    \begin{itemize}
        \item In the case of our Gaussian model, we placed conjugate priors on $\theta$ with $\sigma ^ 2 \sim \text{InverseGamma}(\alpha_p, \beta_p)$ and $\mu \sim \mathcal{N}(\mu_p, \sigma_p * \sigma)$. Here we set $\alpha_p = 2.0, \beta_p = 4.0, \mu_p \in \{0.0, 1.0, 3.0\}, \sigma_p \in \{1, 10, 30\}$.
        \item We used uninformative Gaussian priors for the logistic regressions' $\alpha$ (the intercept) and $\theta$ (other parameters), e.g. $\alpha, \theta \sim \mathcal{N}(0, 50)$
    \end{itemize}
    \item For the logistic regression models, we initialised $\alpha$ and $\theta$ through 3 varying approaches:
    \begin{enumerate}
            \item Using MLJ's \texttt{LinearRegression} model to calculate the MLE given the step's amount of real data $n$.
            \item Setting $\theta$ to be a vector of 0's matching the dimension of the dataset
            \item Randomly initialising $\theta$ within a locality of 0 using a standard Gaussian model.
    \end{enumerate}
    It is worth noting that this initialisation was not seen to have much observable effect in terms of MCMC convergence; the number of samples were carefully chosen alongside very effect sampling schemes such as NUTS and HMC to ensure convergence in almost all cases, even with very little or noisy data.
    \item We repeated all of the configurations defined by combinations of the parameter values specified above at least 100 times to ensure reasonable certainty in our results in the presence of multiple sources of noise (data generation, privatisation and MCMC). During each of these `full iterations' we specified and recorded a randomised seed to ensure that the real data used was reshuffled or different each time for the logistic regression and simulated Gaussian experiments respectively. This then allows us to calculate expected curves across different realisations of varying amounts of real data.
\end{itemize}

\subsubsection{Datasets Used for Logistic Regression}

\paragraph{The Framingham Cohort Dataset} contains 4240 rows and 15 predictors, a mix of binary labels and continuous or discrete numerics. The label is a binary indicator for someone's ten year risk of coronary heart disease. Many of the columns such as age, education, cigarettes smoked per day and more pose genuine privacy concerns to the subjects of this dataset.

\paragraph{The UCI Heart Dataset} contains 303 rows and 14 predictors, again a mix of binary labels and continuous or discrete numerics. The label is a binary indicator for the presence of heart disease in a subject. Many of the attributes pose genuine privacy concerns to the subjects of this dataset.

\subsection{Elaboration on the Formulation and Meaning of the Figures}\label{sec:FigMeaning}

The following section further details how each type of figure shown in the paper is made, and how they should be interpreted:

\begin{itemize}
    \item \textbf{The `branching' plots} (as in Figures \ref{Fig:Branching1}, \ref{Fig:Branching2}, \ref{Fig:Branching3}, \ref{Fig:Branching4}, \ref{Fig:Branching5}, \ref{Fig:Branching6}, \ref{Fig:Branching7}, \ref{Fig:Branching8}, \ref{Fig:Branching9}, \ref{Fig:Branching10} and Figure 1 of the main paper (and Figure 4 which is a special case)) show the total amounts of data on the x axis used to train various model via MCMC, each model corresponding to a single point on the plot. This amount is totalled in the sense that it corresponds to some amount of real data $n_L$ added on to some amount of synthetic data $m$. Each `branch' of these plots fixes the amount of real data it represents at the root of its branch from the black line which represents a varying amount of real data and \textit{no} synthetic samples. Each branch is then colour coded and corresponds to some fixed quantity of real data $n_L$ plus a varying amount of synthetic data, such that the amount of synthetic samples included in learning up to a point on the x axis can be calculated by subtracting the fixed real data quantity $n_L$ from the x axis value. The y axis is relatively clear in corresponding to the relevant criteria value when the model trained at each point of the branching curves is evaluated. Note that these plots directly show the `learning trajectory' as depicted in Figure \ref{Fig:StatisticalGeometrySchematic}
    \item \textbf{The model comparison plots} (as in Figures \ref{Fig:ModelComp1}, \ref{Fig:ModelComp2}, \ref{Fig:ModelComp3}, \ref{Fig:ModelComp4}, \ref{Fig:ModelComp5}, \ref{Fig:ModelComp6}, \ref{Fig:ModelComp7}, \ref{Fig:ModelComp8}, \ref{Fig:ModelComp9}, \ref{Fig:ModelComp10} and Figure 3 of the main paper) are in some ways just a more specific view of the `branching' plots discussed above, in that they fix the real amount of data to some $n_L$ and simply illustrate the performance of the models under varying synthetic data quantity $m$ alongside one another. This is essentially a layering of a single consistent branch from each `branching' plot layered on top of each other across all of the model configurations of interest.
    \item \textbf{The $n$-effective plots} (as in Figures \ref{Fig:Neff1}, \ref{Fig:Neff2}, \ref{Fig:Neff3} and Figure 2 of the main paper) illustrate the maximal effective number of real samples that can be gained through the use of synthetic data under varying real data quantity $n_L$. In order to illustrate this, we calculate bootstrapped \citep{efron1994introduction} mean and variance of the minima / maxima by first taking the expectation over each seed / iteration / realisation of a curve alongside the synthetic data varying; this is done separately for each `branch' (i.e. fixed real data quantity $n_L$) to get an expected curve for each. The turning point of these curves is then matched with the closest realised point along the black line representing varying amounts of real data without any synthetic data in order to produce an estimate for the amount of real data samples this turning point effectively corresponds to. Namely, the additional number of real samples required to achieve the same minimum / maximum criteria value when learning using the optimal amount of synthetic data.
    
    This process is done in the bootstrap paradigm such that we repeatedly sample $N=100$ seeds / iterations / curves from those collected during the full experiment, the turning point corresponding to this expectation over $N$ curves is then calculated; this is repeated $B=200$ times to then calculated a bootstrapped mean and variance for the best (i.e. at the turning point in synthetic data) effective number of real samples for each amount of fixed real data.
    
    This can be expressed as below, by taking an evaluation set $x^\prime_{1:n^\prime} \sim F_0$ and then by calculating $B$ $n_\text{eff}^{(b)}$s for each real data quantity $n_L$ that we take, alongside varying synthetic samples. Each $n_\text{eff}^{(b)}$ is a bootstrap $n$-effective sample formulated using the the minimum of the expected curve arising across a randomly sampled $N=100$ seeds / iterations / curves that is then `matched' with a similar expected curve arising from $R=100$ sampled real-only (black lines) seeds / iterations / curves to provide an estimation for the number of real samples each minimum represents:
    
    \begin{align*}
        n_\text{eff}^{(b)} = \argmin_t \left|\frac{1}{M}\sum_{j=1}^{M} s(x^\prime,p(\cdot\mid x_{1:n_L+t}^{(j)})) -  \text{min}_{m}\frac{1}{N}\sum_{i=1}^{N}s(x^\prime,\tilde{p}(\cdot\mid x^{(i)}_{1:n_L} z_{1:m}^{(i)}))\right|
    \end{align*}
    
    This gives rise to a collection of $B$ bootstrap samples for each value of $n_L$ from which we can compute a bootstrapped mean and variance to present in our $n$-effective plots.
\end{itemize}

\subsection{Further Results and Figures}\label{Sec:FurtherResults}

\begin{figure}[H]
\centering
\includegraphics[width=0.85\columnwidth]{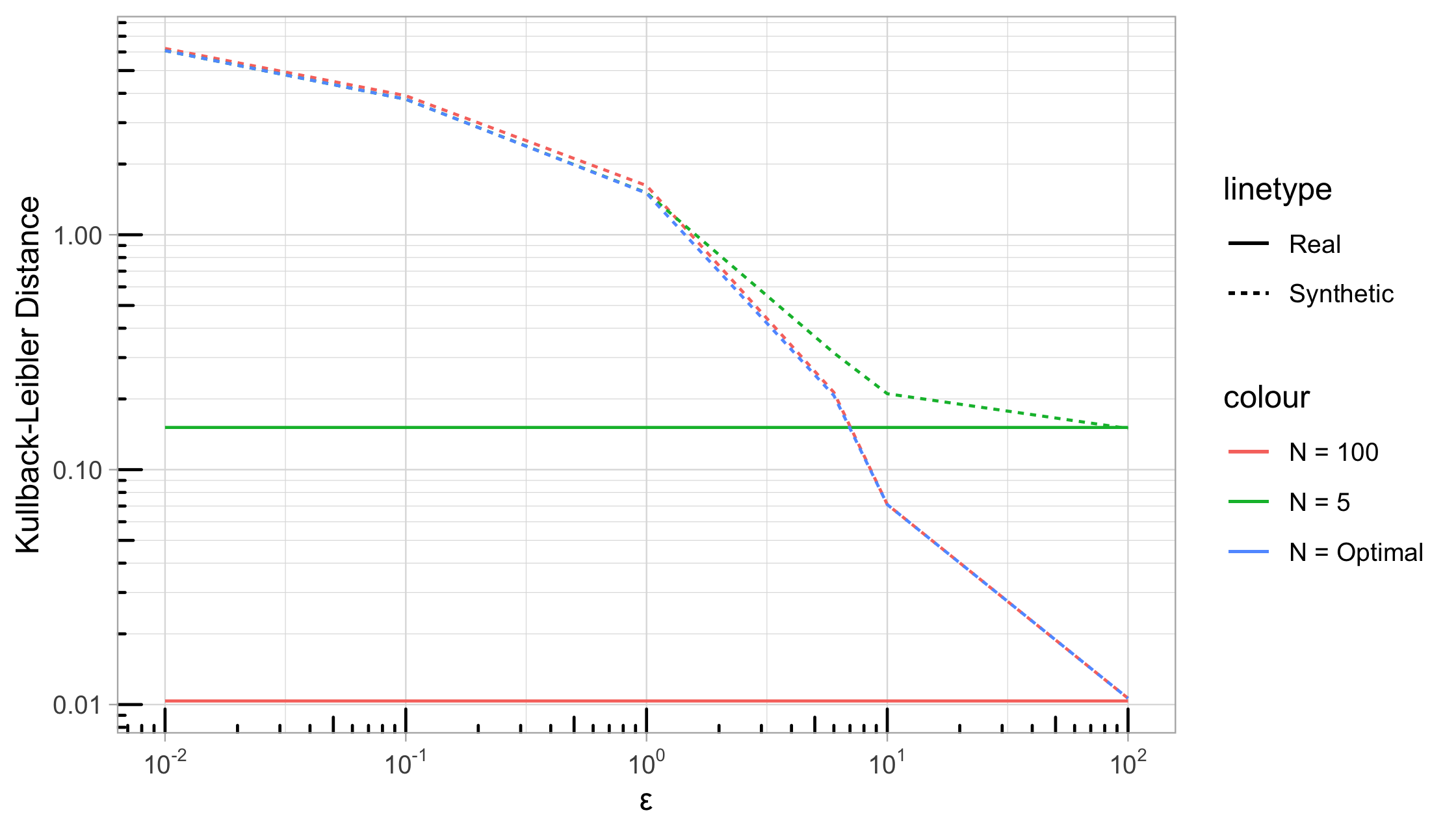}
\caption{ This plot shows the average \KLD between $F_0$ and the models arising from the specified amounts of real or synthetic data under varying \DP-guarantees, the line types distinguish between real and synthetic, and colour indicates the data quantity used for learning. Here `optimal' indicates that $\hat{m}$ samples are used under each $\varepsilon$. As $\varepsilon\to\infty$ the synthetic data becomes arbitrarily close to samples from $F_0$ through the Laplace mechanism such that the question of whether to use any or all of the synthetic data is most interesting at lower $\varepsilon$\protect\footnotemark.%\\
}
\label{Fig:NoiseDemo} 
\end{figure}

Figure \ref{Fig:NoiseDemo} shows the relationship between privacy and model misspecification in terms of the \KLD. We observe that initially, using a small amount of synthetic data is preferable due to the amount of noise a small $\varepsilon$ introduces; there is then a cross over point around $\varepsilon = 1$ to $\varepsilon = 10$ where using more data becomes more desirable as the level of noise decreases, until eventually at $\varepsilon = 100$ we see comparable performance from using the synthetic data to using the same amount of real data. This pattern of usefulness is slightly more complex in the case of a \GAN based model as the usefulness of the data also relies on the convergence of the \GAN and the overall representativeness of its generated samples through the effectiveness of training, regardless of the value of $\varepsilon$.

\begin{figure}[H]
\centering
\includegraphics[width=0.85\columnwidth]{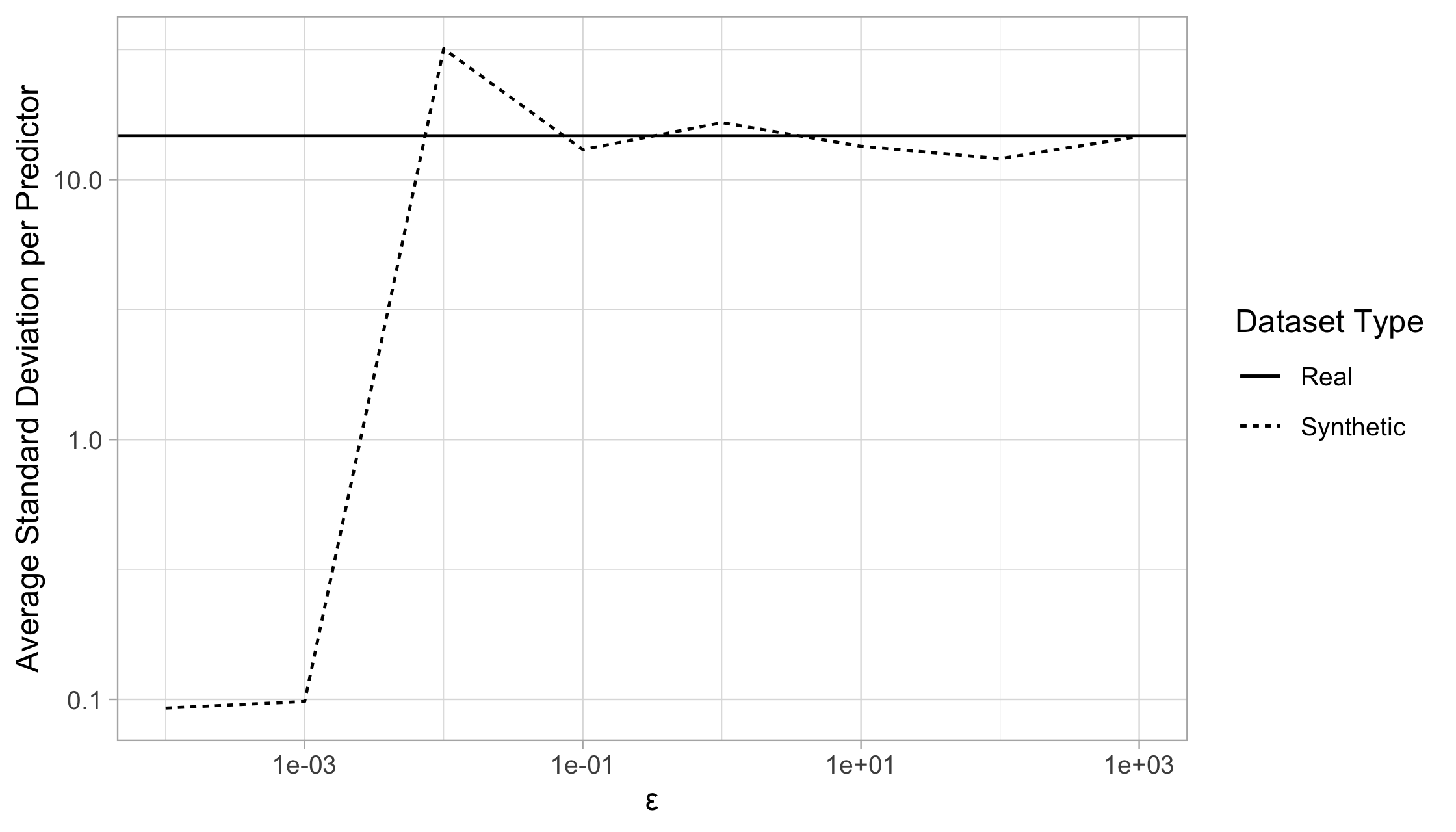}
\caption{ Here, we present the averaged predictor standard deviation for datasets arising from various $\varepsilon$ values. Similarly to Figure \ref{Fig:NoiseDemo}, as $\varepsilon \to \infty$ the synthetic data should more closely resemble the real dataset that was used to train it, plus whatever complications arise by nature of this training. It can be seen that as privacy increases with $\varepsilon \to 0$ that there is a point $\varepsilon^\ast \in [10^{-3}, 10^{-2}]$ where the predictors' standard deviation collapses.
}
\label{Fig:GanError} 
\end{figure}

We can conduct a more fundamental investigation of the \PATEGAN's behaviour under varying $\varepsilon$ by observing through Figure \ref{Fig:GanError} the average predictor standard deviation in the resulting datasets generated by the \GAN under different $\varepsilon$ specifications. This allows us to observe the `mode collapse' of the generative model when $\varepsilon$ is sufficiently small and privacy is sufficiently high. Interestingly, as observed in Figure 4 from the main paper, this data is still somewhat useful, at least in small quantities, in learning about $F_0$.

\subsubsection{Branching Plots}

\begin{figure}[H]
\includegraphics[width=0.97\textwidth]{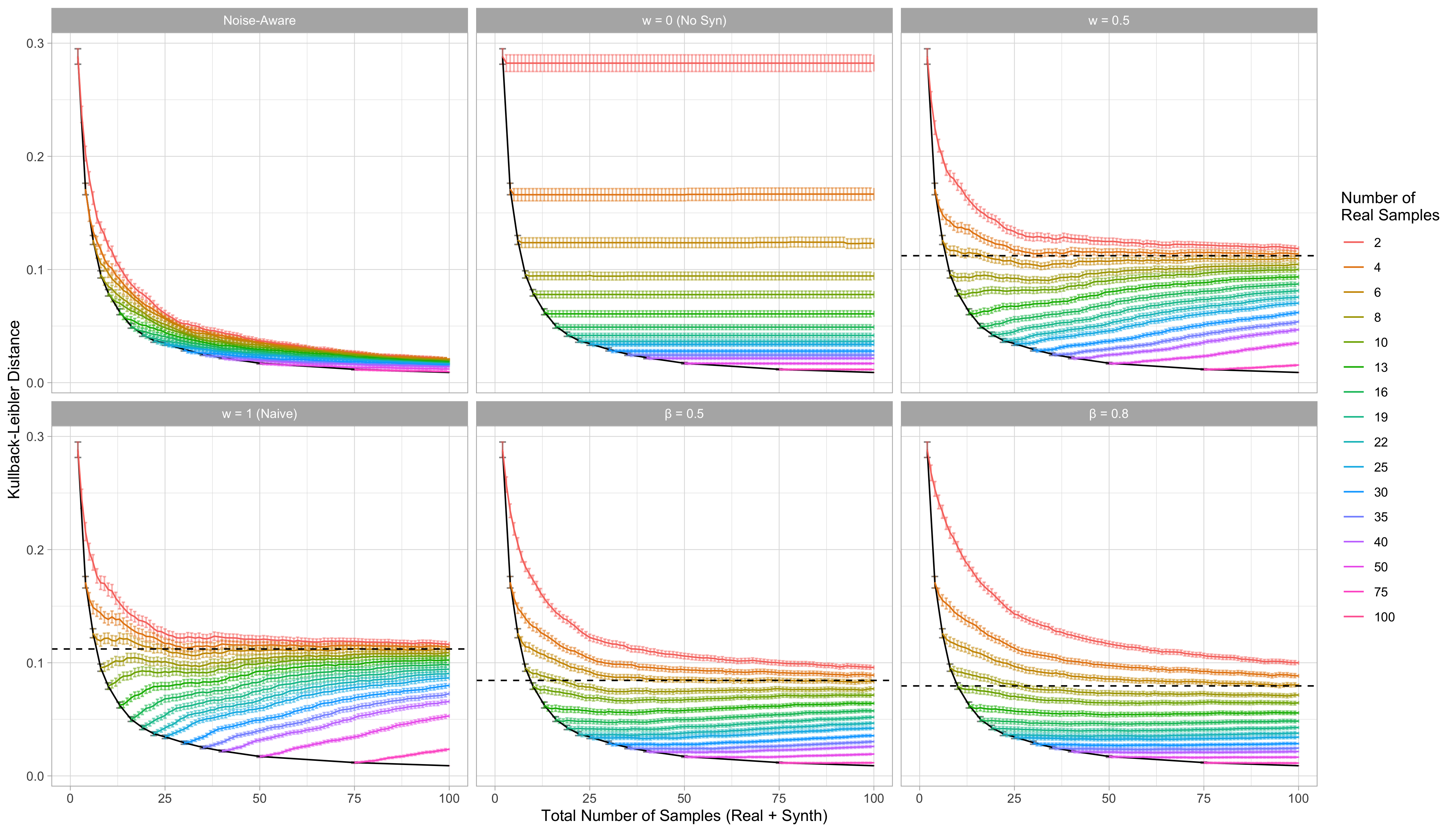}
\caption{
Branching plots for each model configuration in the case of the simulated Gaussian experiments illustrating the \KLD against the total number of samples where \DP of $\varepsilon = 8$ is achieved by the Laplace mechanism via noise of scale $\lambda = 0.75$.
}
\label{Fig:Branching1} 
\end{figure}

\begin{figure}[H]
\includegraphics[width=0.97\textwidth]{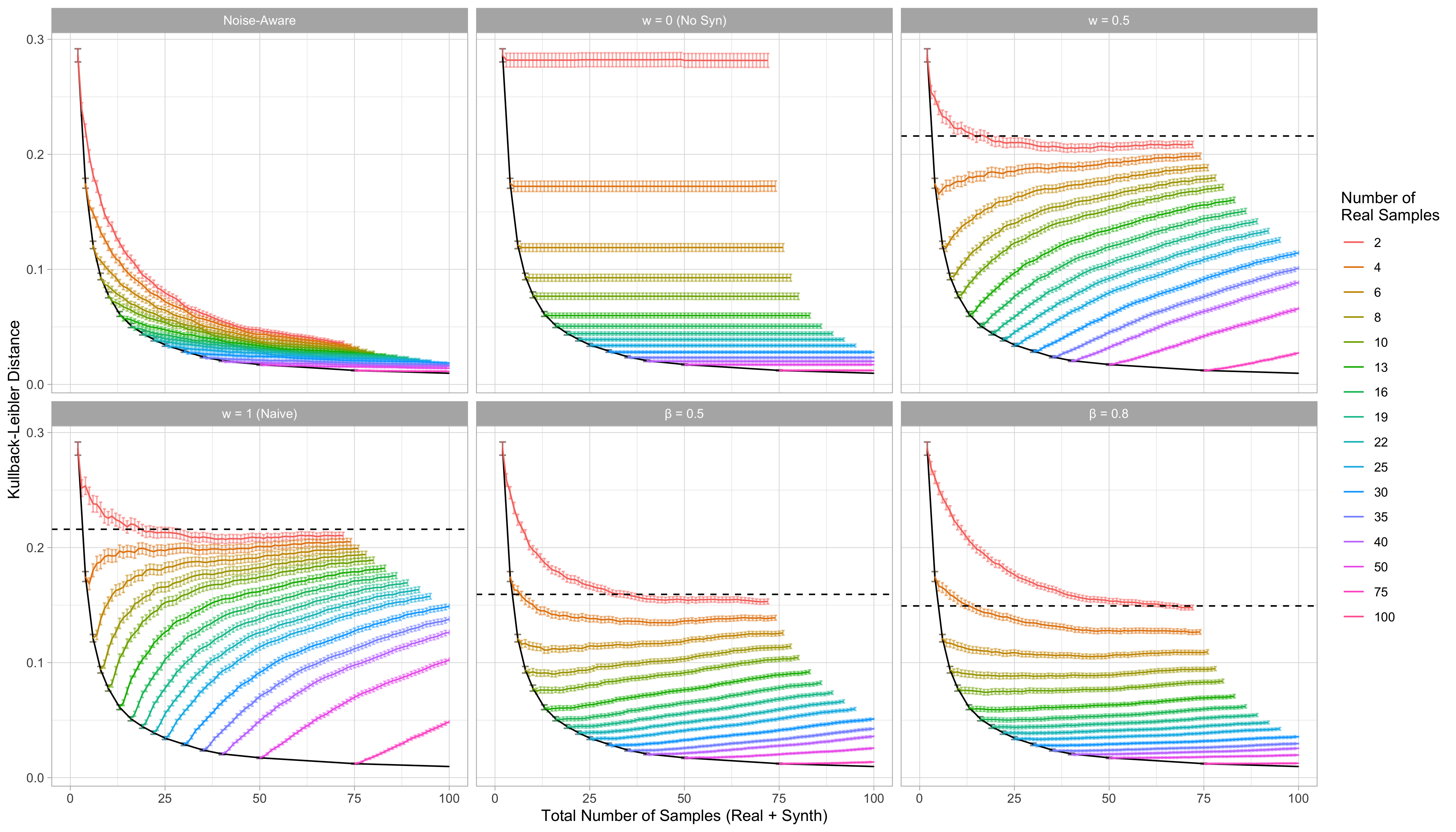}
\caption{
Branching plots for each model configuration in the case of the simulated Gaussian experiments illustrating the \KLD against the total number of samples where \DP of $\varepsilon = 6$ is achieved by the Laplace mechanism via noise of scale $\lambda = 1.0$.
}
\label{Fig:Branching2} 
\end{figure}

\begin{figure}[H]
\includegraphics[width=0.97\textwidth]{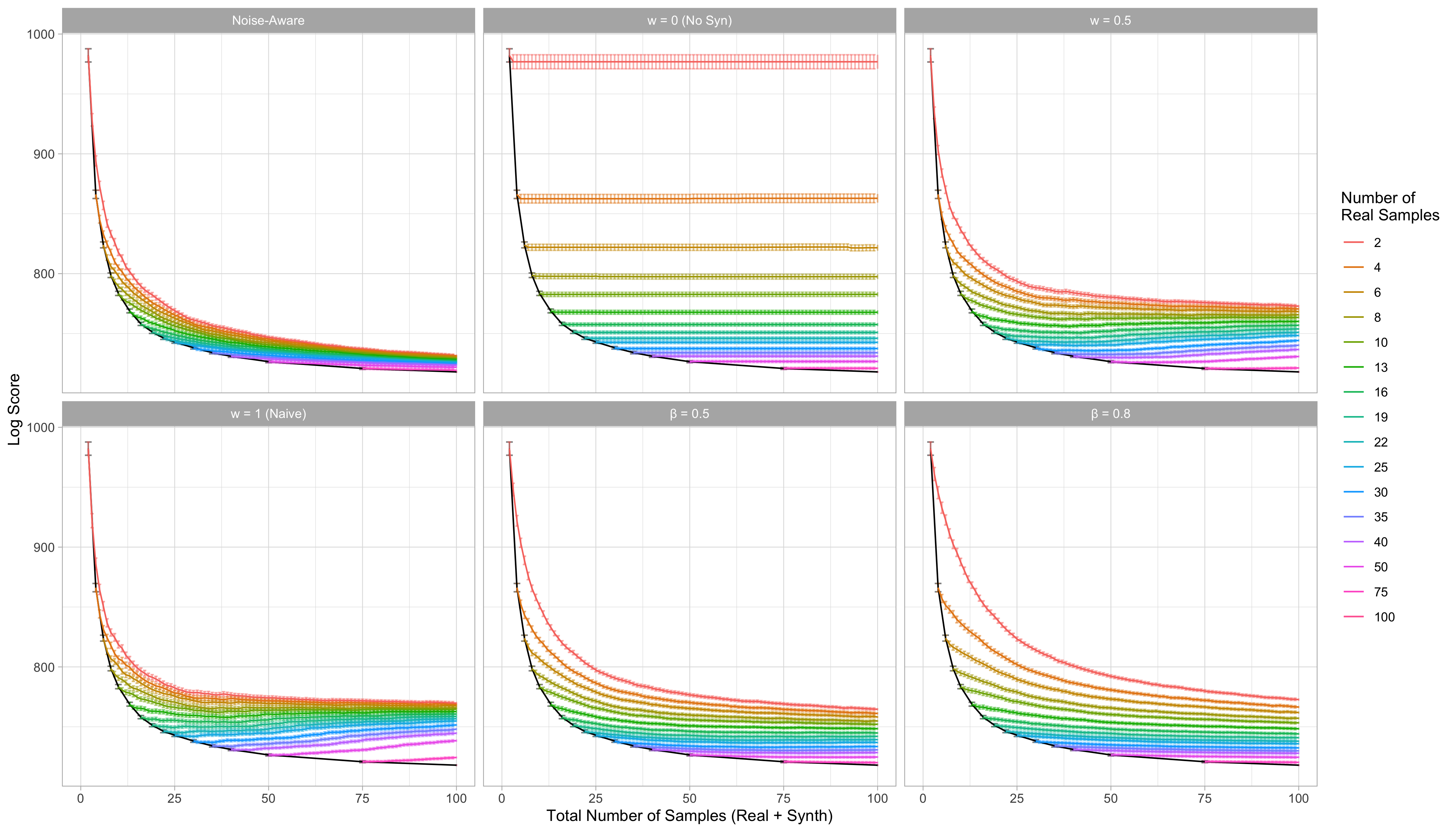}
\caption{
Branching plots for each model configuration in the case of the simulated Gaussian experiments illustrating the log score against the total number of samples where \DP of $\varepsilon = 8$ is achieved by the Laplace mechanism via noise of scale $\lambda = 0.75$.
}
\label{Fig:Branching3} 
\end{figure}

\begin{figure}[H]
\includegraphics[width=0.97\textwidth]{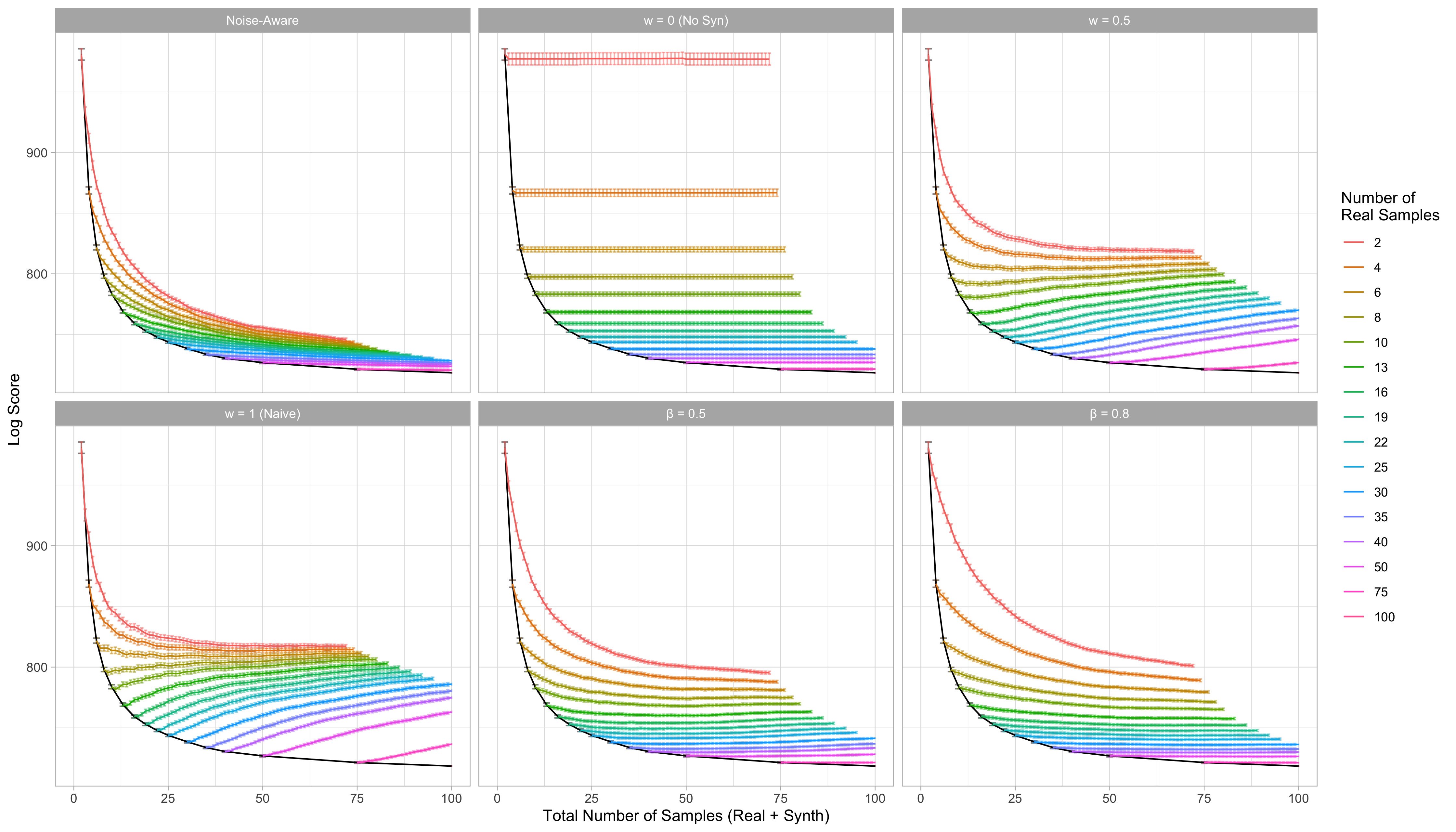}
\caption{
Branching plots for each model configuration in the case of the simulated Gaussian experiments illustrating the log score against the total number of samples where \DP of $\varepsilon = 6$ is achieved by the Laplace mechanism via noise of scale $\lambda = 1.0$.
}
\label{Fig:Branching4} 
\end{figure}

\begin{figure}[H]
\includegraphics[width=0.97\textwidth]{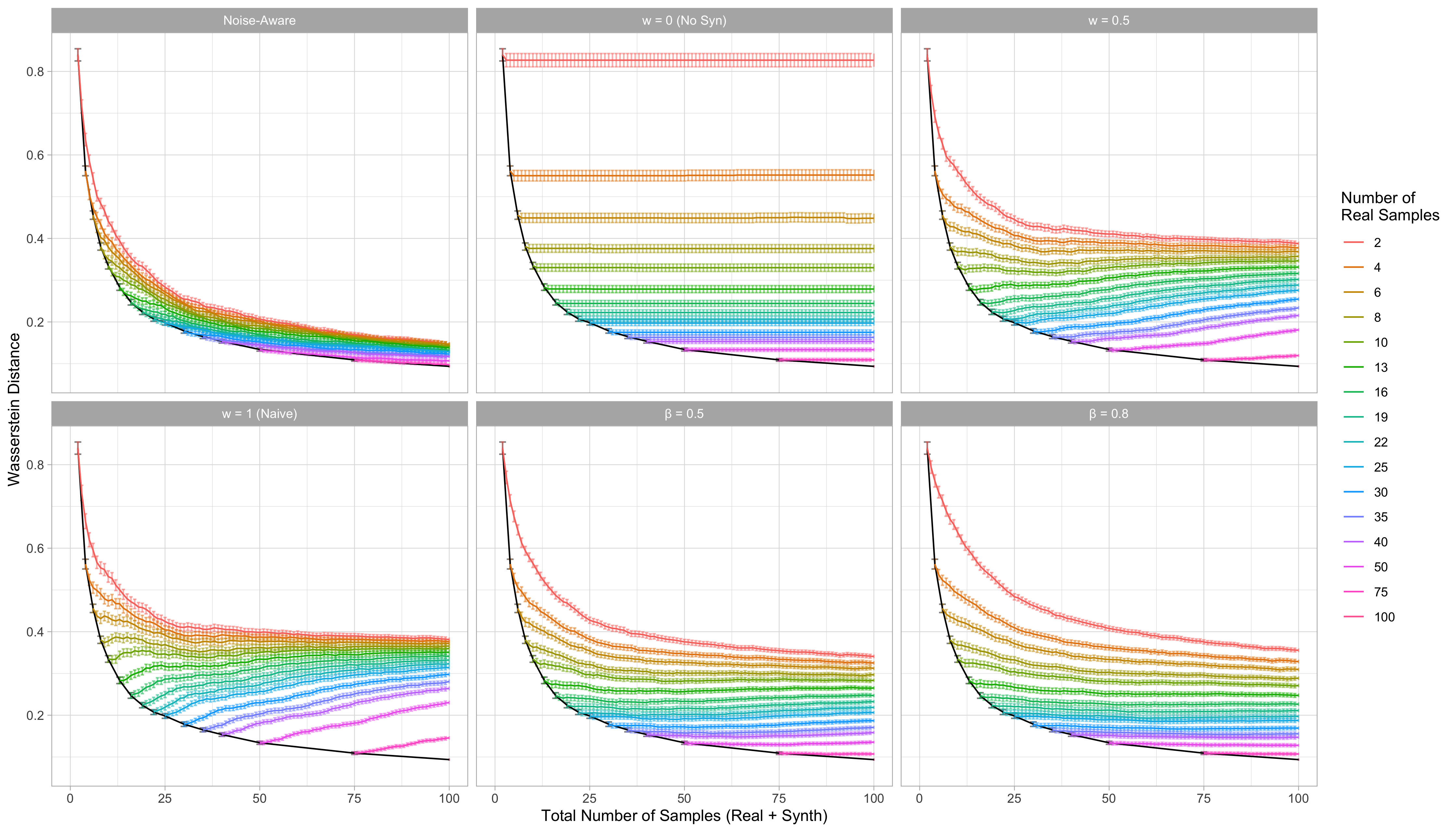}
\caption{
Branching plots for each model configuration in the case of the simulated Gaussian experiments illustrating the Wasserstein distance against the total number of samples where \DP of $\varepsilon = 8$ is achieved by the Laplace mechanism via noise of scale $\lambda = 0.75$.
}
\label{Fig:Branching5} 
\end{figure}

\begin{figure}[H]
\includegraphics[width=0.97\textwidth]{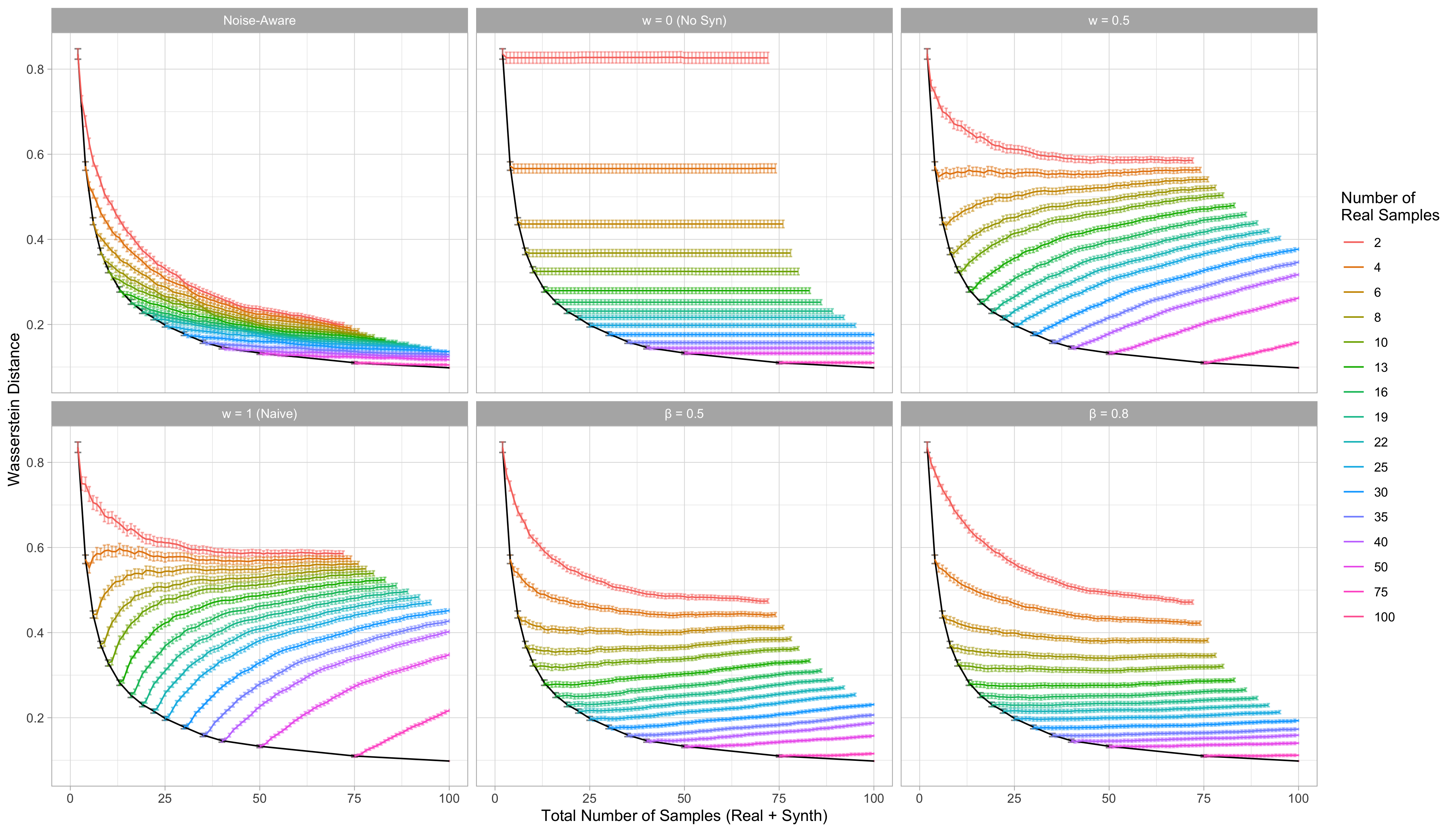}
\caption{
Branching plots for each model configuration in the case of the simulated Gaussian experiments illustrating the Wasserstein distance against the total number of samples where \DP of $\varepsilon = 6$ is achieved by the Laplace mechanism via noise of scale $\lambda = 1.0$.
}
\label{Fig:Branching6} 
\end{figure}

\begin{figure}[H]
\includegraphics[width=\textwidth]{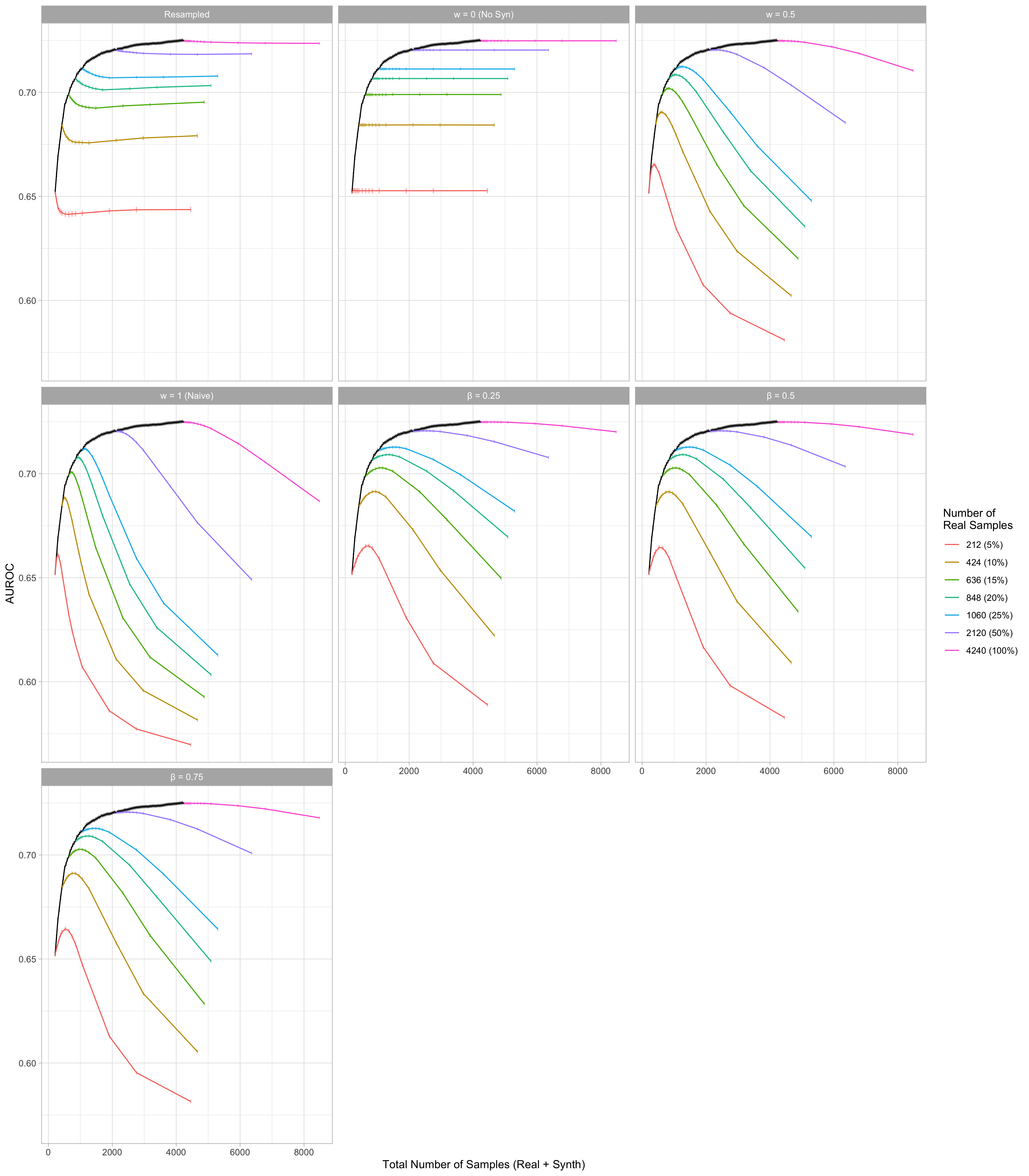}
\caption{
Branching plots for each model configuration in the case of the logistic regression experiments on the Framingham dataset illustrating the AUROC against the total number of samples where \DP of $\varepsilon = 6$ is achieved via generation of synthetic datasets using the \PATEGAN.
}
\label{Fig:Branching7} 
\end{figure}

\begin{figure}[H]
\includegraphics[width=\textwidth]{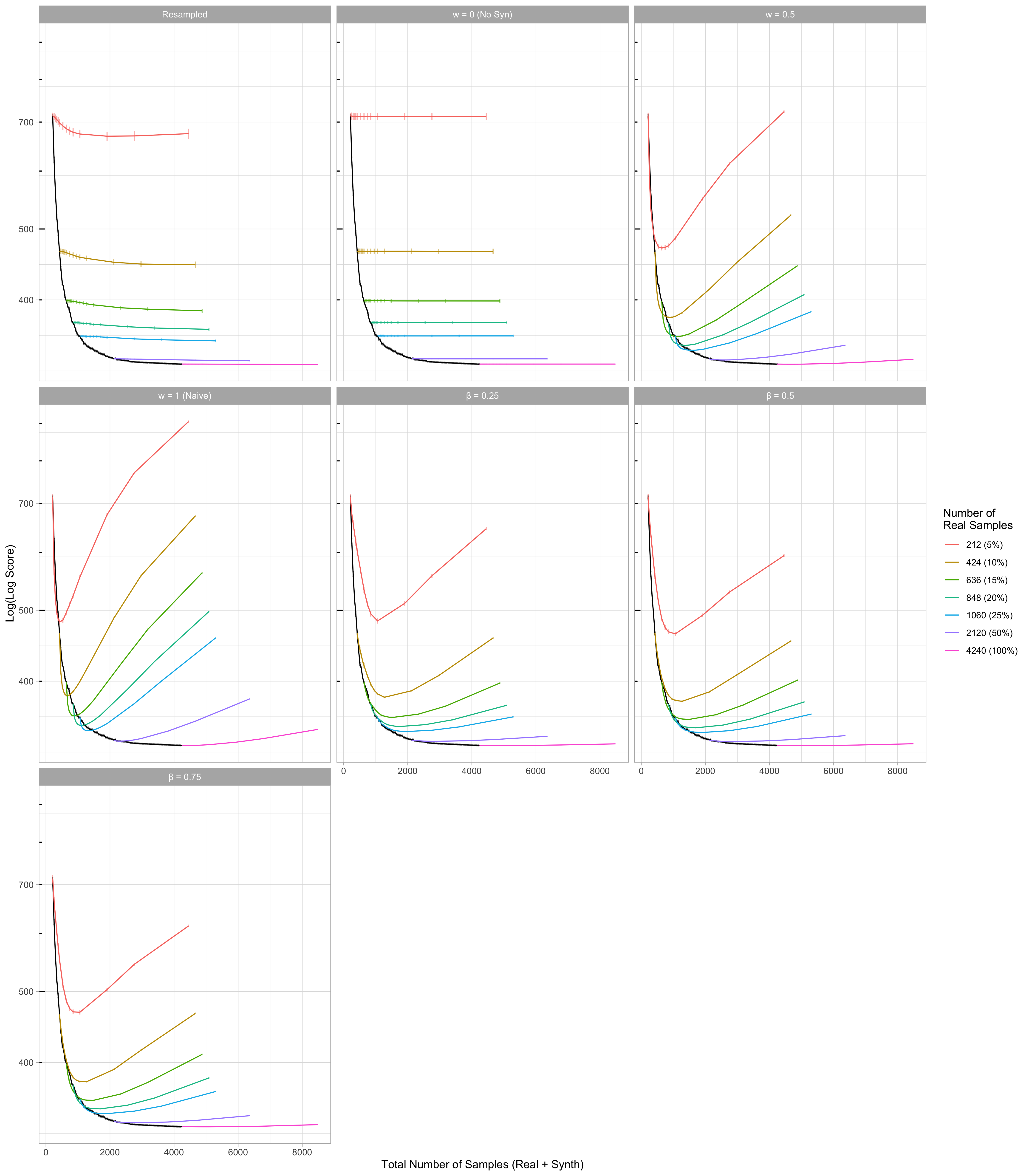}
\caption{
Branching plots for each model configuration in the case of the logistic regression experiments on the Framingham dataset illustrating the log score against the total number of samples where \DP of $\varepsilon = 6$ is achieved via generation of synthetic datasets using the \PATEGAN.
}
\label{Fig:Branching8} 
\end{figure}

\begin{figure}[H]
\includegraphics[width=\textwidth]{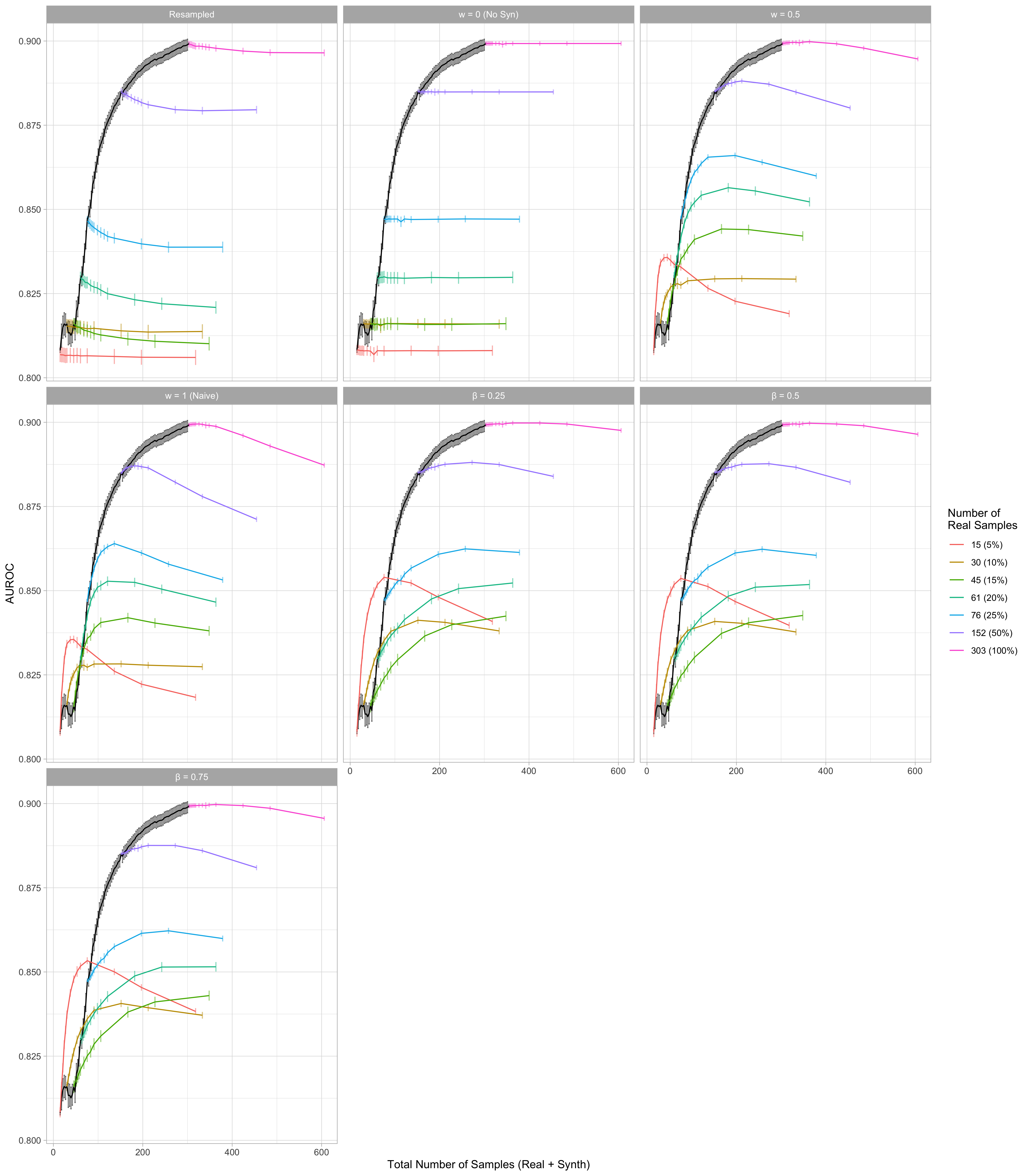}
\caption{
Branching plots for each model configuration in the case of the logistic regression experiments on the UCI Heart dataset illustrating the AUROC against the total number of samples where \DP of $\varepsilon = 6$ is achieved via generation of synthetic datasets using the \PATEGAN.
}
\label{Fig:Branching9} 
\end{figure}

\begin{figure}[H]
\includegraphics[width=\textwidth]{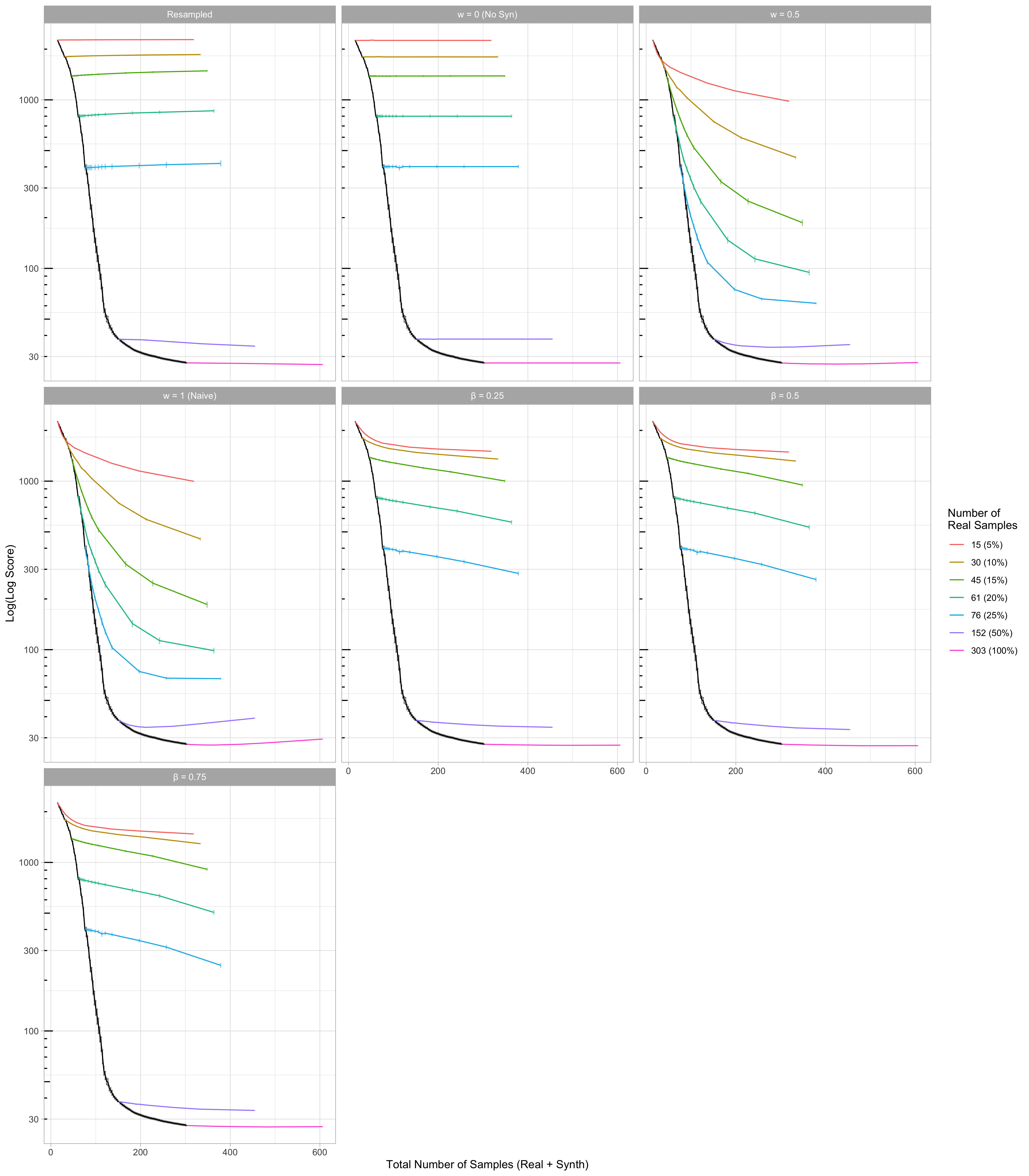}
\caption{
Branching plots for each model configuration in the case of the logistic regression experiments on the UCI Heart dataset illustrating the log score against the total number of samples where \DP of $\varepsilon = 6$ is achieved via generation of synthetic datasets using the \PATEGAN.
}
\label{Fig:Branching10} 
\end{figure}

Figures \ref{Fig:Branching1}, \ref{Fig:Branching2}, \ref{Fig:Branching3}, \ref{Fig:Branching4}, \ref{Fig:Branching5}, \ref{Fig:Branching6}, \ref{Fig:Branching7}, \ref{Fig:Branching8}, \ref{Fig:Branching9}, \ref{Fig:Branching10} show the full suite of branching plots for all of our experiments. In each of these plots we investigate different privatisation levels and criteria of interest; we then draw comparisons amongst the model configurations. In particular we see the finite and asymptotic effectiveness of the \BD across a wide range of $\beta$ when compared to a range of standard and reweighted approaches. The `Noise-Aware` models perform the best as is expected as these models are aware of the privatisation process and thus can go some way towards modelling it. In terms of \KLD especially, we can see this asymptotic effectiveness via the black dashed line representing $\KLD(F_0\,\|\,f_{\theta^{\ast}})$ for $\theta^{\ast} = \theta^{\KLD}_{\mathcal{G}_{\varepsilon, \delta}}$ in the case of models involving $w$ and $\theta^{\ast} = \theta^{\BD}_{\mathcal{G}_{\varepsilon, \delta}}$ for models involving $\beta$. These quantities represent the approximation to $F_0$ given an infinite sample from $\mathcal{G}_{\varepsilon, \delta}$ under the two model types. It can be seen that a relatively small increase in noise / privatisation in the Gaussian experiments can quite drastically change the effectiveness of using synthetic data, such that only when very little real data is available should its use be considered at all. For the logistic regression experiment datasets, we actually see a reduction in performance when using the \BD on the UCI Heart dataset. This is likely due to the \BD's natural tendency to downweight samples, as upon inspection it appears that the two model families will converge to roughly similar criterion values.

\subsubsection{Model Comparison Plots}

\begin{figure}[H]
\includegraphics[width=\textwidth]{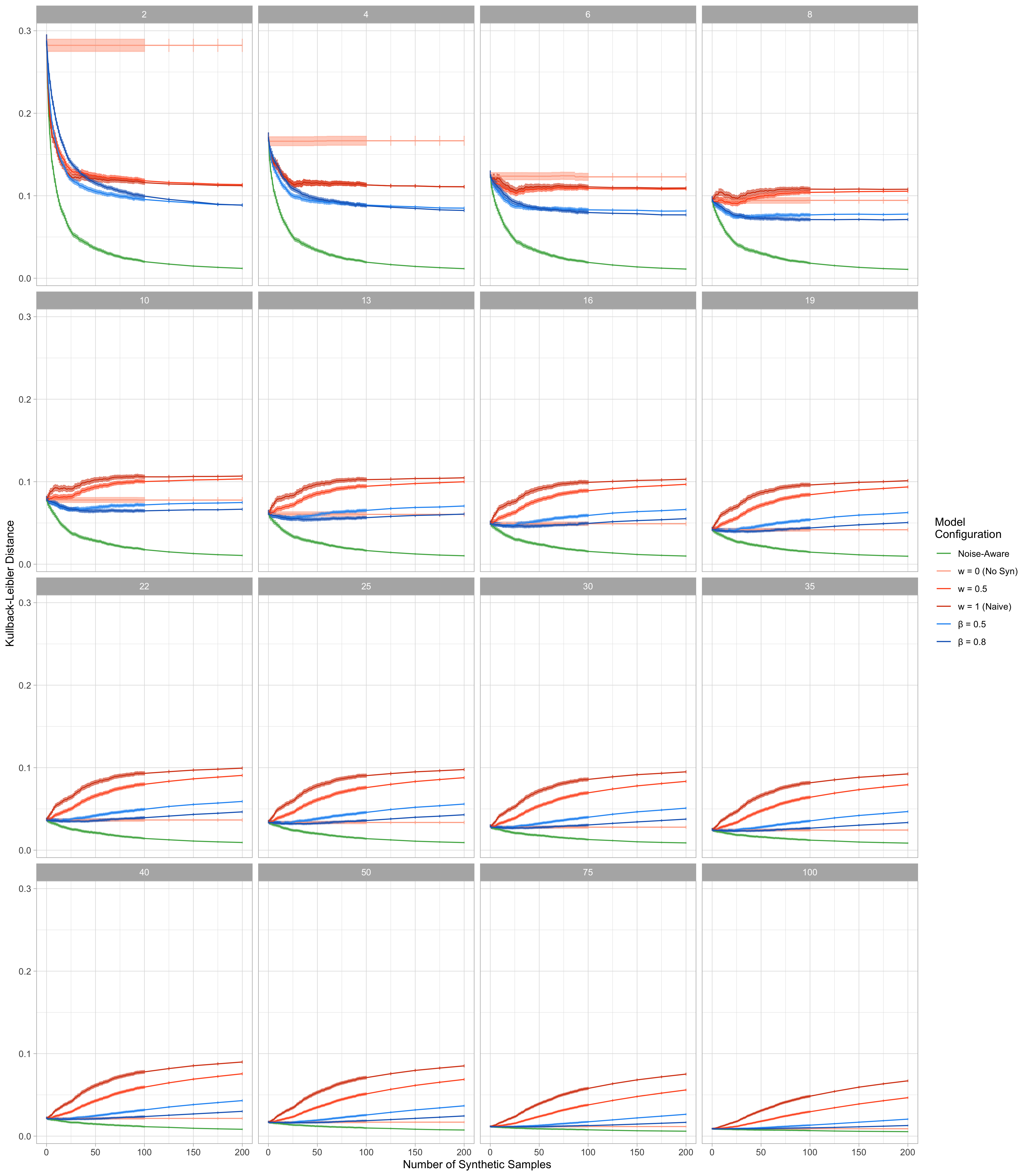}
\caption{
Model comparison plots for each real data quantity $n_L$ in the case of the simulated Gaussian experiments illustrating the \KLD against the number of synthetic samples where \DP of $\varepsilon = 8$ is achieved by the Laplace mechanism via noise of scale $\lambda = 0.75$.
}
\label{Fig:ModelComp1} 
\end{figure}

\begin{figure}[H]
\includegraphics[width=\textwidth]{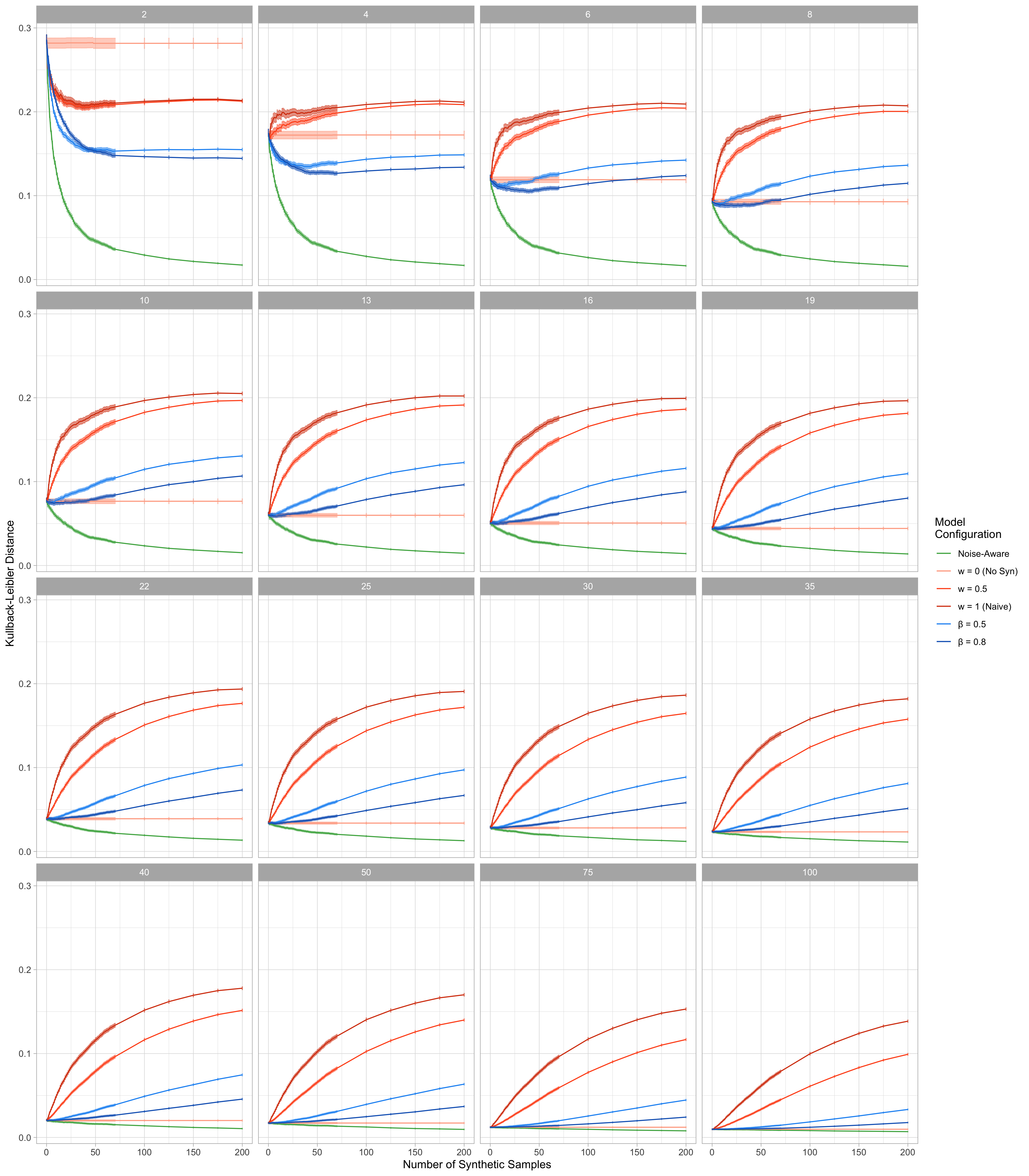}
\caption{
Model comparison plots for each real data quantity $n_L$ in the case of the simulated Gaussian experiments illustrating the \KLD against the number of synthetic samples where \DP of $\varepsilon = 6$ is achieved by the Laplace mechanism via noise of scale $\lambda = 1.0$.
}
\label{Fig:ModelComp2} 
\end{figure}

\begin{figure}[H]
\includegraphics[width=\textwidth]{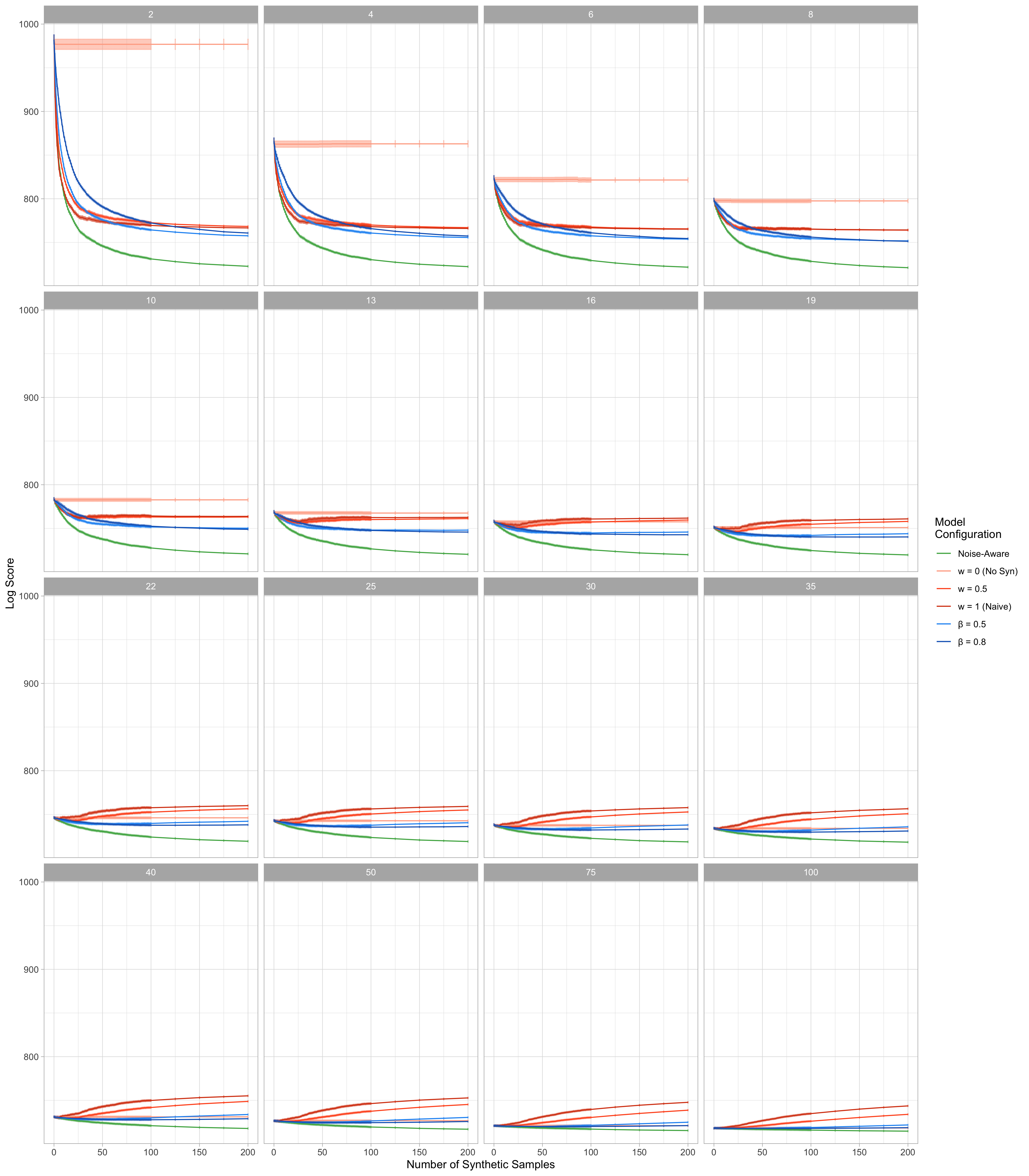}
\caption{
Model comparison plots for each real data quantity $n_L$ in the case of the simulated Gaussian experiments illustrating the log score against the number of synthetic samples where \DP of $\varepsilon = 8$ is achieved by the Laplace mechanism via noise of scale $\lambda = 0.75$.
}
\label{Fig:ModelComp3} 
\end{figure}

\begin{figure}[H]
\includegraphics[width=\textwidth]{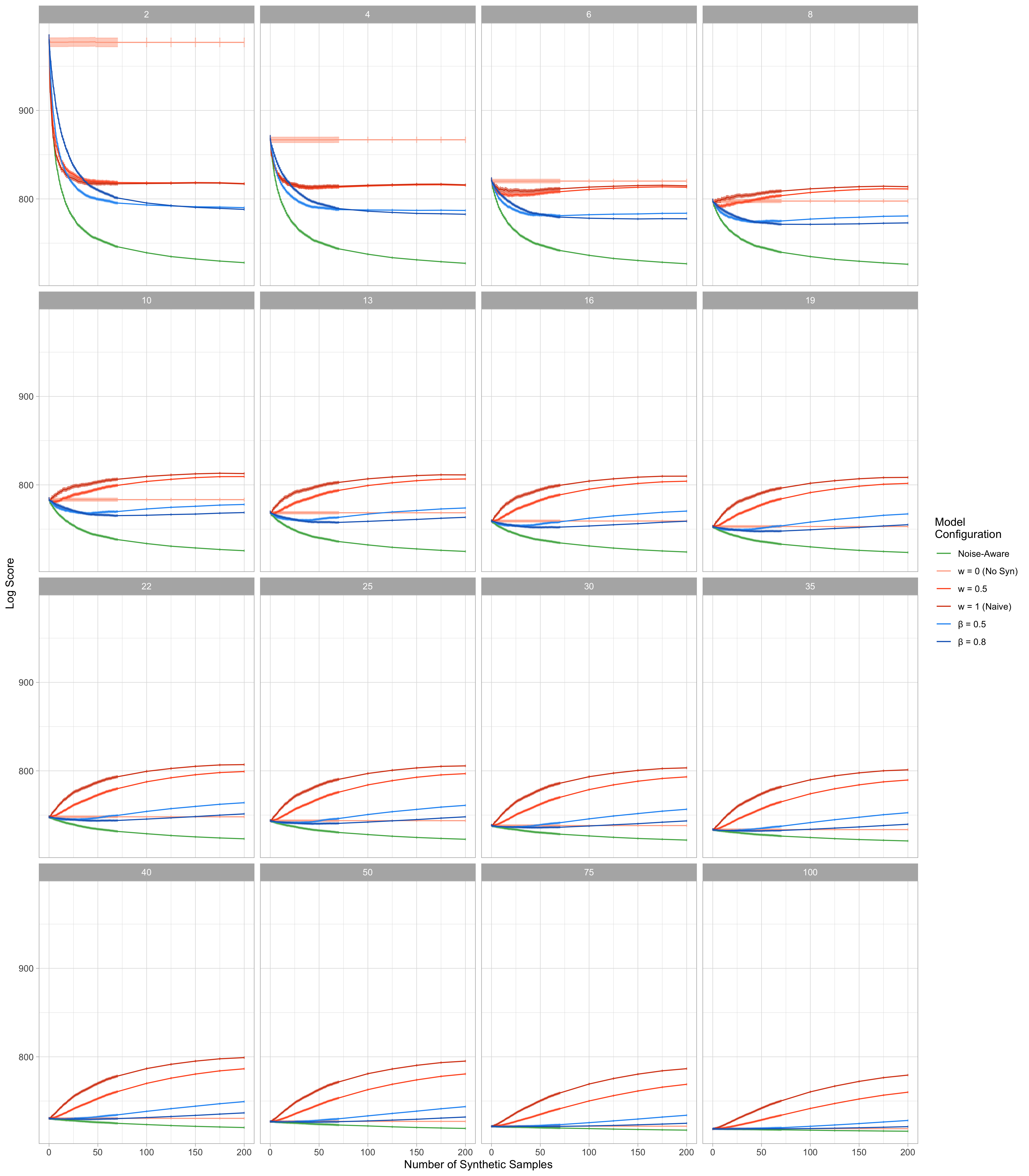}
\caption{
Model comparison plots for each real data quantity $n_L$ in the case of the simulated Gaussian experiments illustrating the log score against the number of synthetic samples where \DP of $\varepsilon = 6$ is achieved by the Laplace mechanism via noise of scale $\lambda = 1.0$.
}
\label{Fig:ModelComp4} 
\end{figure}

\begin{figure}[H]
\includegraphics[width=\textwidth]{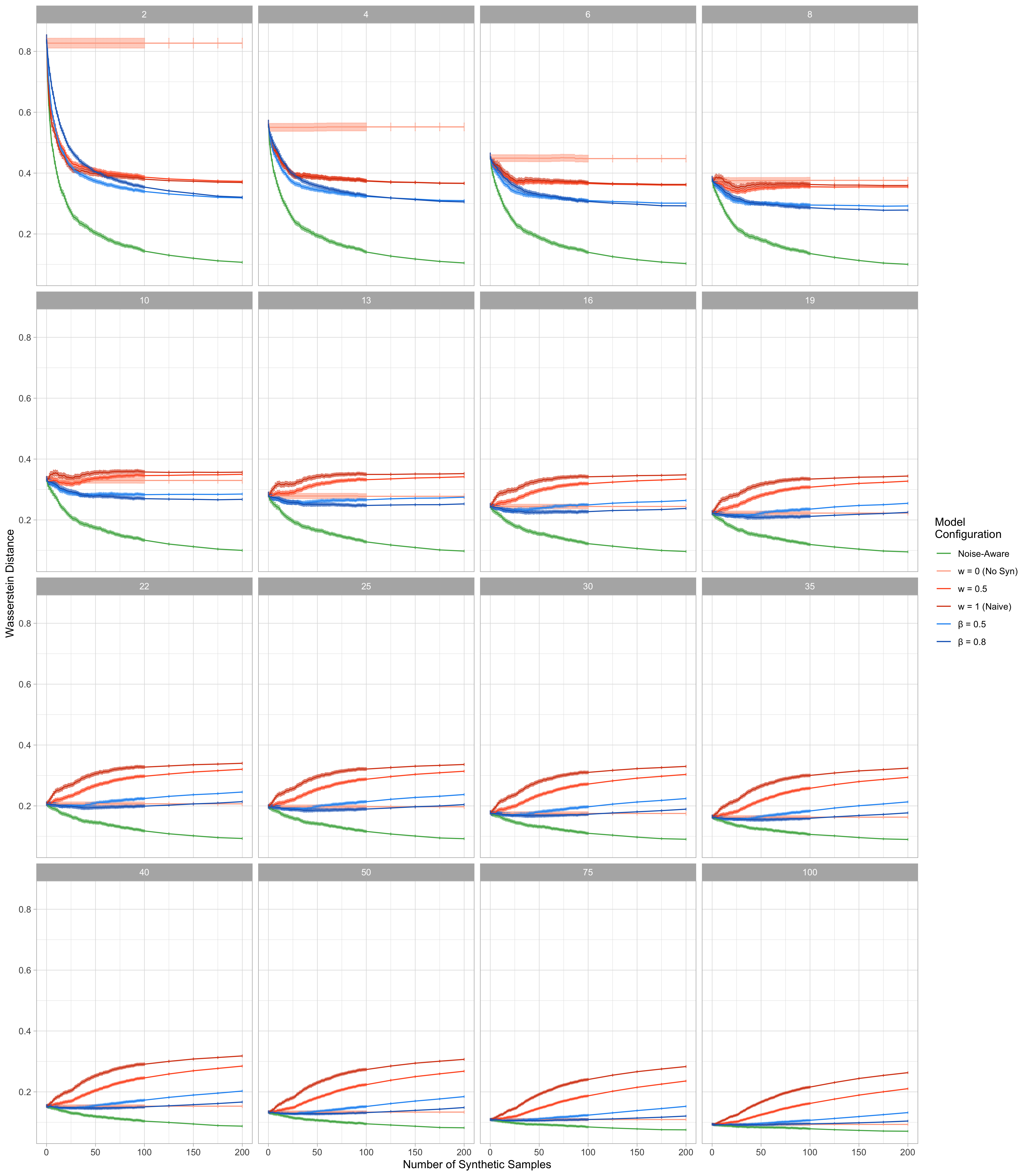}
\caption{
Model comparison plots for each real data quantity $n_L$ in the case of the simulated Gaussian experiments illustrating the Wasserstein distance against the number of synthetic samples where \DP of $\varepsilon = 8$ is achieved by the Laplace mechanism via noise of scale $\lambda = 0.75$.
}
\label{Fig:ModelComp5} 
\end{figure}

\begin{figure}[H]
\includegraphics[width=\textwidth]{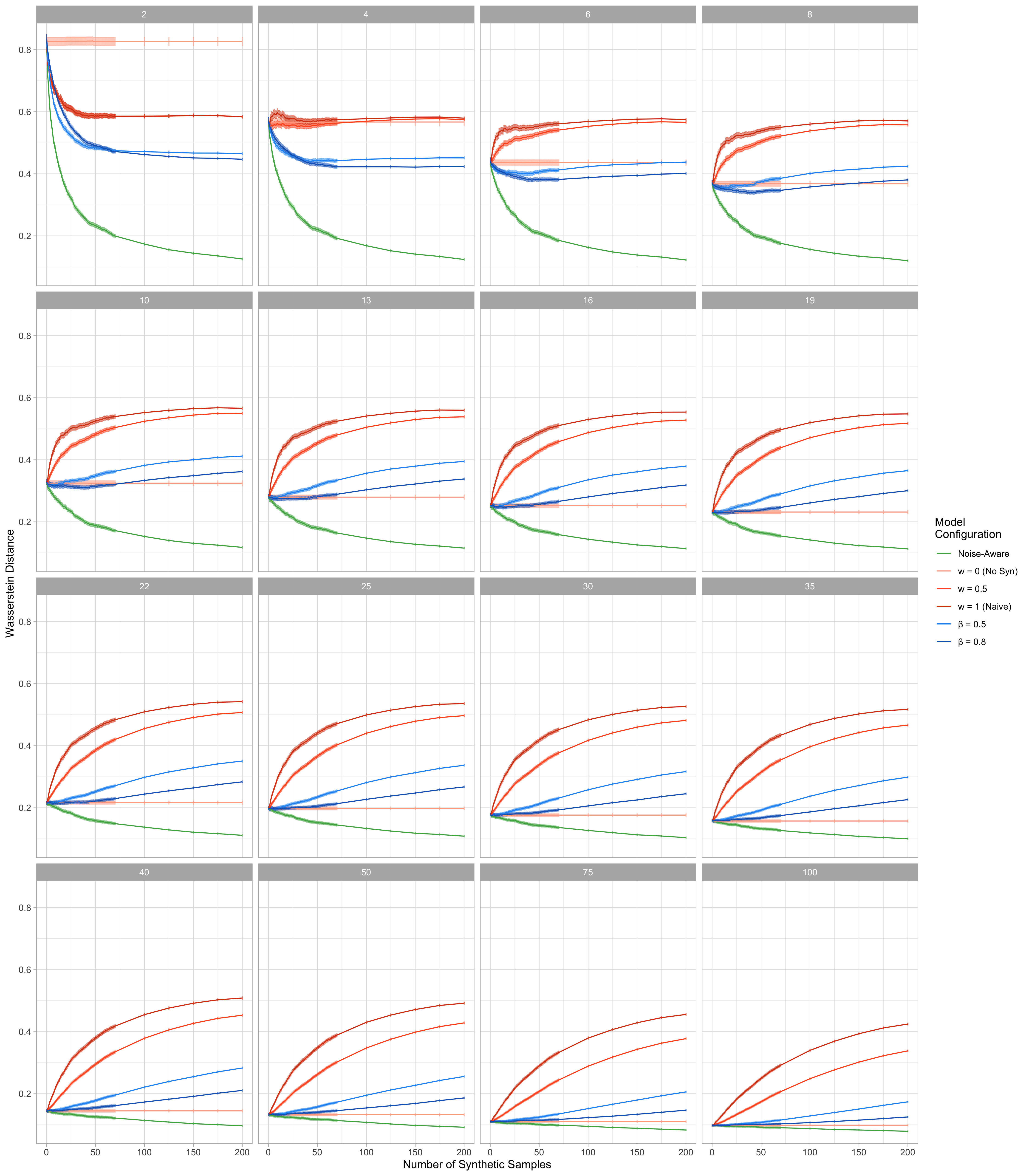}
\caption{
Model comparison plots for each real data quantity $n_L$ in the case of the simulated Gaussian experiments illustrating the Wasserstein distance against the number of synthetic samples where \DP of $\varepsilon = 6$ is achieved by the Laplace mechanism via noise of scale $\lambda = 1.0$.
}
\label{Fig:ModelComp6} 
\end{figure}

\begin{figure}[H]
\includegraphics[width=\textwidth]{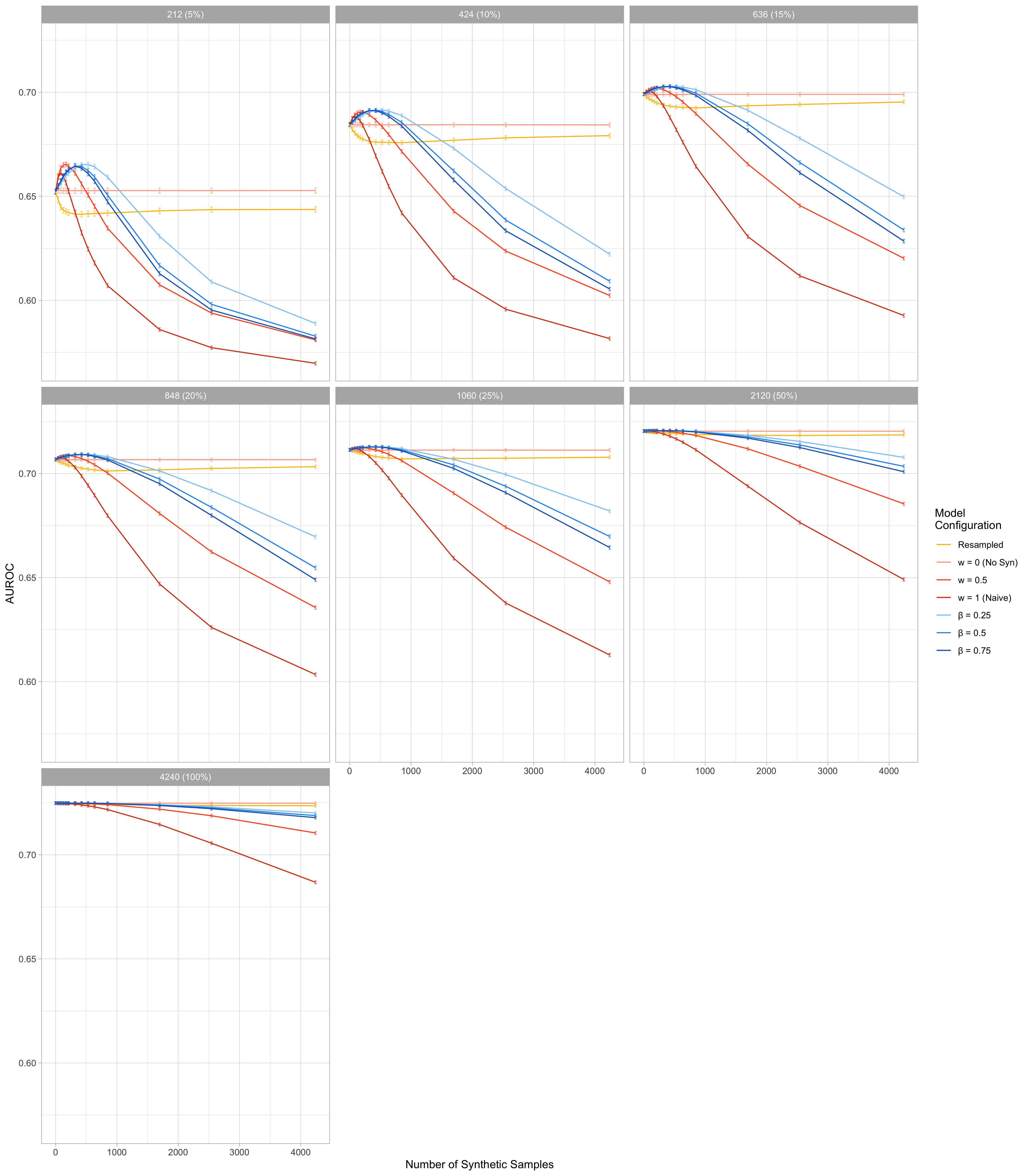}
\caption{
Model comparison plots for each real data quantity $n_L$ in the case of the logistic regression experiments on the Framingham dataset illustrating the AUROC against the number of synthetic samples where \DP of $\varepsilon = 6$ is achieved via generation of synthetic datasets using the \PATEGAN.
}
\label{Fig:ModelComp7} 
\end{figure}

\begin{figure}[H]
\includegraphics[width=\textwidth]{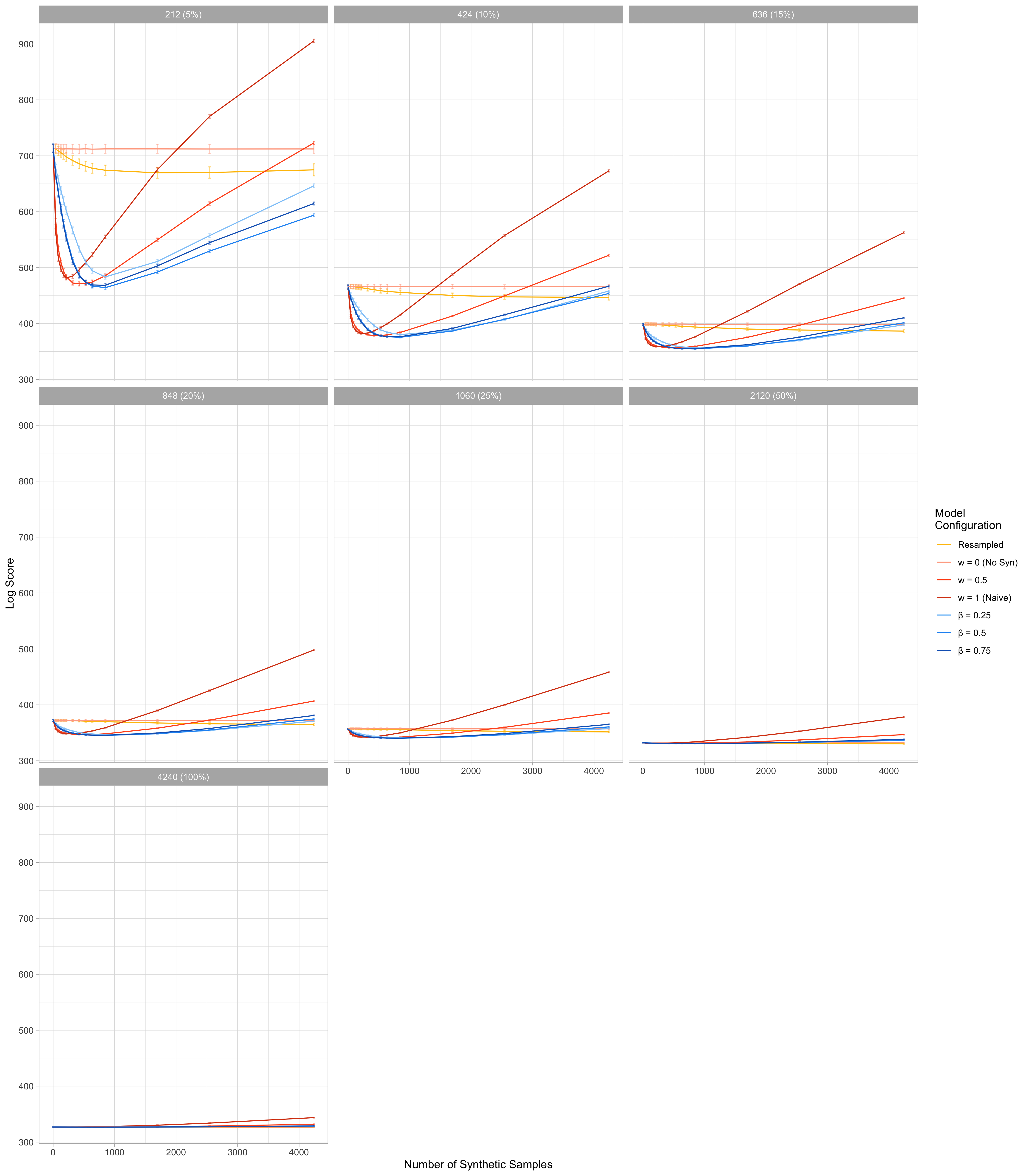}
\caption{
Model comparison plots for each real data quantity $n_L$ in the case of the logistic regression experiments on the Framingham dataset illustrating the AUROC against the number of synthetic samples where \DP of $\varepsilon = 6$ is achieved via generation of synthetic datasets using the \PATEGAN.
}
\label{Fig:ModelComp8} 
\end{figure}

\begin{figure}[H]
\includegraphics[width=\textwidth]{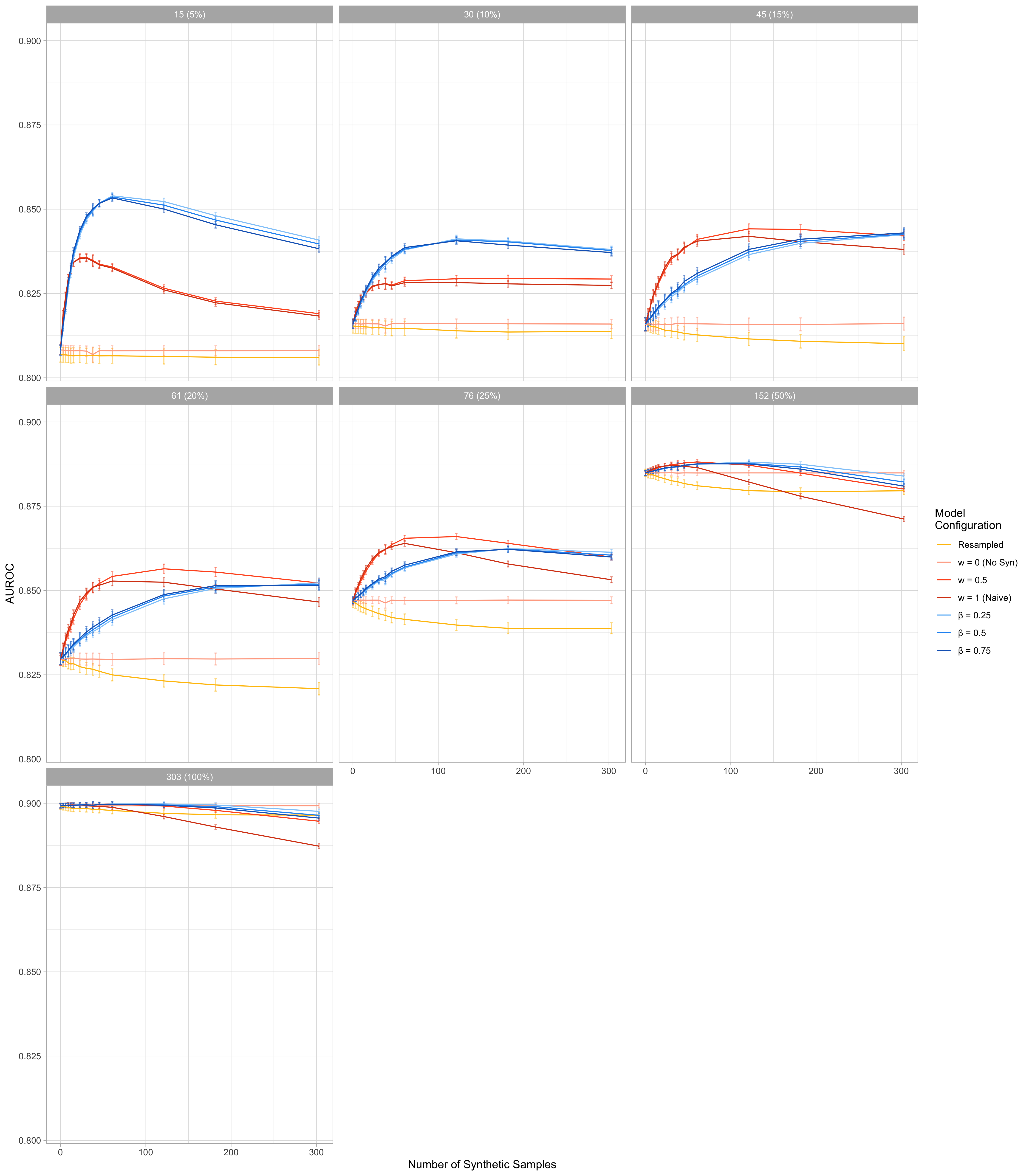}
\caption{
Model comparison plots for each real data quantity $n_L$ in the case of the logistic regression experiments on the UCI Heart dataset illustrating the AUROC against the number of synthetic samples where \DP of $\varepsilon = 6$ is achieved via generation of synthetic datasets using the \PATEGAN.
}
\label{Fig:ModelComp9} 
\end{figure}

\begin{figure}[H]
\includegraphics[width=\textwidth]{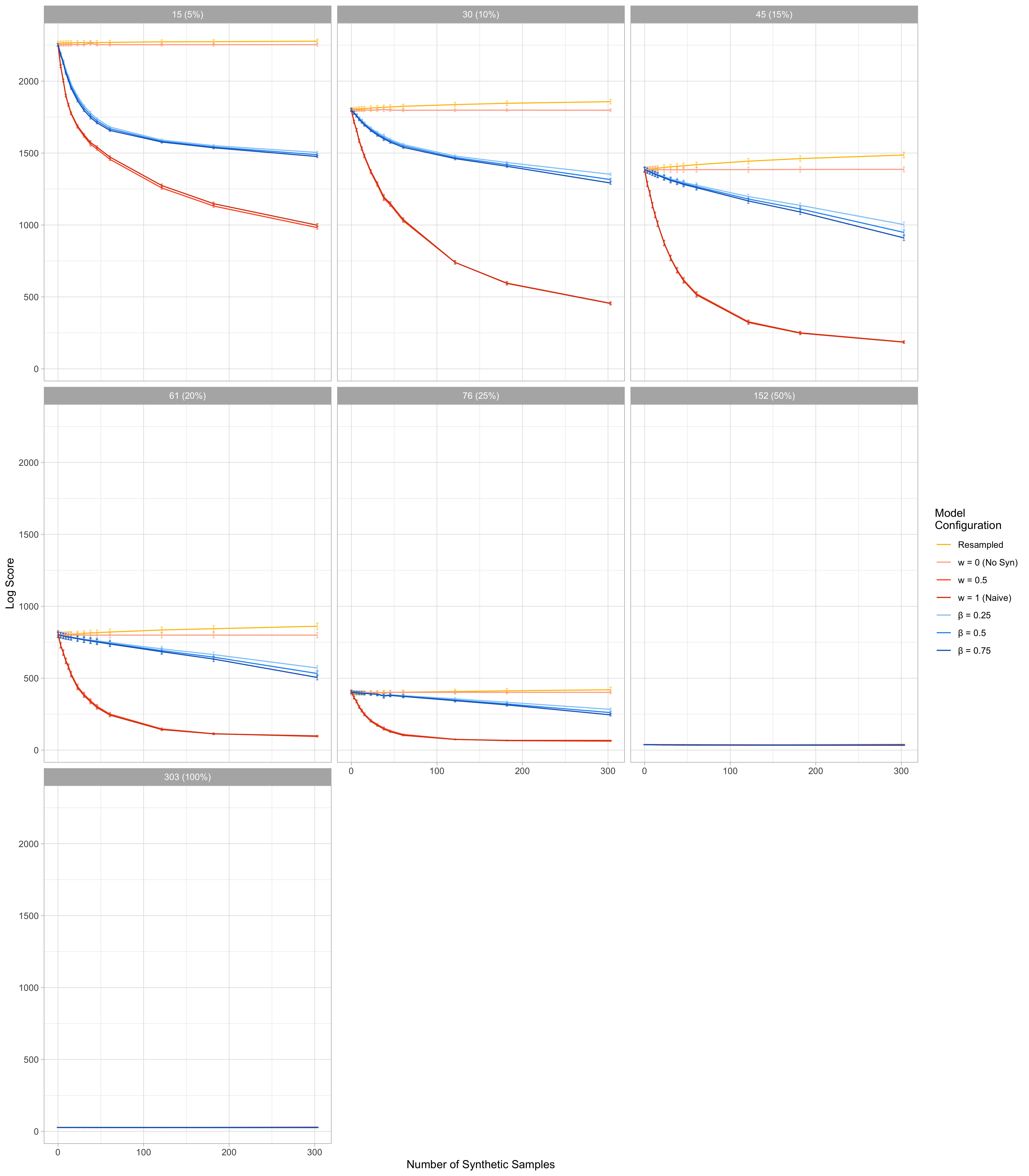}
\caption{
Model comparison plots for each real data quantity $n_L$ in the case of the logistic regression experiments on the UCI Heart dataset illustrating the AUROC against the number of synthetic samples where \DP of $\varepsilon = 6$ is achieved via generation of synthetic datasets using the \PATEGAN.
}
\label{Fig:ModelComp10} 
\end{figure}

Figures \ref{Fig:ModelComp1}, \ref{Fig:ModelComp2}, \ref{Fig:ModelComp3}, \ref{Fig:ModelComp4}, \ref{Fig:ModelComp5}, \ref{Fig:ModelComp6}, \ref{Fig:ModelComp7}, \ref{Fig:ModelComp8}, \ref{Fig:ModelComp9}, \ref{Fig:ModelComp10} show the full suite of model comparison plots for all of our experiments. These plots allow us to directly compare the performance of different model configurations given an identical amount of real data $n_L$ and an accompanying, varying amount of synthetic data. We can compare the most desirable $\hat{m}$ values achieved by all of the models to observe that again, other than in the case of the UCI Heart dataset, the \BD consistently performs well in comparison to other approaches. The models where $w$ is set to 0 offer a baseline of sorts, with the points where each other model's curve crosses its straight line representing when the various other model configurations stop being justifiable and synthetic data should not be used at all. It can be seen that not only does the \BD achieve more desirable performance in the majority of cases, but it also remains robust to large amounts of synthetic data where other approaches fail to be effective and quickly lose out through the use of synthetic data. We offer all of these plots on fixed axes across each grid so that a reader can notice the narrowing scope and magnitude to which synthetic data is useful as more real data samples become available; this highlights the advantages of the \BD in that it is a `safer' option even in the situations where synthetic data cannot help the inference much in that it is more robust to the damage it can cause, especially when a user may not be able to calculate $\hat{m}$ explicitly.

\subsubsection{$n$-Effective Plots}

\begin{figure}[H]
\includegraphics[width=\textwidth]{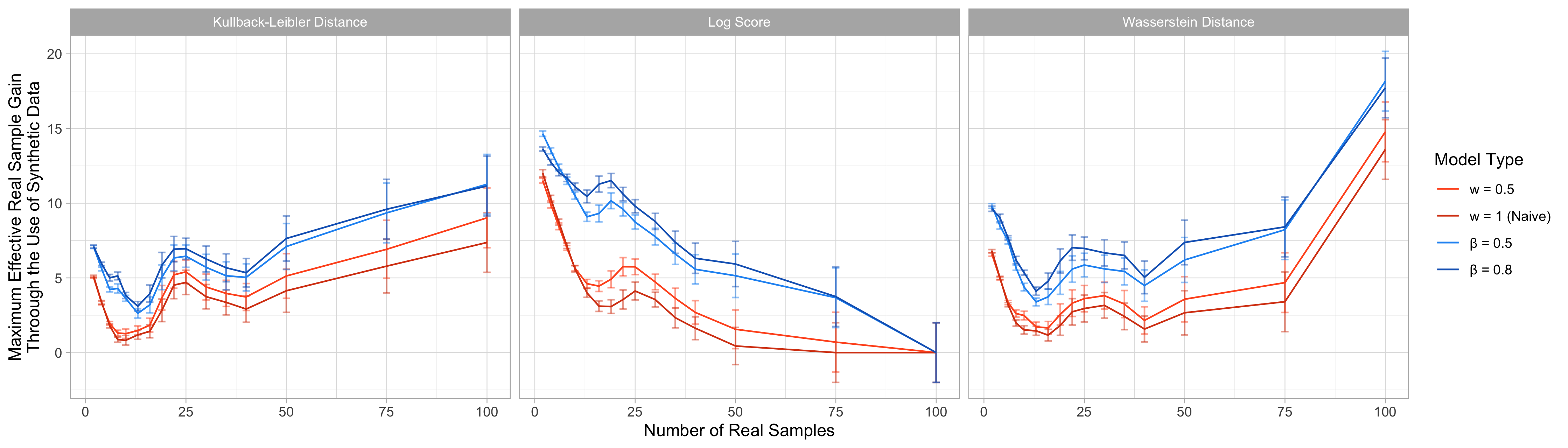}
\caption{
$n$-effective plots plots for each of the relevant criteria in the case of the simulated Gaussian experiments illustrating the effective number of real samples to be gained through the use of synthetic data at each amount of real data $n_L$ where \DP of $\varepsilon = 8$ is achieved by the Laplace mechanism via noise of scale $\lambda = 0.75$.
}
\label{Fig:Neff1} 
\end{figure}

\begin{figure}[H]
\includegraphics[width=\textwidth]{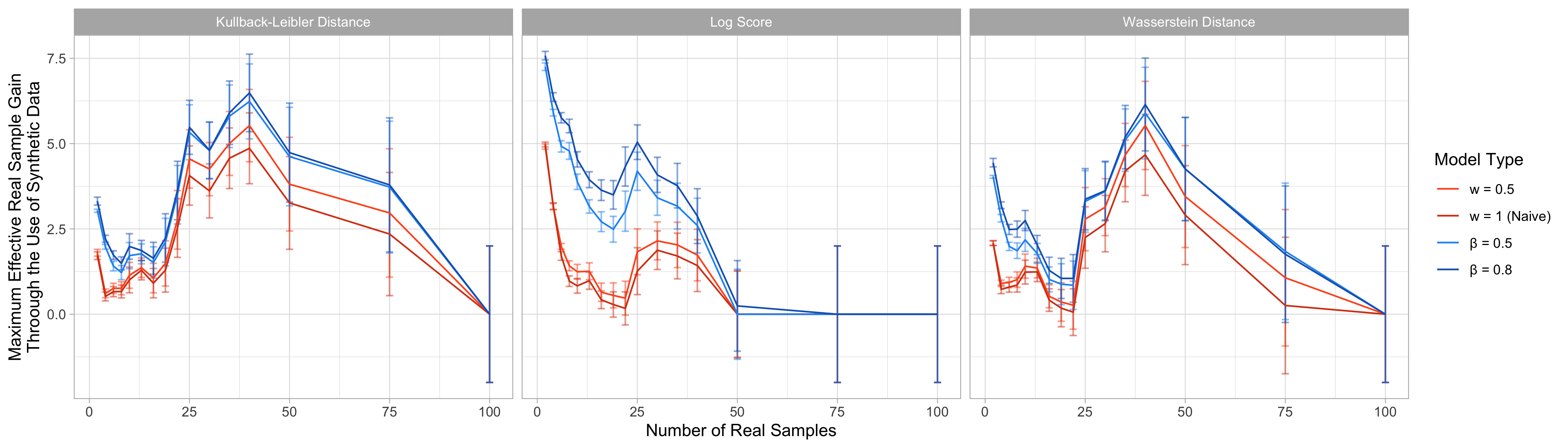}
\caption{
$n$-effective plots plots for each of the relevant criteria in the case of the simulated Gaussian experiments illustrating the effective number of real samples to be gained through the use of synthetic data at each amount of real data $n_L$ where \DP of $\varepsilon = 6$ is achieved by the Laplace mechanism via noise of scale $\lambda = 1.0$.
}
\label{Fig:Neff2} 
\end{figure}

\begin{figure}[H]
\includegraphics[width=\textwidth]{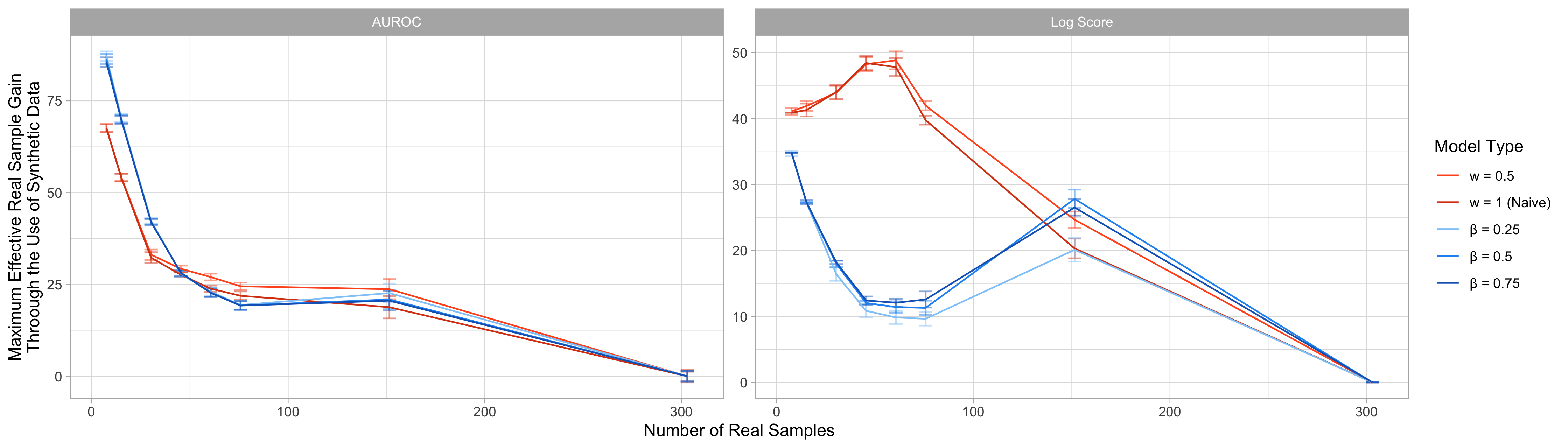}
\caption{
$n$-effective plots plots for each of the relevant criteria in the case of the logistic regression experiments on the UCI Heart dataset illustrating the effective number of real samples to be gained through the use of synthetic data at each amount of real data $n_L$ where \DP of $\varepsilon = 6$ is achieved via generation of synthetic datasets using the \PATEGAN.
}
\label{Fig:Neff3} 
\end{figure}

Figures \ref{Fig:Neff1}, \ref{Fig:Neff2}, \ref{Fig:Neff3} show the full suite of $n$-effective plots for all of our experiments, other than for the Framingham dataset which is already included in the main text. In these plots we observe some interesting phenomena, primarily in that the number of real effective samples to gain through synthetic data is related to the amount of real data that has already been used; in general we observe asymptotic behaviour in the criteria as the amount of real data increases meaning variation in the performance of synthetic data and the resulting turning points can indicate a greater amount of effective samples gained despite the actual criterion value improvement being marginal. As such, reading the x axis is in some ways misleading in that effective samples often `mean more' in the sense that they indicate a greater improvement in the criteria under a smaller total amount of data. Again, we see that in the case of the UCI Heart dataset, the improvements offered by the \BD are less significant and in some cases non-existent over standard approaches.

\newpage

\footnotetext{Note that this is not consistently the case for \GAN based methods as the utility of synthetic data is also limited by how well the \GAN can initially capture $F_0$ from its training data regardless of the chosen $\varepsilon$.}

\newpage

\bibliography{aistats_2021.bib}

% \section{FORMATTING INSTRUCTIONS}

% To prepare a supplementary pdf file, we ask the authors to use \texttt{aistats2021.sty} as a style file and to follow the same formatting instructions as in the main paper.
% The only difference is that the supplementary material must be in a \emph{single-column} format.
% You can use \texttt{supplement.tex} in our starter pack as a starting point, or append the supplementary content to the main paper and split the final PDF into two separate files.

% Note that reviewers are under no obligation to examine your supplementary material.

% \section{MISSING PROOFS}

% The supplementary materials may contain detailed proofs of the results that are missing in the main paper.

% \subsection{Proof of Lemma 3}

% \textit{In this section, we present the detailed proof of Lemma 3 and then [ ... ]}

% \section{ADDITIONAL EXPERIMENTS}

% If you have additional experimental results, you may include them in the supplementary materials.

% \subsection{The Effect of Regularization Parameter}

% \textit{Our algorithm depends on the regularization parameter $\lambda$. Figure 1 below illustrates the effect of this parameter on the performance of our algorithm. As we can see, [ ... ]}

% \vfill